\documentclass{elsarticle}
\usepackage{graphicx}
\usepackage[Symbol]{upgreek}
\usepackage{array}
\usepackage{graphicx}                    
\usepackage{enumerate}
\usepackage{amsmath}                     
\usepackage{amsfonts}                    
\usepackage{amssymb}                     
\usepackage{verbatim}
\usepackage{tabularx}
\usepackage{arydshln}
\usepackage{makeidx}
\usepackage{blkarray}
\usepackage{amsthm}
\usepackage[dvipdfm,
            bookmarksnumbered=false,
            bookmarksopen=true,
            colorlinks=true, 
            linkcolor=blue,
            ]{hyperref}
\usepackage{multirow}
\usepackage{bigdelim}
\usepackage{algorithm}
\usepackage{algorithmic}
\usepackage{bm}
\usepackage{natbib}
\usepackage{subfigure}
\usepackage{cases}
\usepackage{mathrsfs}

\newtheorem{thm}{Theorem}
\newtheorem{exm}{Example}
\newtheorem{dfn}{Definition}
\newtheorem*{pf}{Proof}

\newtheorem{cor}{Corollary}

\begin{document}

\begin{frontmatter}
\title{Heuristic algorithms for finding distribution reducts in probabilistic rough
set model
}
\author[swjt]{Xi'ao Ma\corref{cor1}}
\ead{maxiao73559{@}163.com}
\author[cqupt]{Guoyin Wang}
\ead{wanggy@ieee.org}
\author[cqupt]{Hong Yu}
\ead{hongyu.cqupt{@}gmail.com}

\address[swjt]{School of Computer $\&$ Information Engineering, Zhejiang Gongshang University, Hangzhou, 310018, P.R. China}
\address[cqupt]{Chongqing Key Laboratory of Computational Intelligence, Chongqing University
of Posts and Telecommunications, Chongqing, 400065, P.R. China}

\cortext[cor1]{Corresponding author. Tel: +86 13588030950}

\begin{abstract}
Attribute reduction is one of the most important topics in rough set
theory. Heuristic attribute reduction algorithms have been presented
to solve the attribute reduction problem. It is generally known that
fitness functions play a key role in developing heuristic attribute
reduction algorithms. The monotonicity of fitness functions can
guarantee the validity of heuristic attribute reduction algorithms.
In probabilistic rough set model, distribution reducts can ensure
the decision rules derived from the reducts are compatible with
those derived from the original decision table. However, there are
few studies on developing heuristic attribute reduction algorithms
for finding distribution reducts. This is partly due to the fact
that there are no monotonic fitness functions that are used to
design heuristic attribute reduction algorithms in probabilistic
rough set model. The main objective of this paper is to develop
heuristic attribute reduction algorithms for finding distribution
reducts in probabilistic rough set model. For one thing, two
monotonic fitness functions are constructed, from which equivalence
definitions of distribution reducts can be obtained. For another,
two modified monotonic fitness functions are proposed to evaluate
the significance of attributes more effectively. On this basis, two
heuristic attribute reduction algorithms for finding distribution
reducts are developed based on addition-deletion method and deletion
method. In particular, the monotonicity of fitness functions
guarantees the rationality of the  proposed heuristic attribute
reduction algorithms. Results of experimental analysis are included
to quantify the effectiveness of the proposed fitness functions and
distribution reducts.
\end{abstract}
\end{frontmatter}

\section{Introduction}

Rough set theory, introduced by Pawlak \cite{Pawlak1982RS}, is a
valid mathematical theory that deals well with imprecise, vague and
uncertain information, and it has become an area of active research
spreading throughout many fields, such as machine learning, data
mining, knowledge discovery, intelligent data analyzing
\cite{Dai2012UMIDTIA,Herawan2010RSASCA,Herbert2009CCRSM,Lin2013UVPCRSGA,Pedrycz2013GCADIS,Jensen2007RTEA}.
In rough set theory, attribute reduction is a popular task to select
the essential attributes that preserve or improve a certain
classification property as the entire set of available attributes.
Hence, an attribute reduct can be defined as a minimal attribute set
that can preserve or improve specific classification criterion of a
given information system. Attribute reduction is often helpful to
reduce the computational cost, save storage space, improve learning
performance and prevent over-fitting \cite{Thangavela2009DRRSTR}.
Studies on attribute reduction in rough set theory can be mainly
divided into two categories. The first category concentrates on the
study of definition of attribute reduct. The second category focuses
on the study of attribute reduction algorithm.

For definition of attribute reduct, one mainly emphasizes on
selecting what kinds of properties of a given information system to
keep unchanged or to improve
\cite{Jia2015GARRST,Min2011TCSAR,Min2012ARDERTC}. For example,
Pawlak \cite{Pawlak1991RSTARD} defined a relative reduct that keeps
the quality of classification or classification positive region
unchanged. Miao et al. \cite{Miao1999HARK} constructed the mutual
information based reducts to extract relevant attribute sets that
preserve the mutual information of a given decision table.
Kryszkiewicz
\cite{Kryszkiewicz2001CSATKRIS,Kryszkiewicz2007CGDMDRVFDIS}
investigated and compared the relationships among possible reduct,
approximate reduct, $\mu$-reduct, $\mu$-decision reduct and
generalized decision reduct. Zhang et al. \cite{Zhang2003AKRIS}
introduced the maximum distribution reduct, which preserves all
maximum decision rules in a decision table. Deng et al.
\cite{Deng2011FSDSBCKG} proposed the notions of conditional
knowledge granularity to reflect the relationship between
conditional attributes and decision attribute, and defined an
attribute reduct based on conditional knowledge granularity. Jiang
et al. \cite{Jiang2015RDEBFSA} proposed a new model of relative
decision entropy by combining roughness with the degree of
dependency, and used it as the reduction criterion.

For attribute reduction algorithm, one mainly focuses on designing
the effective reduct construction methods for finding a specific
type of reduct
\cite{Shu2015IAARDIDSRST,Teng2016EARVD,Zheng2014EHARARS}. According
to the different methods, research efforts on attribute reduction
algorithm can be divided into three categories. The first class of
methods is based on discernibility matrix. Miao et al.
\cite{Miao2009RRCIDTPRSM} discussed the structures of discernibility
matrices for three different reducts, including region preservation
reduct, decision preservation reduct and relationship preservation
reduct. Yao and Zhao \cite{Yao2009DMSCAR} introduced a reduct
construction method based on discernibility matrix simplification.
The second class of methods is based on heuristics. Qian et al.
\cite{Qian2010PAAARRST} proposed the concept of positive
approximation for accelerating heuristic attribute reduction
algorithms. Parthalain et al.
\cite{Parthalain2009EBRTRSFS,Parthalain2010DMAERSBRAR} proposed a
distance measure-based attribute reduction algorithm by considering
the proximity of objects in the boundary region to those in the
lower approximation. The third class of methods is based on
stochastic optimization. Chen et al. \cite{Chen2010RSAFSBACO}
proposed a novel attribute reduction algorithm based on ant colony
optimization for finding a reduct that keeps the mutual information
unchanged. Ye et al. \cite{Ye2013NBFERSMARP} introduced a novel
fitness function for designing the particle swarm optimization-based
and genetic-based attribute reduction algorithms.

In particular, the studies mentioned above are mainly focused on
attribute reduction in classical rough set model. Relative to the
classical rough set model, probabilistic rough set model allows
certain acceptable levels of errors by using the threshold
parameters
\cite{Dou2015DTRSMS,Liu2012PMCDTRS,Ma2012PRSOTURE,Wang2015GVPFRSGFR}.
Hence, the probabilistic rough set model can effectively deal with
data sets which have the noisy and uncertain data. By setting the
different threshold parameters, one can derive many existing
probabilistic rough set models, such as 0.5 probabilistic rough set
model \cite{Pawlak1988RSPVDA}, decision-theoretic rough set model
\cite{Yao1990ADTRS}, variable precision rough set model
\cite{Ziarko1993VPRS}, Bayesian rough set model
\cite{Slezak2005IBRS} and game rough set model
\cite{Herbert2011GTRS}. Although attribute reduction in
probabilistic rough set model has gained considerable attention in
recently, these works generally focus on the study of definition of
attribute reduct
\cite{Inuiguch2005SAARVPRSM,Jia2013MCARDTRSM,Li2011NMARDTRS,Mi2004AKRVPRS,Wang2009RRFVPRS,Yao2008ARDTRS,Zhang2014RBQHARTCDTRSM,Zhao2011NARDTRS,Ziarko1993VPRS}.
For attribute reduction algorithm, the research efforts mainly
concentrate on the discernibility matrix based methods
\cite{Inuiguch2005SAARVPRSM,Mi2004AKRVPRS,Yang2014NAARVPRSM,Zhao2011NARDTRS}
and stochastic optimization based methods
\cite{Jia2014OORDTRSM,Liu2014SCBAFRVPRS,Chebrolu2011ARDTRSMUGA}. Few
attempts have been made to study the heuristic attribute reduction
algorithms in probabilistic rough set model. This is partly because
the attribute reduction in probabilistic rough set model becomes
more complex after introducing the threshold parameters.

This paper concentrates on constructing heuristic attribute
reduction algorithms for finding low and upper distribution reducts
in probabilistic rough set model, which are defined by Mi et al. in
\cite{Mi2004AKRVPRS}. Generally speaking, the most common heuristic
attribute reduction algorithms are the addition-deletion method and
the deletion method \cite{Yao2008RCA}. The addition-deletion method
starts from an empty set or the core, then it adds attributes one by
one on the basis of the significance of attributes until a stopping
criterion is reached. On the contrary, the deletion method starts
with a set containing all the attributes, then it delete the
attributes one by one according to the significance of attributes.
In fact, the heuristic attribute reduction algorithms mainly
included two aspects: the stopping criterion and the significance
measures of attributes. The stopping criterion is implemented by
checking the jointly sufficient condition in the definition of
attribute reduct, and the significance measures of attributes is
used to rank the attributes. As we known, the fitness functions play
a non-trivial role in designing stopping criteria and constructing
significance measures of attributes. Furthermore, the monotonicity
of fitness functions is very important to guarantee the validity of
heuristic attribute reduction algorithms. However, very little work
have considered the monotonicity of fitness functions in attribute
reduction of probabilistic rough set model. In this paper, we first
construct two monotonic fitness functions in probabilistic rough set
model. The equivalent definitions of the low and upper distribution
reducts are obtained based on the constructed monotonic fitness
functions. Then we use them to design the stopping criterion of
heuristic attribute reduction algorithms. The monotonicity of
fitness functions guarantees the validity of heuristic attribute
reduction algorithms. In addition, we propose two modified monotonic
fitness functions to evaluate the significance of attributes more
effectively. On this basis, we develop two heuristic attribute
reduction algorithms for finding the low and upper distribution
reducts based on addition-deletion method and deletion method. In
the end, some experimental analyses are covered to validate the
effectiveness of the proposed fitness functions and distribution
reducts.

The remaining sections of this paper are organized as follows.
Section \ref{Preliminaries} briefly reviews some basic notions
related to the definition of distribution reducts and heuristic
attribute reduction algorithms. Section \ref{HeuristicAlgorithm}
develops two heuristic attribute reduction algorithms for finding
the low and upper distribution reducts in probabilistic rough set
model by constructing the monotonic fitness functions. Several
experiments are given to illustrate the effectiveness of the
constructed fitness functions and distribution reducts in Section
\ref{experimental}. Section \ref{conclusion} concludes and offers
suggestions for future research.

\section{Preliminary knowledge} \label{Preliminaries}
In this section, we recall the basic notions related to attribute
reduction on rough set theory. \cite{Liu2012MCDTRS,Yao1990ADTRS}.

\subsection{The definition of distribution reducts}
An information system is a four-tuple $IS = (U,A,V,f)$, where $U$ is
a finite nonempty set of objects called universe, $A$ is a nonempty
finite set of attributes, $V = \bigcup\nolimits_{a \in A} {{V_a}} $,
where ${V_a}$ is a nonempty set of values of attribute $a \in A$,
called the domain of $a$, $f:U \times A \to V$ is a mapping that
maps an object in $U$ to exactly one value in ${V_a}$ such that
$\forall a \in A$, $x \in U$, $f(x,a) \in {V_a}$. For brevity, $IS =
(U,A,V,f)$ can be written as $IS = (U,A)$.

For any subset of attributes $R \subseteq A$, an indiscernibility
relation $IND(R)$ on $U$ is defined as:
$$IND(R) = \{ (x,y) \in U \times U|\forall a \in R,f(x,a) =
f(y,a)\}.
$$

It can be easily shown that $IND(R)$ is an equivalence relation on
$U$. For $R \subseteq A$, the equivalence relation $IND(R)$
partitions $U$ into some equivalence classes denoted by $U/IND(R) =
\{ {[x]_R}|u \in U\} $, for simplicity, $U/IND(R)$ will be replaced
by $U/R$, where ${[x]_R}$ is an equivalence class determined by $x$
with respect to $R$, i.e., ${[x]_R} = \{ y \in U|(x,y) \in IND(R)\}
$.

To describe a concept, rough set theory introduces a pair of lower
approximation and upper approximation as follows.

\begin{dfn}
\label{Approximations} Given an information system $IS = (U,A)$, $R
\subseteq A$ and $X \subseteq U$, the lower approximation and upper
approximation of $X$ with respect to $R$ are defined as:
\begin{eqnarray*}
{\underline {apr} _R}(X) &=& \{ x \in U|{[x]_R} \subseteq X\},\\
{\overline {apr} _R}(X) &=& \{ x \in U|{[x]_R} \cap X \ne \emptyset
\}.
\end{eqnarray*}
\end{dfn}

The lower approximation of $X$ is a set of objects that belong to
$X$ with certainty, and the upper approximation of $X$ is a set of
objects that possibly belong to $X$.

Given an information system $IS = (U,A)$, $P,Q \subseteq A$, one can
define a partial relation $\underline  \prec  $ on ${2^A}$ as
follows \cite{Dai2012AUMIIS}:
$$P\underline  \prec  Q \Leftrightarrow \forall x \in U,{[x]_P} \subseteq {[x]_Q}.$$
If $P\underline  \prec  Q $, Q is said to be coarser than P (or P is
finer than Q). If $P\underline \prec  Q $ and $P \ne Q$, Q is said
to be strictly coarser than P or P is strictly finer than Q, denoted
by $P \prec Q$. In fact, $P \prec Q \Leftrightarrow \forall x \in
U$, we have that ${[x]_P} \subseteq {[x]_Q}$ and there exists $y \in
U$ such that ${[y]_P} \subset {[y]_Q}$.

A decision table is a four-tuple $DT = (U,C \cup D,V,f)$, where $C$
is condition attribute set, $D$ is decision attribute set, and $C
\cap D = \emptyset $, $V$ is the union of attribute domain, $V =
{V_C} \cup {V_D} =  \{ {V_a}|a \in C\}  \cup \{ {V_d}|d \in D\} $.
$DT = (U,C \cup D,V,f)$ can be written as $DT = (U,C \cup D)$ more
simply.

\begin{dfn}
\label{PositiveRegion} {\rm{\cite{Liang2004IEREKGRST}}} Given a
decision table $DT = (U,C \cup D)$, $R \subseteq C$ and $U/D = \{
{Y_1},{Y_2}, \ldots ,{Y_M}\} $ is a classification of the universe
$U$. The positive region of $U/D$ with respect to $R$ is defined as
follows:
\begin{equation*}
PO{S_R}(D) = \bigcup\limits_{1 \le i \le M} {{{\underline {apr}
}_R}({Y_i})}.
\end{equation*}
\end{dfn}

Pawlak rough set model does not allow any tolerance of errors. The
probabilistic rough set model, which is a main extension of Pawlak
rough set model, shows certain levels of tolerance for errors.

\begin{dfn}
\label{ProbabilisticApproximations} Given an information system $IS
= (U,A)$, for any $0 \le \beta  < \alpha  \le 1$, $R \subseteq A$
and $X \subseteq U$. The probabilistic lower approximation and
probabilistic upper approximation of $X$ with respect to $R$ are
defined as follows:
\begin{eqnarray*}
\underline {apr} _R^{(\alpha ,\beta )}(X) = \{ x \in U|p(X|{[x]_R}) \ge \alpha \}, \\
\overline {apr} _R^{(\alpha ,\beta )}(X) = \{ x \in U|p(X|{[x]_R})
> \beta \},
\end{eqnarray*}
\end{dfn}
\noindent where $p(X|{[x]_R}) = {{|{{[x]}_R} \cap X|} \over
{|{{[x]}_R}|}}$.

\begin{dfn}
\label{ProbabilisticPositiveRegion} {\rm{\cite{Liang2004IEREKGRST}}}
Given a decision table $DT = (U,C \cup D)$, for any $0 \le \beta  <
\alpha \le 1$, $R \subseteq C$ and $U/D = \{ {Y_1},{Y_2}, \ldots
,{Y_M}\} $ is a classification of the universe $U$. The
probabilistic positive region of $U/D$ with respect to $R$ is
defined as follows:
\begin{equation*}
POS_R^{(\alpha ,\beta )}(D) = \bigcup\limits_{1 \le i \le M}
{\underline {apr} _R^{(\alpha ,\beta )}({Y_i})}.
\end{equation*}
\end{dfn}

Attribute reduction is one of the most important topics in rough set
theory. An attribute reduct is defined as a subset of attributes
that are jointly sufficient and individually necessary for
preserving or improving a particular property of the given
information system \cite{Yao2008ARDTRS}. A general definition of an
attribute reduct is given as follows.

\begin{dfn}\label{GeneralDefinitionOfAttributeReduct} \cite{Zhao2007GDAR}
Given an information system $IS = (U,A)$, $R \subseteq A$ and
consider a certain property $\mathbb{P}$, which can be represented
by an evaluation function $e:{2^A} \to (L,\underline  \prec  )$, of
$IS$. An attribute set $R$ is called a reduct of $IS$ if it
satisfies the following two conditions:
\begin{enumerate}\upshape
\renewcommand{\labelenumi}{\rm{(\theenumi)}}
\item Jointly sufficient condition: $e(A)\underline  \prec  e(R)$,
\item Individually necessary condition: for any $R' \subset R$, $\neg (e(A)\underline  \prec  e(R))$.
\end{enumerate}
\end{dfn}

An evaluation or fitness function, $e:{2^A} \to (L,\underline  \prec
)$, maps an attribute set to an element of a poset $L$ equipped with
the partial order relation $\underline \prec$ i.e., $\underline
\prec$ is reflexive, anti-symmetric and transitive. For a certain
property $\mathbb{P}$, various fitness functions can be used to
evaluate the degree of satisfiability of the property by an
attribute set. Generally, the fitness function is not unique. The
jointly sufficient condition guarantees that the evaluation of the
reduct $R$ with respect to $e$ is the same or superior to $e(A)$,
and it has $e(R) = e(A)$ in many cases. The individually necessary
condition guarantees that the reduct is minimal, namely, there is no
redundant or superfluous attribute in the reduct $R$.

According to the different properties $\mathbb{P}$ of an information
system, the different reducts can be defined. There are a huge
amount of known properties, such as a description of an object
relation \cite{Pawlak1982RS}, partitions of an information system
\cite{Zhao2007GDAR}, a classification of a set of concepts
\cite{Pawlak1991RSTARD}, where the classification of a set of
concepts is the most common property in rough set theory.

In classical rough set model, the classification of a set of
concepts can be evaluated by using the positive region of the
classification (Definition \ref{PositiveRegion}). In probabilistic
rough set model, the classification of a set of concepts can be
evaluated by using the probabilistic positive region of the
classification (Definition \ref{ProbabilisticPositiveRegion}).
However, the decision rules derived from the reduct preserving
probabilistic positive region maybe in conflict with those derived
from the original decision table because of non-monotonicity of
probabilistic positive region with respect to the set inclusion of
attribute sets \cite{Mi2004AKRVPRS}.

To derive the conflict free decision rules, Mi et al.
\cite{Mi2004AKRVPRS} presented the concepts of distribution reducts
based on variable precision rough set model which is a typical
probabilistic rough set model.

\begin{dfn}
Given a decision table $DT = (U,C \cup D)$, for any $0 \le \beta <
\alpha \le 1$, $R \subseteq C$ and $U/D = \{ {Y_1},{Y_2}, \ldots
,{Y_M}\} $ is a classification of the universe $U$. The $(\alpha,
\beta)$ lower and upper approximation distribution functions with
respect to $R$ are defined as:
\begin{eqnarray*}
\underline {apr} _R^{(\alpha ,{\rm{ }}\beta )} &=& {\rm{
}}(\underline {apr} _R^{(\alpha ,{\rm{ }}\beta )}({Y_1}),\underline
{apr} _R^{(\alpha ,{\rm{ }}\beta
)}({Y_2}),\cdot\cdot\cdot,\underline {apr} _R^{(\alpha ,{\rm{
}}\beta )}({Y_M})), \\
\overline {apr} _R^{(\alpha ,{\rm{ }}\beta )} &=& (\overline {apr}
_R^{(\alpha ,{\rm{ }}\beta )}({Y_1}),\overline {apr} _R^{(\alpha
,{\rm{ }}\beta )}({Y_2}),\cdot\cdot\cdot,\overline {apr} _R^{(\alpha
,{\rm{ }}\beta )}({Y_M})).
\end{eqnarray*}
\end{dfn}

The $(\alpha, \beta)$ lower or upper approximation distribution
functions can be seen as fitness functions for evaluating the
classification of a set of concepts in probabilistic rough set
model. By the $(\alpha, \beta)$ lower and upper approximation
distribution functions, the $(\alpha, \beta)$ lower and upper
approximation distribution reducts based on probabilistic rough set
model are defined as follows.

\begin{dfn} \cite{Mi2004AKRVPRS}
\label{DistributeReduct1} Given a decision table $DT = (U,C \cup
D)$, for any $0 \le \beta < \alpha  \le 1$ and $R \subseteq C$, we
have
\begin{enumerate}\upshape
\renewcommand{\labelenumi}{\rm{(\theenumi)}}
\item If $\underline {apr} _R^{(\alpha ,{\rm{ }}\beta )} = \underline {apr} _C^{(\alpha ,{\rm{ }}\beta )}$, $R$ is referred to as an $(\alpha
,\beta )$ lower distribution consistent set of $DT$; if $\underline
{apr} _R^{(\alpha ,{\rm{ }}\beta )} = \underline {apr} _C^{(\alpha
,{\rm{ }}\beta )}$ and $\underline {apr} _{R'}^{(\alpha ,\beta )}
\ne \underline {apr} _C^{(\alpha ,\beta )}$ for all ${R'} \subset
R$, then $R$ is referred to as an $(\alpha ,\beta )$ lower
distribution reduct of $DT$.
\item If $\overline {apr} _R^{(\alpha ,{\rm{ }}\beta )} = \overline {apr} _C^{(\alpha ,{\rm{ }}\beta )}$, $R$ is referred to as an $(\alpha
,\beta )$ upper distribution consistent set of $DT$; if $\overline
{apr} _R^{(\alpha ,{\rm{ }}\beta )} = \overline {apr} _C^{(\alpha
,{\rm{ }}\beta )}$ and $\overline {apr} _{R'}^{(\alpha ,\beta )} \ne
\overline {apr} _C^{(\alpha ,\beta )}$ for all ${R'}$$\subset$$R$,
then $R$ is referred to as an $(\alpha ,\beta )$ upper distribution
reduct of $DT$.
\end{enumerate}
\end{dfn}

An $(\alpha ,\beta )$ lower (upper) distribution reduct is a minimal
subset of attribute set that preserves the $(\alpha ,\beta )$ lower
(upper) approximations of all decision classes. For the sake of the
simplicity, the $(\alpha ,\beta )$ lower and upper distribution
reducts are collectively called distribution reducts in the rest of
this paper.

It is important to note that the monotonicity property of the
fitness function used in the definition of attribute reduct with
respect to the set inclusion of attribute sets should receive enough
attention when we design the heuristic attribute reduction
algorithms. If the fitness function is monotonic regarding the set
inclusion of attribute sets, individually necessary condition only
need to consider the subsets $R - \{a\}$ for all $a \in R$ to
guarantee a reduct $R$ is minimal. If the fitness function is not
monotonic regarding the set inclusion of attribute sets,
individually necessary condition must consider all subsets of a
reduct $R$ to make sure it is a minimal set
\cite{Jia2013MCARDTRSM,Zhao2011NARDTRS}.

In Definition \ref{DistributeReduct1}, individually necessary
conditions must consider all subsets of the reduct $R$ because the
$(\alpha, \beta)$ lower and upper approximation distribution
functions are not monotonic regarding the set inclusion of attribute
sets, which will complicate the algorithm design.

\subsection{Typical heuristic attribute reduction algorithms}
So far, Yao et al. \cite{Yao2008RCA} have summarized three groups of
heuristic attribute reduction algorithms based on the
addition-deletion method, the deletion method and the addition
method, where the addition-deletion method based algorithm and the
deletion method based algorithm are two most widely used heuristic
attribute reduction algorithms by the rough set community. Hence, we
mainly discuss the first two methods based heuristic attribute
reduction algorithms in this subsection.

The addition-deletion method based algorithm and the deletion method
based algorithm are displayed in Algorithm \ref{addition-deletion}
and Algorithm \ref{deletion}, respectively.

\begin{algorithm}[tb!]
\caption{The addition-deletion method for computing a
reduct}\label{addition-deletion}
  \textbf{Input}: A decision table $DT = (U, C
\cup D)$, threshold $(\alpha, \beta)$\\
  \textbf{Output}: A reduct R \\
\textbf{Method}: Addition-deletion method\\

  \begin{algorithmic}[1]
    \STATE // Addition
    \STATE Let $R = \emptyset $, $CA = C$
         \WHILE {$R$ is not jointly sufficient and $CA \ne \emptyset $}
        \STATE Compute fitness values of all the attributes in $CA$
        regarding the property $\mathbb{P}$ using a fitness function $\sigma $
        \STATE Select an attribute $c$ according to its fitness, let $CA = CA - \{ c\} $
        \STATE Let $R = R \cup \{ c\} $
     \ENDWHILE
    \STATE // Deletion
    \STATE Let $CD = R$
     \WHILE {$CD \ne \emptyset $}
        \STATE Compute fitness values of all the attributes in $CD$
        regarding the property $\mathbb{P}$ using a fitness function $\delta $
        \STATE Select an attribute $a$ according to its fitness, let $CD = CD - \{ a\} $
        \IF{$R - \{ a\}$ is jointly sufficient}
            \STATE $R = R - \{a \}$
        \ENDIF
     \ENDWHILE
     \STATE Return $R$
  \end{algorithmic}
\end{algorithm}

\begin{algorithm}[tb!]
\caption{The deletion method for computing a reduct}\label{deletion}
  \textbf{Input}: A decision table $DT = (U, C
\cup D)$, threshold $(\alpha, \beta)$\\
  \textbf{Output}: A reduct $R$ \\
\textbf{Method}: Deletion method\\

  \begin{algorithmic}[1]
   \STATE // Deletion
    \STATE Let $R = C$, $CD = C$
     \WHILE {$CD \ne \emptyset $}
        \STATE Compute fitness values of all the attributes in $CD$
        regarding the property $\mathbb{P}$ using a fitness function $\delta $
        \STATE Select an attribute $a$ according to its fitness, let $CD = CD - \{ a\} $
        \IF{$R - \{ a\}$ is jointly sufficient}\label{jointlySufficient1}
            \STATE $R = R - \{a \}$ \label{jointlySufficient2}
        \ENDIF
     \ENDWHILE
     \STATE Return $R$
  \end{algorithmic}
\end{algorithm}

Algorithm \ref{addition-deletion} starts with an empty set or the
core, and consequently adds attributes to the subset of selected
attributes until a candidate reduct satisfies the jointly sufficient
condition in the definition of attribute reduct. Each selected
attribute maximizes the increment of fitness values of the current
attribute subset. One needs the deleting process to delete the
superfluous attributes in the candidate reduct one by one after the
addition process because the addition process may add the
superfluous attributes.

Algorithm \ref{deletion} takes the entire condition attributes as a
candidate reduct, then selects the attributes for deleting one by
one according to the fitness values. If the subset of the remaining
attributes satisfies the jointly sufficient condition in the
definition of attribute reduct after deleting the selected
attribute, then the attribute is the superfluous attribute and can
be deleted. A reduct is obtained if and only if each attribute has
been checked once.

Algorithm \ref{addition-deletion} and Algorithm \ref{deletion}
mainly consist of two key steps: checking the jointly sufficient
condition and evaluating the significance of attributes. Checking
the jointly sufficient condition is to ensure that the candidate
reduct $R$ meet jointly sufficient condition in the definition of
attribute reduct. Evaluating the significance of attributes is to
sort attributes and provide the heuristic information for searching
a reduct, and the step can be implemented by designing the effective
fitness functions. Different fitness functions may get the different
orders of attributes, that may obtain the different reducts.

It is worth pointing out that the monotonicity of the fitness
function in Definition \ref{GeneralDefinitionOfAttributeReduct} with
respect to the set inclusion of attribute sets is very important for
the completeness of Algorithm \ref{addition-deletion} and Algorithm
\ref{deletion} \cite{Ma2014DRDPRDTRS}. If the fitness function is
monotonic regarding the set inclusion of attribute sets, Algorithm
\ref{addition-deletion} and Algorithm \ref{deletion} must obtain a
reduct, namely, Algorithm \ref{addition-deletion} and Algorithm
\ref{deletion} are complete. If the fitness function is not
monotonic regarding the set inclusion of attribute sets, Algorithm
\ref{addition-deletion} and Algorithm \ref{deletion} may obtain a
super reduct that includes redundancy attributes, namely, Algorithm
\ref{addition-deletion} and Algorithm \ref{deletion} are incomplete.
Moreover, the fitness functions for evaluating the significance of
attributes should also satisfy the monotonicity with respect to the
set inclusion of attribute sets and provide enough precision to sort
the attributes more effectively
\cite{Jiang2015RDEBFSA,Li2015CIMACDT}.

\section{Heuristic algorithms to find distribution reducts in probabilistic rough set model} \label{HeuristicAlgorithm}
Checking the jointly sufficient condition and evaluating the
significance of attributes are two key steps in heuristic attribute
reduction algorithms based on the addition-deletion strategy and the
deletion strategy. Furthermore, the monotonicity of the fitness
functions for checking the jointly sufficient condition and
evaluating the significance of attributes is very important for the
validity of the heuristic attribute reduction algorithms. To obtain
the distribution reducts with heuristic attribute reduction
algorithms, in this section, we first construct two monotonic
fitness functions. Then we give the equivalent definition of
distribution reducts based on the monotonic fitness functions
constructed. After that, we further proposed two significance
measures of attributes by multiplying measures of granularity of
partitions by the monotonic fitness functions constructed. Moreover,
the core and core computation algorithm for distribution reducts are
also presented. On this basis, two heuristic attribute reduction
algorithms to find distribution reducts are developed based on
addition-deletion method and deletion method. Finally, an
illustrative example to the heuristic attribute reduction algorithms
proposed is provided step by step.

\subsection{The equivalent definition of distribution
reducts with monotonic fitness functions}

For a certain property in the definition of attribution reduct, the
different fitness functions can be used as its indicator. The
monotonicity of the fitness functions is very important to guarantee
the completeness of heuristic attribute reduction algorithms. To
develop the complete heuristic attribute reduction algorithms to
obtain distribution reducts, we construct two monotonic fitness
functions in this subsection. The equivalent definition of
distribution reducts are further given based on the monotonic
fitness functions constructed.

Now, let us first give two monotonic fitness functions as follow.

\begin{dfn} \label{MonotonicityFitnessFunction}
Given a decision table $DT = (U,C \cup D)$, for any $0 \le \beta <
\alpha  \le 1$, $R \subseteq C$ and $U/D = \{ {Y_1},{Y_2}, \ldots
,{Y_M}\} $ is a classification of the universe $U$, we denote
\begin{eqnarray*}
\eta _R^{(\alpha ,\beta )} = {{\sum\nolimits_{{Y_i} \in U/D}
{|{{\underline {apr} }_R}(\underline {apr} _C^{(\alpha ,\beta
)}({Y_i}))|} } \over {|U||U/D|}}, \\
\mu _R^{(\alpha ,\beta )} = {{\sum\nolimits_{{Y_i} \in U/D}
{|{{\overline {apr} }_R}(\overline {apr} _C^{(\alpha ,\beta
)}({Y_i}))|} } \over {|U||U/D|}}.
\end{eqnarray*}

If $R = \emptyset$, then define $\eta _R^{(\alpha ,\beta )} = 0$ and
$\mu _R^{(\alpha ,\beta )} = 1$. Moreover, $\eta _R^{(\alpha ,\beta
)}$ and $\mu _R^{(\alpha ,\beta )}$ are denoted by $\eta$ and $\mu$
respectively if there is no confusion arisen.
\end{dfn}

It is easy to see from above that $\eta _R^{(\alpha ,\beta )}$ and
$\mu _R^{(\alpha ,\beta )}$ are defined by using the lower
approximations of all elements in $\underline {apr} _C^{(\alpha
,\beta )}$ with respect $R$ and upper approximations of all elements
in $\overline {apr} _C^{(\alpha ,\beta )}$ with respect to $R$,
respectively. ${{\underline {apr} }_R}(\underline {apr} _C^{(\alpha
,\beta )}({Y_i}))$ represents the change of certain knowledge with
respect to $\underline {apr} _C^{(\alpha ,\beta )}({Y_i})$ after
removing attributes $C - R$ from the decision table, and
${{\overline {apr} }_R}(\overline {apr} _C^{(\alpha ,\beta
)}({Y_i}))$ represents the change of relevant knowledge with respect
to $\overline {apr} _C^{(\alpha ,\beta )}({Y_i})$ after removing
attributes $C - R$ from the decision table. Therefore, $\eta
_R^{(\alpha ,\beta )}$ represents the change of certain knowledge
with respect to $\underline {apr} _C^{(\alpha ,\beta )}$. $\mu
_R^{(\alpha ,\beta )}$ represents the change of relevant knowledge
with respect to $\overline {apr} _C^{(\alpha ,\beta )}$. Hence, we
can use $\eta _R^{(\alpha ,\beta )}$ and $\mu _R^{(\alpha ,\beta )}$
to compute the $(\alpha ,\beta )$ lower and upper distribution
reducts, respectively.

\begin{thm}
\label{MonotonicityTheorem} Given a decision table $DT = (U,C \cup
D)$, for any $0 \le \beta  < \alpha  \le 1$ and $P,Q \subseteq C$,
we have
\begin{enumerate}
\renewcommand{\labelenumi}{\rm{(\theenumi)}}
\item $P\underline  \prec  Q \Rightarrow \eta _P^{(\alpha ,\beta )} \ge \eta _Q^{(\alpha ,\beta )}$,
\item $P\underline  \prec  Q \Rightarrow \mu _P^{(\alpha ,\beta )} \le \mu _Q^{(\alpha ,\beta )}$.
\end{enumerate}
\end{thm}
\begin{pf}
$(1)$ Suppose $P\underline  \prec  Q$. In terms of the definition of
$\underline  \prec$, we have ${[x]_P} \subseteq {[x]_Q}$. Let $U/D =
\{ {Y_1},{Y_2}, \ldots ,{Y_M}\} $ and ${Y_i} \in U/D, 1 \le i \le
M$.

On the one hand, for $\forall x \in {\underline {apr} _Q}(\underline
{apr} _C^{(\alpha ,\beta )}(Y_i))$, one can obtain ${[x]_Q}
\subseteq \underline {apr} _C^{(\alpha ,\beta )}(Y_i)$, hence
${[x]_P}$$\subseteq$$\underline {apr} _C^{(\alpha ,\beta )}(Y_i)$,
then we obtain that $x$$\in$${\underline {apr} _P}(\underline {apr}
_C^{(\alpha ,\beta )}(Y_i))$, thus ${\underline {apr} _P}(\underline
{apr} _C^{(\alpha ,\beta )}(Y_i)) \supseteq {\underline {apr}
_Q}(\underline {apr} _C^{(\alpha ,\beta )}(Y_i))$.

In the other hand, for $\forall x \in {\overline {apr} _P}(\overline
{apr} _C^{(\alpha ,\beta )}(Y_i))$, we have ${[x]_P} \cap \overline
{apr} _C^{(\alpha ,\beta )}(Y_i) \ne \emptyset $. Since ${[x]_P}
\subseteq {[x]_Q}$, ${[x]_Q} \cap \overline {apr} _C^{(\alpha ,\beta
)}(Y_i) \ne \emptyset $, we have $ x \in {\overline {apr}
_Q}(\overline {apr} _C^{(\alpha ,\beta )}(Y_i))$. It follows that
${\overline {apr} _P}(\overline {apr} _C^{(\alpha ,\beta )}(Y_i))
\subseteq {\overline {apr} _Q}(\overline {apr} _C^{(\alpha ,\beta
)}(Y_i))$.

As a result, we have
$$|{\underline {apr} _P}(\underline {apr} _C^{(\alpha ,\beta )}({Y_i}))| \ge |{\underline {apr} _Q}(\underline {apr} _C^{(\alpha ,\beta )}({Y_i}))|,$$
$$|{\overline {apr} _P}(\overline {apr} _C^{(\alpha ,\beta )}({Y_i}))| \le |{\overline {apr} _Q}(\overline {apr} _C^{(\alpha ,\beta )}({Y_i}))|.$$

Thus,
$$\sum\nolimits_{{Y_i} \in U/D} {|{{\underline {apr}
}_P}(\underline {apr} _C^{(\alpha ,\beta )}({Y_i}))|}  \ge
\sum\nolimits_{{Y_i} \in U/D} {|{{\underline {apr} }_Q}(\underline
{apr} _C^{(\alpha ,\beta )}({Y_i}))|}, $$
$$\sum\nolimits_{{Y_i} \in U/D} {|{{\overline {apr} }_P}(\overline {apr} _C^{(\alpha ,\beta )}({Y_i}))|}  \le \sum\nolimits_{{Y_i} \in U/D} {|{{\overline {apr} }_Q}(\overline {apr} _C^{(\alpha ,\beta )}({Y_i}))|}. $$

Hence,
$$\eta _P^{(\alpha ,\beta )} = {{\sum\nolimits_{{Y_i} \in U/D}
{|{{\underline {apr} }_P}(\underline {apr} _C^{(\alpha ,\beta
)}({Y_i}))|} } \over {|U||U/D|}} \ge \eta _Q^{(\alpha ,\beta )} =
{{\sum\nolimits_{{Y_i} \in U/D} {|{{\underline {apr} }_Q}(\underline
{apr} _C^{(\alpha ,\beta )}({Y_i}))|} } \over {|U||U/D|}},$$
$$\mu _P^{(\alpha ,\beta )} = {{\sum\nolimits_{{Y_i} \in U/D}
{|{{\underline {apr} }_R}(\underline {apr} _C^{(\alpha ,\beta
)}({Y_i}))|} } \over {|U||U/D|}} \le \mu_Q^{(\alpha ,\beta )} =
{{\sum\nolimits_{{Y_i} \in U/D} {|{{\underline {apr} }_R}(\underline
{apr} _C^{(\alpha ,\beta )}({Y_i}))|} } \over {|U||U/D|}}.$$

This completes the proof.

The proof of $(2)$ is similar to that of $(1)$.
\end{pf}

By Theorem \ref{MonotonicityTheorem} we immediately get the
following corollary.

\begin{cor}
\label{MonotonicityCorollary} Given a decision table $DT = (U,C \cup
D)$, for any $0 \le \beta  < \alpha  \le 1$ and $P,Q \subseteq C$,
we have
\begin{enumerate}
\renewcommand{\labelenumi}{\rm{(\theenumi)}}
\item $P \supseteq Q \Rightarrow \eta _P^{(\alpha ,\beta )} \ge \eta _Q^{(\alpha ,\beta )}$,
\item $P \supseteq Q \Rightarrow \mu _P^{(\alpha ,\beta )} \le \mu _Q^{(\alpha ,\beta )}$.
\end{enumerate}
\end{cor}

Theorem \ref{MonotonicityTheorem} and Corollary
\ref{MonotonicityCorollary} show that the fitness function $\eta
_R^{(\alpha ,\beta )}$ increases and the fitness function $\mu
_R^{(\alpha ,\beta )}$ decreases as the equivalence classes become
smaller through finer partitioning, which means that adding a new
attribute into the existing subset of condition attributes at least
does not decrease $\eta _R^{(\alpha ,\beta )}$ or increase $\mu
_R^{(\alpha ,\beta )}$, and that deleting an attribute from the
existing subset of condition attributes at least does not increases
$\eta _R^{(\alpha ,\beta )}$ or decreases $\mu _R^{(\alpha ,\beta
)}$. The property is very important for constructing heuristic
attribute reduction algorithms.

In the following, the performance of Theorem
\ref{MonotonicityTheorem} is shown through an illustrative example.

\begin{exm}
\label{MonotonicityExample} Given a decision table $DT = (U,C \cup
D)$ showed in Table \ref{DecisionSystem1}, where $U = \{
{x_1},{x_2}, \cdots,{x_{12}}\} $, and $C = \{
{a_1},{a_2},{a_3},{a_4},{a_5},{a_6}\}$. Suppose that $\alpha  =
0.60$ and $\beta  = 0.40$, $P,Q \subseteq C$, where $P = \{
{a_1},{a_2},{a_3}\} $ and $Q = \{ {a_1},{a_2}\} $. As we can see, $P
\supseteq Q$, which means $P\underline  \prec  Q$.

\setlength{\tabcolsep}{14pt}
\begin{table}[htbp] \small\vspace{-0.3cm}
\caption{A decision table} \label{DecisionSystem1}
\begin{center}
\begin{tabular}{cccccccc}
\hline
$U$ & ${a_1}$ & ${a_2}$ & ${a_3}$ & ${a_4}$ & ${a_5}$ & ${a_6}$ & $d$ \\
\hline
$x_{1}$& 1 & 1 & 1 & 1 & 0 & 0 & 0\\
$x_{2}$& 0 & 1 & 1 & 0 & 1 & 0 & 0\\
$x_{3}$& 0 & 1 & 1 & 1 & 0 & 0 & 0 \\
$x_{4}$& 0 & 1 & 1 & 1 & 0 & 0 & 1 \\
$x_{5}$& 0 & 0 & 0 & 1 & 0 & 1 & 0 \\
$x_{6}$& 0 & 0 & 1 & 1 & 1 & 1 & 1 \\
$x_{7}$& 0 & 0 & 1 & 1 & 0 & 1 & 1 \\
$x_{8}$& 1 & 1 & 0 & 0 & 1 & 1 & 0 \\
$x_{9}$& 1 & 1 & 0 & 0 & 1 & 1 & 1 \\
$x_{10}$& 1 & 1 & 0 & 0 & 0 & 0 & 1 \\
$x_{11}$& 1 & 1 & 0 & 0 & 0 & 0 & 1 \\
\hline
\end{tabular}
\end{center}
\end{table}

By calculating, one can have

\begin{eqnarray*}
U/C &=& \{ \{{x_1}\}, \{{x_2}\}, \{{x_3},{x_4}\},\{{x_5}\}, \{{x_6}\}, \{{x_7}\},\{{x_8}, {x_9}\}, \{{x_{10}},{x_{11}}\}\}, \\
U/D &=& \{
\{{x_1},{x_2},{x_3},{x_5},{x_{8}}\},\{x_{4},{x_6},{x_7},{x_9},x_{10},{x_{11}}\}
\}.
\end{eqnarray*}

Hence, we have
\begin{eqnarray*}
\underline {apr} _C^{(0.60 , 0.40)} &=& ( \{{x_1},{{x_2}},{x_{5}}\}, \{{x_6},{x_7},{x_{10}},{x_{11}}\} ), \\
\overline {apr} _C^{(0.60 , 0.40)} &=& (
\{{x_1},{x_2},{x_3},{x_4},x_{5},x_{8},{x_{9}}\},\{{x_3},{x_4},{x_6},x_{7},x_{8},x_{9},x_{10},x_{11}\}
).
\end{eqnarray*}

It can be easily calculated that
\begin{eqnarray*}
U/P &=& \{ \{{x_1}\}, \{{x_2},{x_3},{x_4}\},\{{x_5}\}, \{{x_6},{x_7}\},\{{x_8}, {x_9},{x_{10}},{x_{11}}\}\}, \\
U/Q &=& \{ \{{x_1}\},
\{{x_2},{x_3},{x_4}\},\{{x_5},{x_6},{x_7}\},\{{x_8},
{x_9},{x_{10}},{x_{11}}\}\}.
\end{eqnarray*}

Obviously, $P \prec  Q$.

According to Definition \ref{MonotonicityFitnessFunction}, we have
\begin{eqnarray*}
\eta _P^{(0.60 , 0.40)} = {{\sum\nolimits_{{Y_i} \in U/D}
{|{{\underline {apr} }_P}(\underline {apr} _C^{(0.60, 0.40
)}({Y_i}))|} } \over {|U||U/D|}} = {{2 + 2} \over {11 \times 2}} = {{4} \over {22}}, \\
\eta _Q^{(0.60 , 0.40)} = {{\sum\nolimits_{{Y_i} \in U/D}
{|{{\underline {apr} }_Q}(\underline {apr} _C^{(0.60, 0.40
)}({Y_i}))|} } \over {|U||U/D|}} = {{1 + 0} \over {11 \times 2}} = {{1} \over {22}}, \\
\mu _P^{(0.60 , 0.40)} = {{\sum\nolimits_{{Y_i} \in U/D}
{|{{\overline {apr} }_P}(\overline {apr} _C^{(0.60, 0.40
)}({Y_i}))|} } \over {|U||U/D|}} = {{9 + 9} \over {11 \times 2}} =
{{18} \over {22}}, \\
\mu _Q^{(0.60 , 0.40)} = {{\sum\nolimits_{{Y_i} \in U/D}
{|{{\overline {apr} }_Q}(\overline {apr} _C^{(0.60, 0.40
)}({Y_i}))|} } \over {|U||U/D|}} = {{11 + 10} \over {11 \times 2}} =
{{21} \over {22}}.
\end{eqnarray*}

Thus, $\eta _P^{(0.60 , 0.40)} > \eta _Q^{(0.60 , 0.40)}$ and $\mu
_P^{(0.60 , 0.40)} < \mu _Q^{(0.60 , 0.40)}$. It is clear that $\eta
_R^{(\alpha ,\beta)}$ increases and $\mu _R^{(\alpha ,\beta)}$
decreases with $R$ becoming finer.

\end{exm}

Now, let us give equivalence descriptions for each distribution
consistent set by the fitness functions $\eta _R^{(\alpha ,\beta )}$
and $\mu _R^{(\alpha ,\beta )}$.

\begin{thm} \label{EquivalentTheorm}
Given a decision table $DT = (U,C \cup D)$, for any $0 \le \beta <
\alpha  \le 1$ and $R \subseteq C$, we have
\begin{enumerate}\upshape
\renewcommand{\labelenumi}{\rm{(\theenumi)}}
\item $R$ is an $(\alpha ,\beta )$ lower distribution consistent set of $DT$ iff $\eta _R^{(\alpha ,\beta )} = \eta _C^{(\alpha ,\beta
)}$,
\item $R$ is an $(\alpha ,\beta )$ upper distribution consistent set of $DT$ iff $\mu _R^{(\alpha ,\beta )} = \mu _C^{(\alpha ,\beta )}$.
\end{enumerate}
\end{thm}

\begin{pf}
Since $R \subseteq C$, it is easy to verify that $\xi ({[x]_R}) = \{
{[y]_C}:{[y]_C} \subseteq {[x]_R}\}$ forms a partition of $[x]_R$.
Let $U/D = \{ {Y_1},{Y_2}, \ldots ,{Y_M}\} $.

$(1)$ $`` \Rightarrow "$ Since $R$ is an $(\alpha ,\beta )$ lower
distribution consistent set, we have $\underline {apr} _R^{(\alpha
,{\rm{ }}\beta )} = \underline {apr} _C^{(\alpha ,{\rm{ }}\beta )}$,
therefore for $\forall {Y_i} \in U/D$, we obtain that $\underline
{apr} _R^{(\alpha ,\beta )}({Y_i}) = \underline {apr} _C^{(\alpha
,\beta )}({Y_i})$.

Hence,
\begin{eqnarray*}
\eta _R^{(\alpha ,\beta )}{\rm{ }} &=& {{\sum\nolimits_{{Y_i} \in
U/D} {|{{\underline {apr} }_R}(\underline {apr} _C^{(\alpha ,\beta
)}({Y_i}))|} } \over {|U||U/D|}} = {{\sum\nolimits_{{Y_i} \in U/D}
{|{{\underline {apr} }_R}(\underline {apr} _R^{(\alpha ,\beta
)}({Y_i}))|} } \over {|U||U/D|}} \\
&=& {{\sum\nolimits_{{Y_i} \in U/D} {|\underline {apr} _R^{(\alpha
,\beta )}({Y_i})|} } \over {|U||U/D|}} = {{\sum\nolimits_{{Y_i} \in
U/D} {|\underline {apr} _C^{(\alpha ,\beta )}({Y_i})|} } \over
{|U||U/D|}} \\
&=& {{\sum\nolimits_{{Y_i} \in U/D} {|{{\underline {apr}
}_C}(\underline {apr} _C^{(\alpha ,\beta )}({Y_i}))|} } \over
{|U||U/D|}} = \eta _C^{(\alpha ,\beta )}{\rm{ }}.
\end{eqnarray*}

$`` \Leftarrow "$ For $\forall {Y_i} \in U/D$, we have ${\underline
{apr} _R}(\underline {apr} _C^{(\alpha ,\beta )}({Y_i})) \subseteq
\underline {apr} _C^{(\alpha ,\beta )}({Y_i})$.

Suppose there exists ${Y_i} \in U/D$ such that ${\underline {apr}
_R}(\underline {apr} _C^{(\alpha ,\beta )}({Y_i})) \subset
\underline {apr} _C^{(\alpha ,\beta )}({Y_i})$.

As a result,
\begin{eqnarray*}
\eta _R^{(\alpha ,\beta )}{\rm{ }} &=& {{\sum\nolimits_{{Y_i} \in
U/D} {|{{\underline {apr} }_R}(\underline {apr} _C^{(\alpha ,\beta
)}({Y_i}))|} } \over {|U||U/D|}} < {{\sum\nolimits_{{Y_i} \in U/D}
{|\underline {apr} _C^{(\alpha
,\beta )}({Y_i})|} } \over {|U||U/D|}} \\
&=& {{\sum\nolimits_{{Y_i} \in U/D} {|{{\underline {apr}
}_C}(\underline {apr} _C^{(\alpha ,\beta )}({Y_i}))|} } \over
{|U||U/D|}} = \eta _C^{(\alpha ,\beta )}{\rm{ }}.
\end{eqnarray*}

It conflicts with condition $\eta _R^{(\alpha ,\beta )} = \eta
_C^{(\alpha ,\beta )}$. Hence, for $\forall {Y_i} \in U/D$, we have
${\underline {apr} _R}(\underline {apr} _C^{(\alpha ,\beta
)}({Y_i})) = \underline {apr} _C^{(\alpha ,\beta )}({Y_i})$.

If $x \in \underline {apr} _C^{(\alpha ,\beta )}({Y_i})$, there are
two cases in which ${[x]_R} \subseteq \underline {apr} _C^{(\alpha
,\beta )}({Y_i})$ and ${[x]_R} \not\subseteq \underline {apr}
_C^{(\alpha ,\beta )}({Y_i})$.

When ${[x]_R} \not\subseteq \underline {apr} _C^{(\alpha ,\beta
)}({Y_i})$, we have $x \notin {\underline {apr} _R}(\underline {apr}
_C^{(\alpha ,\beta )}({Y_i}))$.

It conflicts with condition  ${\underline {apr} _R}(\underline {apr}
_C^{(\alpha ,\beta )}({Y_i})) = \underline {apr} _C^{(\alpha ,\beta
)}({Y_i})$. Thus ${[x]_R} \subseteq \underline {apr} _C^{(\alpha
,\beta )}({Y_i})$. Since ${[x]_R} = \cup \{ {[y]_C}:{[y]_C} \in \xi
({[x]_R})\} $, we obtain that ${[y]_C} \subseteq {[x]_R} \subseteq
\underline {apr} _C^{(\alpha ,\beta )}({Y_i})$ for all ${[y]_C} \in
\xi ({[x]_R})$. That is to say, for all ${[y]_C} \in \xi ({[x]_R})$,
it has $p({Y_i}|{[y]_C}) \ge {\alpha}$. Therefore we have that
\begin{eqnarray*}
p({Y_i}|{[x]_R}) & = & {{\left( {\sum {\{ |{{[y]}_C} \cap
{Y_i}|:{{[y]}_C} \in \xi ({{[x]}_R})\} } } \right)} \mathord{\left/
 {\vphantom {{\left( {\sum {\{ |{{[y]}_C} \cap {D_j}|:{{[y]}_C} \in \xi ({{[x]}_R})\} } } \right)} {|{{[x]}_R}|}}} \right.
 \kern-\nulldelimiterspace} {|{{[x]}_R}|}} \\
& = & \sum {\left\{ {p({Y_i}|{{[y]}_C}) \cdot {{|{{[y]}_C}|} \over
{|{{[x]}_R}|}}:{{[y]}_C} \in \xi ({{[x]}_R})} \right\}} \\
& \ge & \alpha \sum {\left\{ {{{|{{[y]}_C}|} \over
{|{{[x]}_R}|}}:{{[y]}_C} \in \xi ({{[x]}_R})} \right\}}  = \alpha.
\end{eqnarray*}
As a result $x \in \underline {apr} _R^{(\alpha ,\beta )}({Y_i})$.

On the other hand, if $x \in \underline {apr} _R^{(\alpha ,\beta
)}({Y_i})$, then we have ${[x]_R} \subseteq \underline {apr}
_R^{(\alpha ,\beta )}({Y_i})$. Since ${\underline {apr}
_R}(\underline {apr} _C^{(\alpha ,\beta )}({Y_i})) = \underline
{apr} _C^{(\alpha ,\beta )}({Y_i})$, there are two cases in which
${[x]_R} \subseteq \underline {apr} _C^{(\alpha ,\beta )}({Y_i})$
and ${[x]_R} \cap \underline {apr} _C^{(\alpha ,\beta )}({Y_i}) =
\emptyset $.

When ${[x]_R} \cap \underline {apr} _C^{(\alpha ,\beta )}({Y_i}) =
\emptyset $, since ${[x]_R} =  \cup \{ {[y]_C}:{[y]_C} \in \xi
({[x]_R})\} $, we have ${[y]_C} \cap \underline {apr} _C^{(\alpha
,\beta )}({Y_i}) = \emptyset $ for all ${[y]_C} \in \xi ({[x]_R})$.
That is to say, for all ${[y]_C} \in \xi ({[x]_R})$, it has
$p({Y_i}|{[y]_C}) < {\alpha}$. Therefore we have that
\begin{eqnarray*}
p({Y_i}|{[x]_R}) & = & {{\left( {\sum {\{ |{{[y]}_C} \cap
{Y_i}|:{{[y]}_C} \in \xi ({{[x]}_R})\} } } \right)} \mathord{\left/
 {\vphantom {{\left( {\sum {\{ |{{[y]}_C} \cap {D_j}|:{{[y]}_C} \in \xi ({{[x]}_R})\} } } \right)} {|{{[x]}_R}|}}} \right.
 \kern-\nulldelimiterspace} {|{{[x]}_R}|}} \\
& = & \sum {\left\{ {p({Y_i}|{{[y]}_C}) \cdot {{|{{[y]}_C}|} \over
{|{{[x]}_R}|}}:{{[y]}_C} \in \xi ({{[x]}_R})} \right\}} \\
& <  & \alpha \sum {\left\{ {{{|{{[y]}_C}|} \over
{|{{[x]}_R}|}}:{{[y]}_C} \in \xi ({{[x]}_R})} \right\}}  = \alpha.
\end{eqnarray*}
As a result ${[x]_R} \cap \underline {apr} _R^{(\alpha ,\beta
)}({Y_i}) = \emptyset $, which contradicts with ${[x]_R} \subseteq
\underline {apr} _R^{(\alpha ,\beta )}({Y_i})$.

Hence ${[x]_R} \subseteq \underline {apr} _C^{(\alpha ,\beta
)}({Y_i})$, and in tune $x \in \underline {apr} _C^{(\alpha ,\beta
)}({Y_i})$.

Thus we conclude that $\underline {apr} _R^{(\alpha ,\beta )}({Y_i})
= \underline {apr} _C^{(\alpha ,\beta )}({Y_i})$ for $\forall {Y_i}
\in U/D$, i.e., $R$ is an $(\alpha ,\beta)$ lower distribution
consistent set.

The proof of $(2)$ is similar to that of $(1)$.
\end{pf}

\begin{thm} \label{Dispensable}
Given a decision table $DT = (U,C \cup D)$, for any $0 \le \beta <
\alpha  \le 1$ and $R \subseteq C$, we have
\begin{enumerate}\upshape
\renewcommand{\labelenumi}{\rm{(\theenumi)}}
\item An attribute $a \in R$ is dispensable in $R$ with respect to $\underline {apr} _R^{(\alpha ,{\rm{ }}\beta )}$ iff $\eta _{R - \{ a\} }^{(\alpha ,\beta )} = \eta _R^{(\alpha ,\beta )}$,
\item An attribute $a \in R$ is dispensable in $R$ with respect to $\overline {apr} _R^{(\alpha ,{\rm{ }}\beta )}$ iff $\mu _{R - \{ a\} }^{(\alpha ,\beta )} = \mu _R^{(\alpha ,\beta )}$.
\end{enumerate}
\end{thm}

By Theorem \ref{Dispensable} we immediately get the following
corollary.

\begin{cor}
\label{Indispensable} Given a decision table $DT = (U,C \cup D)$,
for any $0 \le \beta < \alpha  \le 1$ and $R \subseteq C$, we have
\begin{enumerate}\upshape
\renewcommand{\labelenumi}{\rm{(\theenumi)}}
\item An attribute $a \in R$ is indispensable in $R$ with respect to $\underline {apr} _R^{(\alpha ,{\rm{ }}\beta )}$ iff $\eta _{R - \{ a\} }^{(\alpha ,\beta )} \ne \eta _R^{(\alpha ,\beta )}$,
\item An attribute $a \in R$ is indispensable in $R$ with respect to $\overline {apr} _R^{(\alpha ,{\rm{ }}\beta )}$ iff $\mu _{R - \{ a\} }^{(\alpha ,\beta )} \ne \mu _R^{(\alpha ,\beta )}$.
\end{enumerate}
\end{cor}

Theorem \ref{EquivalentTheorm} and Corollary \ref{Indispensable}
yield the following theorem.
\begin{thm}
\label{DistributionReductEquivalentRepresentation} Given a decision
table $DT = (U,C \cup D)$, for any $0 \le \beta < \alpha  \le 1$ and
$R \subseteq C$, we have
\begin{enumerate}\upshape
\renewcommand{\labelenumi}{\rm{(\theenumi)}}
\item $R$ is an $(\alpha ,\beta )$ lower distribution reduct of $DT$ iff \\
(\uppercase\expandafter{\romannumeral1})  $\eta _R^{(\alpha ,\beta
)} = \eta _C^{(\alpha ,\beta
)}$; \\
(\uppercase\expandafter{\romannumeral2}) $\eta _{R - \{ a\}
}^{(\alpha ,\beta )} \ne \eta _C^{(\alpha ,\beta )}$ for $\forall a
\in R$.
\item $R$ is an $(\alpha ,\beta )$ upper distribution reduct of $DT$ iff \\
(\uppercase\expandafter{\romannumeral1}) $\mu _R^{(\alpha ,\beta )} = \mu _C^{(\alpha ,\beta )}$; \\
(\uppercase\expandafter{\romannumeral2}) $\mu _{R - \{ a\}
}^{(\alpha ,\beta )} \ne \mu _C^{(\alpha ,\beta )}$for $\forall a
\in R$.
\end{enumerate}
\end{thm}

Theorem \ref{DistributionReductEquivalentRepresentation} gives the
equivalent definition of distribution reducts.Compared with
Definition \ref{DistributeReduct1}, individually necessary
conditions only need to check the subsets $R - \{a\}$ for all $a \in
R$, not all subsets of $R$ because the fitness functions $\eta
_R^{(\alpha ,\beta )}$ and $\mu _R^{(\alpha ,\beta )}$ are monotonic
with respect to the set inclusion of attribute sets. The condition
simplifies the algorithm design.

In fact, Theorem \ref{DistributionReductEquivalentRepresentation}
provides concrete methods to design heuristic algorithms for
obtaining the distribution reducts, and guarantees the completeness
of heuristic attribute reduction algorithms.

\subsection{The significance measures of attributes}
Evaluating the significance of attributes is one of the most
important problem to design the efficient heuristic attribute
reduction algorithms. In this subsection, we construct the
significance measures of attributes on the basis of the monotonic
fitness functions proposed in order to provide the heuristic
information that can guide search to distribution reducts.

For the significance measures of attributes, the monotonicity of
measures are very important. The monotonicity can guarantee the
rationality of the measures to evaluate the significance of
attributes. Hence, we can use the fitness functions $\eta
_R^{(\alpha ,\beta )}$ and $\mu _R^{(\alpha ,\beta )}$ as the
significance measures of attributes for searching the $(\alpha
,\beta )$ lower and upper distribution reducts to a certain degree,
respectively.

The corresponding significance measures of attributes are defined as
follows.

\begin{dfn}
\label{significanceOut} Given a decision table $DT = (U,C \cup D)$,
for any $0 \le \beta < \alpha \le 1$, $R \subset C$ and $\forall a
\in C - R$, the significance measures of attribute $a$ in $R$ with
respect to $\underline {apr} _C^{(\alpha ,{\rm{ }}\beta )}$ and
$\overline {apr} _C^{(\alpha ,{\rm{ }}\beta )}$ are defined as
follows:
\begin{enumerate}\upshape
\renewcommand{\labelenumi}{\rm{(\theenumi)}}
\item $SI{G_\eta}(a,R,{\underline {apr} _C^{(\alpha ,{\rm{ }}\beta )}}) = \eta_{R \cup \{ a\} }^{(\alpha ,\beta
)} - \eta_R^{(\alpha ,\beta )}$.
\item $SI{G_\mu}(a,R,{\overline {apr} _C^{(\alpha ,{\rm{ }}\beta )}}) = \mu_R^{(\alpha ,\beta )} - \mu_{R \cup \{ a\} }^{(\alpha ,\beta
)}$.
\end{enumerate}
\end{dfn}

Definition \ref{significanceOut} can be used to provide heuristics
to guide the mechanism of searching an attribute.
$SI{G_\eta}(a,R,{\underline {apr} _C^{(\alpha ,{\rm{ }}\beta )}})$
can serve as the heuristic information for searching the $(\alpha
,\beta )$ lower distribution reduct. $SI{G_\mu}(a,R,$ ${\overline
{apr} _C^{(\alpha ,{\rm{ }}\beta )}})$ can serve as the heuristic
information for searching the $(\alpha ,\beta )$ upper distribution
reduct. For convenience, the corresponding significance measures of
attributes are also denoted as $SIG_{\eta}(a,{\underline {apr}
_C^{(\alpha ,{\rm{ }}\beta )}}) = \eta_{\{ a\} }^{(\alpha ,\beta )}$
and $SIG_{\mu}(a,{\overline {apr} _C^{(\alpha ,{\rm{ }}\beta )}}) =
1 - \mu_{\{ a\} }^{(\alpha ,\beta )}$ for any singleton attribute $a
\in C$.

As to the monotonicity of the fitness functions $\eta _R^{(\alpha
,\beta )}$ and $\mu _R^{(\alpha ,\beta )}$, they can be used for
evaluating the significance of attributes. However, the accuracy of
fitness functions is very important to effectively evaluate the
significance of attributes. In some situations, the fitness
functions $\eta _R^{(\alpha ,\beta )}$ and $\mu _R^{(\alpha ,\beta
)}$ cannot supply enough information for evaluating, because they do
not taking into full account the granularity of partitions. The
limitations are revealed by the following example.

\begin{exm}
\label{NonStrictMonotonicityExample} Given a decision table $DT =
(U,C \cup D)$ showed in Table \ref{DecisionSystem2}, where $U = \{
{x_1},{x_2}, \cdots,{x_{12}}\} $, and $C = \{
{a_1},{a_2},{a_3},{a_4},{a_5},{a_6}\}$. Suppose that $\alpha  =
0.60$ and $\beta  = 0.40$, $P,Q \subseteq C$, where $P = \{
{a_1},{a_2},{a_3},{a_4}\} $ and $Q = \{ {a_1},{a_2},{a_3}\} $. As we
can see, $P \supseteq Q$, which means $P\underline  \prec  Q$.

\setlength{\tabcolsep}{14pt}
\begin{table}[htbp] \small\vspace{-0.3cm}
\caption{A decision table} \label{DecisionSystem2}
\begin{center}
\begin{tabular}{cccccccc}
\hline
$U$ & ${a_1}$ & ${a_2}$ & ${a_3}$ & ${a_4}$ & ${a_5}$ & ${a_6}$ & $d$ \\
\hline
$x_{1}$& 1 & 0 & 0 & 1 & 1 & 0 & 0\\
$x_{2}$& 0 & 0 & 1 & 0 & 1 & 1 & 0\\
$x_{3}$& 0 & 1 & 1 & 0 & 1 & 0 & 0 \\
$x_{4}$& 0 & 1 & 1 & 0 & 1 & 0 & 1 \\
$x_{5}$& 0 & 1 & 1 & 0 & 1 & 0 & 1 \\
$x_{6}$& 0 & 1 & 1 & 1 & 0 & 0 & 1 \\
$x_{7}$& 1 & 0 & 0 & 1 & 0 & 1 & 0 \\
$x_{8}$& 1 & 0 & 0 & 1 & 0 & 1 & 1 \\
$x_{9}$& 1 & 0 & 0 & 0 & 1 & 1 & 0 \\
$x_{10}$& 1 & 0 & 0 & 0 & 1 & 0 & 0 \\
$x_{11}$& 1 & 0 & 0 & 0 & 1 & 0 & 1 \\
$x_{12}$& 1 & 0 & 0 & 0 & 1 & 0 & 1 \\
\hline
\end{tabular}
\end{center}
\end{table}

By calculating, one can have

\begin{eqnarray*}
U/C &=& \{ \{{x_1}\}, \{{x_2}\}, \{{x_3},{x_4},{x_5}\}, \{{x_6}\}, \{{x_7},{x_8}\}, \{{x_9}\}, \{{x_{10}},{x_{11}},{x_{12}}\} \}, \\
U/D &=& \{
\{{x_1},{x_2},{x_3},{x_7},{x_{9}},{x_{10}}}\},\{x_{4},{x_5},{x_6},{x_8},{x_{11},{x_{12}}\}
\}.
\end{eqnarray*}

Hence, we have
\begin{eqnarray*}
\underline {apr} _C^{(0.60 , 0.40)} &=& ( \{{x_1}},{{x_2}},{x_{10}},{x_{11}},{x_{12}}\}, \{{x_3},{x_4},{x_5},{x_6},{x_{9}\} ), \\
\overline {apr} _C^{(0.60 , 0.40)} &=& (
\{{x_1},{x_2},{x_7},{x_8},{x_{10}},{x_{11}},{x_{12}}\},\{{x_3},{x_4},{x_5},{x_6},{x_{7},{x_{8},{x_{9}}}}\}
).
\end{eqnarray*}

It can be easily calculated that
\begin{eqnarray*}
U/P &=& \{ \{{x_1},{x_7},{x_8}\}, \{{x_2}\}, \{{x_3},{x_4},{x_5}\}, \{{x_6}\}, \{{x_9},{x_{10}},{x_{11}},{x_{12}}\} \}, \\
U/Q &=& \{
\{{x_1},{x_7},{x_8},{x_9},{x_{10}},{x_{11}},{x_{12}}\},\{x_{2}\},\{x_{3},x_{4},{x_5},{x_6}\}
\}.
\end{eqnarray*}

Obviously, $P \prec  Q$.

According to Definition \ref{MonotonicityFitnessFunction}, we have
\begin{eqnarray*}
\eta _P^{(0.60 , 0.40)} = {{\sum\nolimits_{{Y_i} \in U/D}
{|{{\underline {apr} }_P}(\underline {apr} _C^{(0.60, 0.40
)}({Y_i}))|} } \over {|U||U/D|}} = {{1 + 4} \over {12 \times 2}} = {{5} \over {24}}, \\
\eta _Q^{(0.60 , 0.40)} = {{\sum\nolimits_{{Y_i} \in U/D}
{|{{\underline {apr} }_Q}(\underline {apr} _C^{(0.60, 0.40
)}({Y_i}))|} } \over {|U||U/D|}} = {{1 + 4} \over {12 \times 2}} = {{5} \over {24}}, \\
\mu _P^{(0.60 , 0.40)} = {{\sum\nolimits_{{Y_i} \in U/D}
{|{{\overline {apr} }_P}(\overline {apr} _C^{(0.60, 0.40
)}({Y_i}))|} } \over {|U||U/D|}} = {{8 + 11} \over {12 \times 2}} =
{{19} \over {24}}, \\
\mu _Q^{(0.60 , 0.40)} = {{\sum\nolimits_{{Y_i} \in U/D}
{|{{\overline {apr} }_Q}(\overline {apr} _C^{(0.60, 0.40
)}({Y_i}))|} } \over {|U||U/D|}} = {{8 + 11} \over {12 \times 2}} =
{{19} \over {24}}.
\end{eqnarray*}

As a result, we have

$\eta _P^{(0.60 , 0.40)} = \eta _Q^{(0.60 , 0.40)}$ and $\mu
_P^{(0.60 , 0.40)} = \mu _Q^{(0.60 , 0.40)}$.

\end{exm}

From Example \ref{NonStrictMonotonicityExample}, we can see that
there is a partial relation between $P$ and $Q$. However, we
obtained the same values of the fitness functions. It indicates that
the fitness functions in Definition
\ref{MonotonicityFitnessFunction} can not discern the attribute
subsets $P$ and $Q$ clearly. The main reason is because they do not
take into full account the granularity of partitions. Hence, it is
easy to obtain that the significance of some attributes is zero by
using the fitness functions in Definition
\ref{MonotonicityFitnessFunction}. Therefore, it is necessary to
introduce more effective fitness functions to evaluate the
significance of attributes.

In what follows, we introduce more effective fitness functions based
on the measures of granularity of partitions. We review several
existing measures of granularity of partitions before introducing
the more efficient fitness functions.

Given a partition, there are many methods that can be used to
measure the granularity of a partition. Currently, Yao and Zhao
\cite{Yao2011MTVGP} provided a unified framework for measures of
granularity of partitions by considering the expected granularity of
blocks in a partition.

First, let us recall the measure of granularity of a set.

\begin{dfn} {\rm{\cite{Yao2011MTVGP}}} \label{SetGranularity}
Suppose $U$ is finite and nonempty universe. A function $m:{2^U} \to
\mathbb{R}$ is called a measure of granularity of a set if it
satisfies the following conditions: for all $X,Y \in {2^U}$.
\begin{enumerate}
\renewcommand{\labelenumi}{\rm{(\theenumi)}}
\item $(nonnegativity)$ $m(X) \ge 0$,
\item $(monotonicity)$ $X \subset Y \Rightarrow m(X) < m(Y)$,
\item $(size \ invariant)$ $X{ \equiv _s}Y \Leftrightarrow m(X) =
m(Y)$,
\end{enumerate}
where the binary relation ${ \equiv _s}$ is defined by: $A { \equiv
_s} B$ if there exists a bijection from A to B.
\end{dfn}

The measure of granularity of a partition is defined as follows
based on the measure of granularity of a set.

\begin{dfn} {\rm{\cite{Yao2011MTVGP}}} \label{PartitionGranularity}
Suppose $\pi  = \{ {X_1},{X_2}, \ldots ,{X_K}\} $ is a partition of
a finite nonempty universe $U$ and $m:{2^U} \to \mathbb{R}$ is a
measure of granularity of subsets of $U$, satisfies $(1)-(3)$ of
Definition {\rm{\ref{SetGranularity}}}. The expected granularity of
blocks of $\pi$ is defined as:
$$E{G_m}(\pi ) = {\bm{\mathrm{{E}}}_{{P_\pi }}}(m( \cdot )) = \sum\limits_{i = 1}^K {m({X_i})p({X_i})}, $$
where ${P_\pi } = (p({X_1}),p({X_2}), \ldots ,p({X_K})) = \left(
{{{|{X_1}|} \over {|U|}},{{|{X_2}|} \over {|U|}}, \ldots ,{{|{X_K}|}
\over {|U|}}} \right)$ is the probability distribution defined by
$\pi$ and ${\bm{\mathrm{{E}}}_{{P_\pi }}(\cdot)}$ is the
mathematical expectation with respect to distribution ${P_\pi }$.
\end{dfn}

Let $\Pi $ be the set of all partitions of $U$. The expected
granularity $E{G_m}(\pi )$ reaches the minimum value if and only if
$\pi  = {\Pi _0} = \{ \{ x\} |x \in U\}$ which is the finest
partition in $\Pi$, and it reaches the maximum value if and only if
$\pi  = {\Pi _1} = \{U\}$ which is the coarsest partition in $\Pi$.
In general, we have $E{G_m}({\Pi _0}) \le E{G_m}(\pi ) \le
E{G_m}({\Pi _1})$ for any $\pi  \in \Pi $.

The expected granularity $E{G_m}(\pi )$ is a class of measures of
granularity of partitions. The different measures of granularity of
partitions can be derived from the expected granularity by
considering various classes of measures of granularity of sets.
Hence, many existing measures of granularity of partitions are
instances of the expected granularity. The three most widely used
measures of granularity of partitions are co-entropy, knowledge
granularity and combination granularity in all existing measures of
granularity of partitions. The corresponding definitions are as
follows.

\begin{dfn}
\label{CoEntropy}
{\rm{\cite{Bianucci2009IEGCPCS,Cattaneo2008ECPCART,Liang2004IEREKGRST}}}
Given an information system $IS = (U,A)$, $R \subseteq A$ and $U/R =
\{ {X_1},{X_2}, \ldots ,{X_K}\} $. The co-entropy of $U/R$ is
defined as follows:
\begin{equation*}
CE(U/R) = \sum\limits_{i = 1}^K {{{|{X_i}|} \over {|U|}}{{\log
}_2}|{X_i}|}.
\end{equation*}
Obviously, one has that $0 \le CE(U/R) \le {{{\log }_2}|U|}$.
\end{dfn}

\begin{dfn}
\label{KnowledgeGranulation}
{\rm{\cite{Liang2004IEREKGRST,Liang2009NMUBKGRS}}} Given an
information system $IS = (U,A)$, $R \subseteq A$ and $U/R = \{
{X_1},{X_2}, \ldots ,{X_K}\} $. The knowledge granulation of $U/R$
is defined as follows:
\begin{equation*}
GK(U/R) = {1 \over {|U{|^2}}}\sum\limits_{i = 1}^K {|{X_i}{|^2}}.
\end{equation*}
Obviously, one has that $1/|U| \le GK(U/R) \le 1$.
\end{dfn}

\begin{dfn}
\label{CombinationGranulation} {\rm{\cite{Qian2008CECGRST}}} Given an
information system $IS = (U,A)$, $R \subseteq A$ and $U/R = \{
{X_1},{X_2}, \ldots ,{X_K}\} $. The combination granulation of $U/R$
is defined as follows:
\begin{equation*}
CG(U/R) = \sum\limits_{i = 1}^K {{{|{X_i}|} \over {|U|}}
{\binom{|X_i|}{2} \over \binom{|U|}{2}}}.
\end{equation*}
Obviously, one has that $0 \le CG(U/R) \le 1$.
\end{dfn}

In the following, we modified the fitness functions $\eta
_R^{(\alpha ,\beta )}$ and $\mu _R^{(\alpha ,\beta )}$ in Definition
\ref{MonotonicityFitnessFunction} based on the expected granularity
to evaluate the significance of attributes more effectively.
Moreover, an example is provided to show that the modified fitness
functions are more effective and suitable for evaluating the
significance of attributes.

\begin{dfn} \label{ModifiedFitnessFunction}
Given a decision table $DT = (U,C \cup D)$, for any $0 \le \beta <
\alpha  \le 1$, $R \subseteq C$ and $U/D = \{ {Y_1},{Y_2}, \ldots
,{Y_M}\} $ is a classification of the universe $U$, we denote
\begin{eqnarray*}
G\eta _R^{(\alpha ,\beta )} &=& E{G_m}({\Pi _1}) - (1 - \eta _R^{(\alpha ,\beta )})E{G_m}(U/R), \\
G\mu _R^{(\alpha ,\beta )} &=& \mu _R^{(\alpha ,\beta )}E{G_m}(U/R).
\end{eqnarray*}

If $R = \emptyset$, then define $G\eta _R^{(\alpha ,\beta )} = 0$
and $G\mu _R^{(\alpha ,\beta )} = E{G_m}({\Pi _1})$. Moreover,
$G\eta _R^{(\alpha ,\beta )}$ and $G\mu _R^{(\alpha ,\beta )}$ are
denoted by $G\eta$ and $G\mu$ respectively if there is no confusion
arisen.
\end{dfn}

Clearly, $G\eta _R^{(\alpha ,\beta )}$ and $G\mu _R^{(\alpha ,\beta
)}$ take into account the granularity of the partition.

\begin{thm}
\label{ModifiedMonotonicityTheorem} Given a decision table $DT =
(U,C \cup D)$, for any $0 \le \beta  < \alpha  \le 1$ and $P,Q
\subseteq C$, we have
\begin{enumerate}
\renewcommand{\labelenumi}{\rm{(\theenumi)}}
\item $P\underline  \prec  Q \Rightarrow G\eta _P^{(\alpha ,\beta )} \ge G\eta _Q^{(\alpha ,\beta )}$,
\item $P\underline  \prec  Q \Rightarrow G\mu _P^{(\alpha ,\beta )} \le G\mu _Q^{(\alpha ,\beta )}$.
\end{enumerate}
\end{thm}
\begin{pf}
It can be easily proved according to Theorem
\ref{MonotonicityTheorem}, Definition \ref{SetGranularity} and
Definition \ref{PartitionGranularity}.
\end{pf}

By Theorem \ref{ModifiedMonotonicityTheorem} we immediately get the
following corollary.

\begin{cor}
\label{ModifiedMonotonicityCorollary} Given a decision table $DT =
(U,C \cup D)$, for any $0 \le \beta  < \alpha  \le 1$ and $P,Q
\subseteq C$, we have
\begin{enumerate}
\renewcommand{\labelenumi}{\rm{(\theenumi)}}
\item $P \supseteq Q \Rightarrow G\eta _P^{(\alpha ,\beta )} \ge G\eta _Q^{(\alpha ,\beta )}$,
\item $P \supseteq Q \Rightarrow G\mu _P^{(\alpha ,\beta )} \le G\mu _Q^{(\alpha ,\beta )}$.
\end{enumerate}
\end{cor}

Theorem \ref{ModifiedMonotonicityTheorem} and Corollary
\ref{ModifiedMonotonicityCorollary} show that the modified fitness
function $G\eta _R^{(\alpha ,\beta )}$ increases and the modified
fitness function $G\mu _R^{(\alpha ,\beta )}$ decreases as the
equivalence classes become smaller through finer partitioning, which
means that adding a new attribute into the existing subset of
condition attributes at least does not decrease $G\eta _R^{(\alpha
,\beta )}$ or increase $G\mu _R^{(\alpha ,\beta )}$, and that
deleting an attribute from the existing subset of condition
attributes at least does not increases $G\eta _R^{(\alpha ,\beta )}$
or decreases $G\mu _R^{(\alpha ,\beta )}$.

\begin{exm}
\label{StrictMonotonicityExample} Continued from Example
{\rm{\ref{NonStrictMonotonicityExample}}}.

We take the co-entropy (Definition \ref{CoEntropy}) as an example of
the expected granularity (Definition \ref{PartitionGranularity}).

According to Definition {\rm{\ref{CoEntropy}}}, we have
\begin{eqnarray*}
E{G_m}(U/P){\rm{ }} &=& {3 \over {12}}{\log _2}3 + {1 \over {12}}{\log _2}1 + {3 \over {12}}{\log _2}3 + {1 \over {12}}{\log _2}1 + {4 \over {12}}{\log _2}4 \\
&\approx& 1.46,\\
E{G_m}(U/Q) &=& {7 \over {12}}{\log _2}7 + {1 \over {12}}{\log _2}1
+ {4 \over {12}}{\log _2}4 \\
&\approx& 2.30.
\end{eqnarray*}

According to Definition {\rm{\ref{ModifiedFitnessFunction}}}, we
have

\begin{eqnarray*}
G\eta _P^{(0.60, 0.40)} &=& E{G_m}(\Pi _1) - \eta _P^{(0.60, 0.40
)}E{G_m}(U/P)
= {\log _2}12 - {5 \over {23}} \times 1.46 \approx 3.27, \\
G\eta _Q^{(0.60, 0.40)} &=& E{G_m}(\Pi _1) - \eta _Q^{(0.60,
0.40)}E{G_m}(U/Q) = {\log _2}12 - {5 \over {23}} \times 2.30
\approx 3.08,\\
G\mu _P^{(0.60, 0.40)} &=& E{G_m}(\Pi _1) - \mu _P^{(0.60, 0.40
)}E{G_m}(U/P)
= {19 \over {24}} \times 1.46 \approx 1.16, \\
G\mu _Q^{(0.60, 0.40)} &=& E{G_m}(\Pi _1) - \mu _Q^{(0.60, 0.40
)}E{G_m}(U/Q) = {19 \over {24}} \times 2.30
\approx 1.82. \\
\end{eqnarray*}

Obviously, $G\eta _P^{(0.60, 0.40)}
> G\eta _Q^{(0.60, 0.40)}$ and $G\mu _P^{(0.60, 0.40)}
< G\mu _Q^{(0.60, 0.40)}$. It is clear that the modified fitness
function $G\eta _R^{(\alpha ,\beta )}$ increases and the modified
fitness function $G\mu _R^{(\alpha ,\beta )}$ decreases with R
becoming finer.
\end{exm}

Example \ref{StrictMonotonicityExample} shows the modified fitness
functions $G\eta _R^{(\alpha ,\beta )}$ and $G\mu _R^{(\alpha ,\beta
)}$ are more powerful for evaluating the attribute subsets in some
cases. Hence, we use them as the significance measures of attributes
for guiding search to the $(\alpha ,\beta )$ lower and upper
distribution reducts, respectively.

The corresponding significance measures of attributes are defined as
follows.

\begin{dfn}
\label{significanceOutModified} Given a decision table $DT = (U,C
\cup D)$, for any $0 \le \beta < \alpha \le 1$, $R \subset C$ and
$\forall a \in C - R$, the significance measures of attribute $a$ in
$R$ with respect to $\underline {apr} _C^{(\alpha ,{\rm{ }}\beta )}$
and $\overline {apr} _C^{(\alpha ,{\rm{ }}\beta )}$ are defined as
follows:
\begin{enumerate}\upshape
\renewcommand{\labelenumi}{\rm{(\theenumi)}}
\item $SI{G_{G\eta}}(a,R,{\underline {apr} _C^{(\alpha ,{\rm{ }}\beta )}}) = G\eta_{R \cup \{ a\} }^{(\alpha ,\beta
)} - G\eta_R^{(\alpha ,\beta )}$.
\item $SI{G_{G\mu}}(a,R,{\overline {apr} _C^{(\alpha ,{\rm{ }}\beta )}}) = G\mu_R^{(\alpha ,\beta )} - G\mu_{R \cup \{ a\} }^{(\alpha ,\beta
)}$.
\end{enumerate}
\end{dfn}

Similar to Definition \ref{significanceOut}, for convenience, the
corresponding significance measures of attributes are also denoted
as $SIG_{G\eta}(a,{\underline {apr} _C^{(\alpha ,{\rm{ }}\beta )}})
= \eta_{\{ a\} }^{(\alpha ,\beta )}$ and $SIG_{G\mu}(a,$ ${\overline
{apr} _C^{(\alpha ,{\rm{ }}\beta )}}) = E{G_m}({\Pi _1}) - \mu_{\{
a\} }^{(\alpha ,\beta )}$ for any singleton attribute $a \in C$.

\subsection{The attribute core}
Attribute core plays important role in heuristic attribute reduction
algorithms. The attribute core is the set of all indispensable
attributes. In other words, the attribute core is included in all
reducts. Hence, it is often selected as the starting point in
heuristic attribute reduction algorithms to narrow the search space
of attributes. Moreover, the different attribute cores can be
defined according to the different definitions of attribute reducts.
In this subsection, we define the attribute cores for distribution
reducts and provide the method for calculating the attribute cores.

\begin{dfn}
\label{AttributeCore} Given a decision table $DT = (U,C \cup D)$,
for any $0 \le \beta < \alpha \le 1$, the attribute cores for the
 $(\alpha ,\beta )$ lower and upper distribution reducts are defined as follows:

\begin{enumerate}\upshape
\renewcommand{\labelenumi}{\rm{(\theenumi)}}
\item $COR{E_{{{\underline {apr} }^{(\alpha ,{\rm{ }}\beta )}}}}(C) =  \cap RE{D_{{{\underline {apr} }^{(\alpha ,{\rm{ }}\beta )}}}}(C)$,
\item $COR{E_{{{\overline {apr} }^{(\alpha ,{\rm{ }}\beta )}}}}(C) =  \cap RE{D_{{{\overline {apr} }^{(\alpha ,{\rm{ }}\beta )}}}}(C)$.
\end{enumerate}
where, $RE{D_{{{\underline {apr} }^{(\alpha ,{\rm{ }}\beta )}}}}(C)$
and $RE{D_{{{\overline {apr} }^{(\alpha ,{\rm{ }}\beta )}}}}(C)$
denote the set of all $(\alpha ,\beta )$ lower and $(\alpha ,\beta
)$ upper distribution reducts, respectively.
\end{dfn}

Theorem \ref{EquivalentTheorm} and Definition \ref{AttributeCore}
yield the following theorem.

\begin{thm}
\label{computeCore} Given a decision table $DT = (U,C \cup D)$, for
any $0 \le \beta < \alpha  \le 1$ and $c \in C$, we have
\begin{enumerate}\upshape
\renewcommand{\labelenumi}{\rm{(\theenumi)}}
\item $c \in COR{E_{{{\underline {apr} }^{(\alpha ,{\rm{ }}\beta )}}}}(C)$ iff  $\eta _{C - \{ c\}
}^{(\alpha ,\beta )} \ne \eta _C^{(\alpha ,\beta )}$,
\item $c \in COR{E_{{{\overline {apr} }^{(\alpha ,{\rm{ }}\beta )}}}}(C)$ iff $\mu _{C - \{ c\}
}^{(\alpha ,\beta )} \ne \mu _C^{(\alpha ,\beta )}$.
\end{enumerate}
\end{thm}
\begin{pf}
$(1)$ $`` \Rightarrow "$ Suppose $\eta _{C - \{ c\} }^{(\alpha
,\beta )} = \eta _C^{(\alpha ,\beta )}$. $c$ is dispensable in $C$
according to Theorem {\rm{\ref{Dispensable}}}. It conflicts with
condition $c \in COR{E_{{{\underline {apr} }^{(\alpha ,{\rm{ }}\beta
)}}}}(C)$, so we have $\eta _{C - \{ c\} }^{(\alpha ,\beta )} \ne
\mu _C^{(\alpha ,\beta )}$.

$`` \Leftarrow "$ If $\eta _{C - \{ c\} }^{(\alpha ,\beta )} \ne
\eta _C^{(\alpha ,\beta )}$, then $c$ is indispensable in $C$
according to Corollary {\rm\ref{Indispensable}}. Hence, $c$ must
appear in all $(\alpha ,\beta )$ lower distribution reducts of $DT$.
Thus, $c \in \cap RE{D_{{{\underline {apr} }^{(\alpha ,{\rm{ }}\beta
)}}}}(C)$, i.e., $c \in COR{E_{{{\underline {apr} }^{(\alpha ,{\rm{
}}\beta )}}}}(C)$.

$(2)$ The proof is similar to that of $(1)$.
\end{pf}

Theorem \ref{computeCore} and Definition \ref{AttributeCore} yield
the following definition.

\begin{dfn}\label{computeCore2}
Given a decision table $DT = (U, C \cup D)$, for any
$0$$\le$$\beta$$<$$\alpha$$\le 1$, the attribute cores for the
 $(\alpha ,\beta )$ lower and upper distribution reducts are defined as follows:
\begin{enumerate}\upshape
\renewcommand{\labelenumi}{\rm{(\theenumi)}}
\item $COR{E_{{{\underline {apr} }^{(\alpha ,{\rm{ }}\beta )}}}}(C) = \{ c \in C| \eta _{C - \{ c\}
}^{(\alpha ,\beta )} \ne \eta _C^{(\alpha ,\beta )}\} $,
\item $COR{E_{{{\overline {apr} }^{(\alpha ,{\rm{ }}\beta )}}}}(C) = \{ c \in C| \mu _{C - \{ c\}
}^{(\alpha ,\beta )} \ne \mu _C^{(\alpha ,\beta )}\} $.
\end{enumerate}
\end{dfn}

Definition \ref{computeCore2} states the calculation method of the
attribute cores for the $(\alpha ,\beta )$ lower and upper
distribution reducts.

An algorithm for calculating the attribute cores for the $(\alpha
,\beta )$ lower and upper distribution reducts is displayed in
Algorithm \ref{algorithm: core computing}.

\begin{algorithm}
\caption{The algorithm for calculating the attribute core for the
$(\alpha ,\beta )$ lower (upper) distribution
reduct}\label{algorithm: core computing}
  \textbf{Input}: A decision table $DT = (U,C \cup
D)$, threshold values $(\alpha, \beta)$\\
  \textbf{Output}: An attribute core for the $(\alpha
,\beta )$ lower (upper) distribution reduct\\
 \textbf{Note}: $(T, \Delta)  \in  \{(\underline {apr}, \eta), (\overline {apr}, \mu)\}$ \\
  \begin{algorithmic}[1]
    \STATE Let $COR{E_{{T^{(\alpha ,\beta )}}}}(C) =
    \emptyset$
    \FOR {Each $c \in C$}
     \STATE Calculate $\Delta _{C - \{ c\}}^{(\alpha ,\beta )}$
       \IF {{$\Delta _{C - \{ c\}}^{(\alpha ,\beta )} \ne \Delta _C^{(\alpha ,\beta)}$}}
          \STATE $COR{E_{{T^{(\alpha ,{\rm{ }}\beta )}}}}(C) = COR{E_{{T^{(\alpha ,{\rm{ }}\beta )}}}}(C) \cup \{ c\} $
        \ENDIF
     \ENDFOR
     \STATE Return $COR{E_{{T^{(\alpha ,\beta)}}}}(C)$
  \end{algorithmic}
\end{algorithm}

\subsection{Attribute reduction algorithms}
In this subsection, we proposed two heuristic attribute reduction
algorithms based on the addition-deletion method (Algorithm
\ref{addition-deletion}) and the deletion method (Algorithm
\ref{deletion}) to obtain the distribution reducts in probabilistic
rough set model. Two algorithm details are shown in Algorithm
\ref{algorithm: addition-deletion} and Algorithm \ref{algorithm:
deletion}.

\begin{algorithm}
\caption{The addition-deletion method for computing the distribution
reducts of $DT$}\label{algorithm: addition-deletion}
  \textbf{Input}: A decision table $DT = (U,C \cup
D)$, threshold values $(\alpha, \beta)$\\
  \textbf{Output}: An $(\alpha, \beta)$ lower (upper) distribution
reduct of $DT$\\
\textbf{Method}: Addition-deletion method\\
\textbf{Note}: $(T, \Delta) \in \{(\underline {apr}, \eta), (\overline {apr}, \mu)\}$ \\
  \begin{algorithmic}[1]
    \STATE Calculate $COR{E_{{T^{(\alpha ,\beta )}}}}(C)$ by Algorithm \ref{algorithm: core computing}
      \STATE Calculate $\Delta _C^{(\alpha ,\beta )}{\rm{ }}$
      \STATE Let $R = COR{E_{{T^{(\alpha ,{\rm{ }}\beta )}}}}(C)$, $CA = C - R$
        \IF {$\Delta _R^{(\alpha ,\beta )} = \Delta _C^{(\alpha ,\beta )}$}
        \STATE go to Step \ref{algorithm: addition-deletion-return}
        \ENDIF
      \STATE // Addition
      \WHILE {$\Delta _R^{(\alpha ,\beta )} \ne \Delta _C^{(\alpha ,\beta )}$}
        \FOR {Each $a \in CA$} \label{code:fram:classify}
            \STATE Calculate $SI{G_{G\Delta}}(a,R,{T _C^{(\alpha ,{\rm{ }}\beta )}})$
        \ENDFOR
        \IF {$SI{G_{G\Delta}}(a,R,{T _C^{(\alpha ,\beta )}}) = {\max _{a \in C - R}}SI{G_{G\Delta}}(a,R,{T _C^{(\alpha ,{\rm{ }}\beta )}})$}
        \STATE $R = R \cup \{ a\} $, $CA = CA - \{a\}$
        \ENDIF
      \ENDWHILE
      \STATE // Deletion
      \STATE Let $CD = R$ \label{backdeletion}
     \FOR {Each $a \in CD$}
     \STATE Calculate $SI{G_{G\Delta}}(a,T)$
     \ENDFOR
     \STATE Sort attributes in $CD$ according to $SI{G_{G\Delta}}(a,T)$ in a ascending order
     \WHILE {$CD \ne \emptyset $}
        \STATE $CD = CD - \{ a\} $, where $a$ is the first element of $CD$
          \IF{$\Delta _{R - \{a \}}^{(\alpha ,\beta )} = \Delta _C^{(\alpha ,\beta )}$}
            \STATE $R = R - \{a \}$
          \ENDIF
     \ENDWHILE
     \STATE Return $R$ \label{algorithm: addition-deletion-return}
  \end{algorithmic}
\end{algorithm}

\begin{algorithm}[tb!]
\caption{The deletion method for computing the distribution reducts
of $DT$}\label{algorithm: deletion}
  \textbf{Input}: A decision table $DT = (U,C \cup
D)$, threshold values $(\alpha, \beta)$\\
  \textbf{Output}: An $(\alpha, \beta)$ lower (upper) distribution
reduct of $DT$\\
\textbf{Method}: Deletion method\\
\textbf{Note}: $(T, \Delta) \in \{(\underline {apr}, \eta), (\overline {apr}, \mu)\}$ \\
  \begin{algorithmic}[1]
    \STATE Calculate $\Delta _C^{(\alpha ,\beta )}{\rm{ }}$
    \STATE Let $R = C$, $CD = C$
     \FOR {Each $a \in CD$}
     \STATE Calculate $SI{G_{G\Delta}}(a,T)$
     \ENDFOR
     \STATE Sort attributes in $CD$ according to $SI{G_{G\Delta}}(a,T)$ in a ascending order
     \WHILE {$CD \ne \emptyset $}
        \STATE $CD = CD - \{ a\} $, where $a$ is the first element of $CD$
          \IF{$\Delta _{R - \{a \}}^{(\alpha ,\beta )} = \Delta _C^{(\alpha ,\beta )}$}
            \STATE $R = R - \{a \}$
          \ENDIF
     \ENDWHILE
     \STATE Return $R$
  \end{algorithmic}
\end{algorithm}

\subsection{An illustrative example}
We have developed two heuristic attribute reduction algorithms to
obtain distribution reducts. In this subsection, we present an
example to show the validity of the proposed algorithms.

\begin{exm}
\label{AlgorithmExample}(Continued from Example
\ref{MonotonicityExample}) For Table \ref{DecisionSystem1} shown in
Example \ref{MonotonicityExample}, we take the $(0.6, 0.4)$ low
distribution reduct as an example. Moreover, the co-entropy
(Definition \ref{CoEntropy}) is used as an example of the expected
granularity (Definition \ref{PartitionGranularity}). We can
calculate the $(0.60, 0.40)$ low distribution reduct by Algorithm
{\rm{\ref{algorithm: addition-deletion}}}.

By computing, we can obtain $\eta _C^{(0.60, 0.40)} \approx 0.32$.

According to core computing step in Algorithm {\rm{\ref{algorithm:
addition-deletion}}}, we first calculate $\eta _{C - \{
a_i\}}^{(0.60 , 0.40)}$ for each attribute $a_i \in C$ as follows:

$\eta _{C - \{ a_1\}}^{(0.60 , 0.40)} \approx 0.27$, $\eta _{C -
\{a_2\}}^{(0.60 , 0.40)} \approx 0.32$, $\eta _{C - \{ a_3\}}^{(0.75
, 0.60)} \approx 0.23$,

$\eta _{C - \{ a_4\}}^{(0.60 , 0.40)} \approx 0.32$, $\eta _{C - \{
a_5\}}^{(0.60 , 0.40)} \approx 0.32$, $\eta _{C - \{ a_6\}}^{(0.75 ,
0.60)} \approx 0.32$.

Therefore, $COR{E_{{{\underline {apr} }^{(0.60, 0.40)}}}}(C) = \{
{a_1},{a_3}\} $.

Let $R = COR{E_{{{\underline {apr} }^{(0.60, 0.40)}}}}(C) = \{
{a_1},{a_3}\} $, then $\eta _R^{(0.60 , 0.40)} \approx 0.09$.

Because $\eta _R^{(0.60 , 0.40)} \ne \eta _C^{(0.60 , 0.40)}$,
according to the addition step in Algorithm {\rm{\ref{algorithm:
addition-deletion}}} we calculate

$G\eta _{R \cup \{ a_2\}}^{(0.60 , 0.40)} \approx 1.64$, $G\eta _{R
\cup \{ a_4\}}^{(0.60 , 0.40)} \approx 1.53$,

$G\eta _{R \cup \{ a_5\}}^{(0.60 , 0.40)} \approx 1.84$, $G\eta _{R
\cup \{ a_6\}}^{(0.60 , 0.40)} \approx 1.91$.

Because $G\eta _{R \cup \{ a_6\}}^{(0.60 , 0.40)}$ is maximum, we
select $a_6$. Let $R = R \cup \{ {a_6}\} = \{{a_1},{a_3},{a_6}\} $,
then $\eta _R^{(0.60 , 0.40)} \approx 0.27$.

Because $\eta _R^{(0.60 , 0.40)} \ne \eta _C^{(0.60 , 0.40)}$, we
calculate $\eta _{R \cup \{ a_i\}}^{(0.60 , 0.40)}$ for each
attribute $a_i \in C - R$ as follows:

$G\eta _{R \cup \{ a_2\}}^{(0.60 , 0.40)} \approx 1.91$, $G\eta _{R
\cup \{ a_4\}}^{(0.60 , 0.40)} \approx 2.05$, $G\eta _{R \cup \{
a_5\}}^{(0.60 , 0.40)} \approx 2.14$.

Because $G\eta _{R \cup \{ a_5\}}^{(0.60 , 0.40)}$ is maximum, we
select $a_5$. Let $R = R \cup \{ {a_6}\} = \{{a_1},{a_3},{a_6}\} $,
then $\eta _R^{(0.60 , 0.40)} \approx 0.32$.

Thus, we have $\eta _R^{(0.60 , 0.40)} = \eta _C^{(0.60 , 0.40)}$
when $R = {\{ {a_1},{a_3},{a_5},{a_6}\}}$.

According to the deletion step in Algorithm {\rm{\ref{algorithm:
addition-deletion}}}, we get a $(0.60, 0.40)$ low distribution
reduct ${\{ {a_1},{a_3},{a_5},{a_6}\}}$ because we have $\eta _{R -
\{a_i\}}^{(0.60 , 0.40)} \ne \eta _C^{(0.60 , 0.40)}$ for $\forall
a_i \in R$.

It can be easily calculated that

$\underline {apr} _{\{ {a_1},{a_3},{a_5},{a_6}\}}^{(0.60 , 0.40)} =
( \{{x_1},{{x_2}},{x_{5}}\}, \{{x_6},{x_7},{x_{10}},{x_{11}}\} )$.

Hence, we have $\underline {apr} _{\{
{a_1},{a_3},{a_5},{a_6}\}}^{(0.60 , 0.40)} = \underline {apr}
_C^{(0.60 , 0.40)}$.

In other words, the attribute set $\{ {a_1},{a_3},{a_5},{a_6}\}$ can
keep the $(0.60 , 0.40)$ lower approximations of all decision
classes unchanged.

Similar to the computation of the $(0.60, 0.40)$ low distribution
reduct, it is not difficult to find that ${\{ {a_1}, {a_3}, {a_4},
{a_6}\}}$ is a $(0.60, 0.40)$upper distribution reduct.

By Algorithm {\rm{\ref{algorithm: deletion}}} we can obtain ${\{
{a_1},{a_3},{a_4},{a_6}\}}$ is a $(0.60, 0.40)$ low distribution
reduct and ${\{ {a_1}, {a_2}, {a_3}, {a_5}\}}$  is a $(0.60, 0.40)$
upper distribution reduct.
\end{exm}

\section{Experimental results} \label{experimental}
In this section, a series of experiments were designed to
demonstrate that our methods proposed are effective and applicable.
Ten benchmark real-world data sets were chosen for experimental
evaluation. All the data sets were obtained from the UCI Repository
of Machine Learning databases \cite{UCI}. These data sets have been
widely used in literatures. The general information about the
selected UCI data sets is summarized in Table \ref{UCIData}, where
$|U|$ and $|C|$ denote the number of objects and the condition
attributes, respectively. $|{V_d}|$ denotes the number of decision
classes.

Since the data sets may contain missing values or continuous
attributes, they would be handled in advance prior to attribute
reduction. Missing values were filled with mean values for
continuous attributes and mode values for nominal attributes.
Continuous attributes were discretized using equal-frequency
discretization method. All preprocessing methods were implemented by
using WEKA filters \cite{WEKA}.

\setlength{\tabcolsep}{18pt}
\begin{table}[h]
\caption{Description of the datasets} \label{UCIData}
\begin{center}
\begin{tabular}{ccccc}
\hline
ID & Data sets & $|U|$ & $|C|$ & $|{V_d}|$\\
\hline
1 & Horse-colic & 368 & 22 & 2\\
2 & Primary-tumor & 339 & 17 & 21\\
3 & Vehicle & 846 & 18 & 4\\
4 & Voting & 436 & 16 & 2\\
5 & Zoo & 101 & 16 & 7\\
6 & Credit Approval & 690 & 15 & 2\\
7 & Segment & 2310 & 19 & 7\\
8 & Kr-vs-kp & 3196 & 36 & 2\\
9 & Hypothyroid & 3772 & 29 & 4\\
10 & German & 1000 & 20 & 2\\
\hline
\end{tabular}
\end{center}
\end{table}

\subsection{The monotonicity experiments} \label{MonotonicityExperiment}
In this subsection, several experiments were performed to verify the
effectiveness of the proposed fitness functions in Section
\ref{HeuristicAlgorithm}. In the experiments, we took the co-entropy
(CE,Definition \ref{CoEntropy}), knowledge granulation
(KG,Definition \ref{KnowledgeGranulation}) and combination
granulation (CG,Definition \ref{CombinationGranulation}) as examples
of the expected granularity. Hence, both of fitness functions
$G\eta$ and $G\mu$ (Definition \ref{ModifiedFitnessFunction}) can
adopt three types of implementations based on CE, KG and CG. The
threshold parameters $\alpha$ and $\beta$ are set to 0.6 and 0.4
respectively.

\begin{figure}[h]
  \subfigure[$\eta$ and $G\eta$]{
    \label{fig:subfig:a}
    \includegraphics[width=2.5in]{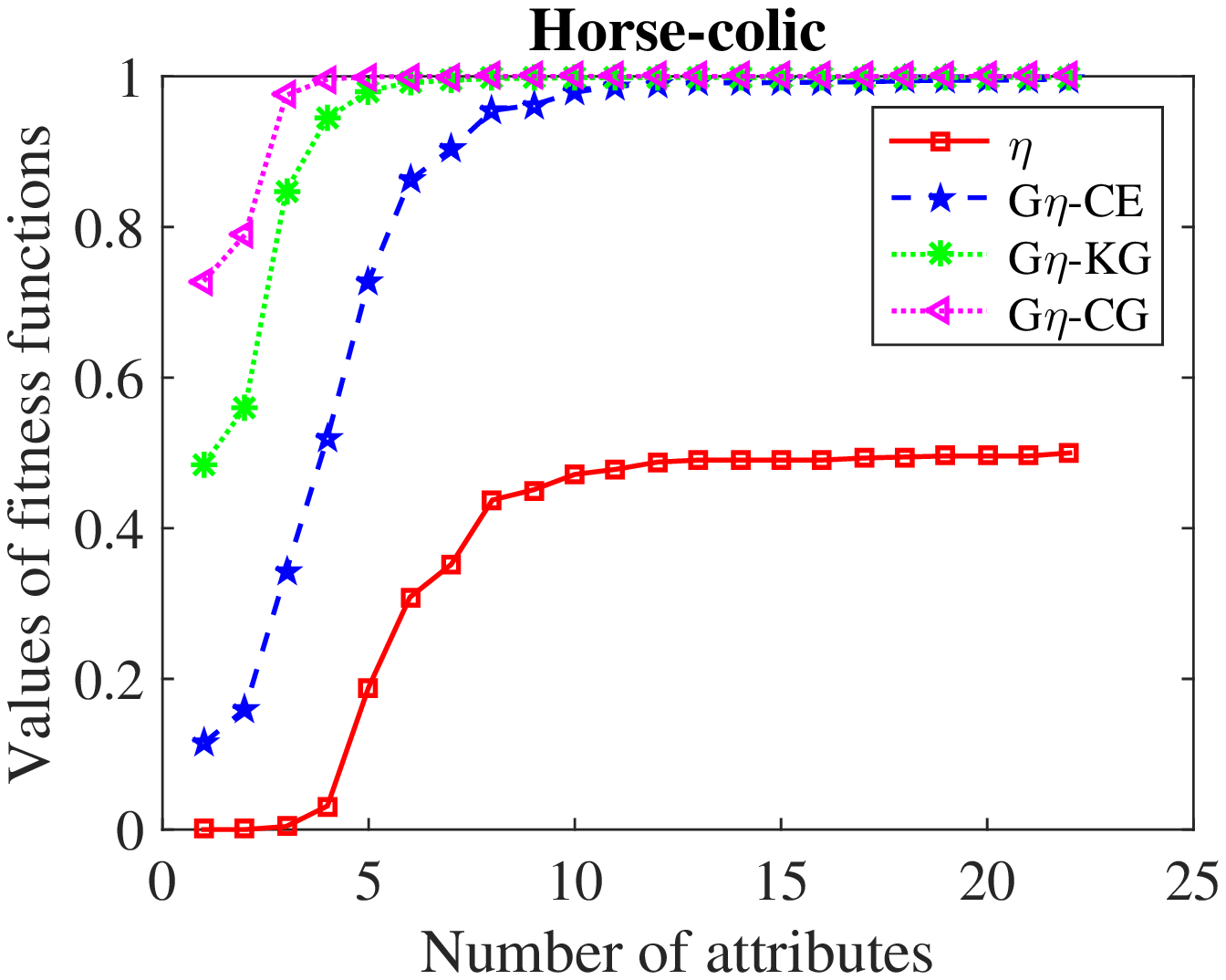}}
  \subfigure[$\mu$ and $G\mu$]{
    \label{fig:subfig:b}
    \includegraphics[width=2.5in]{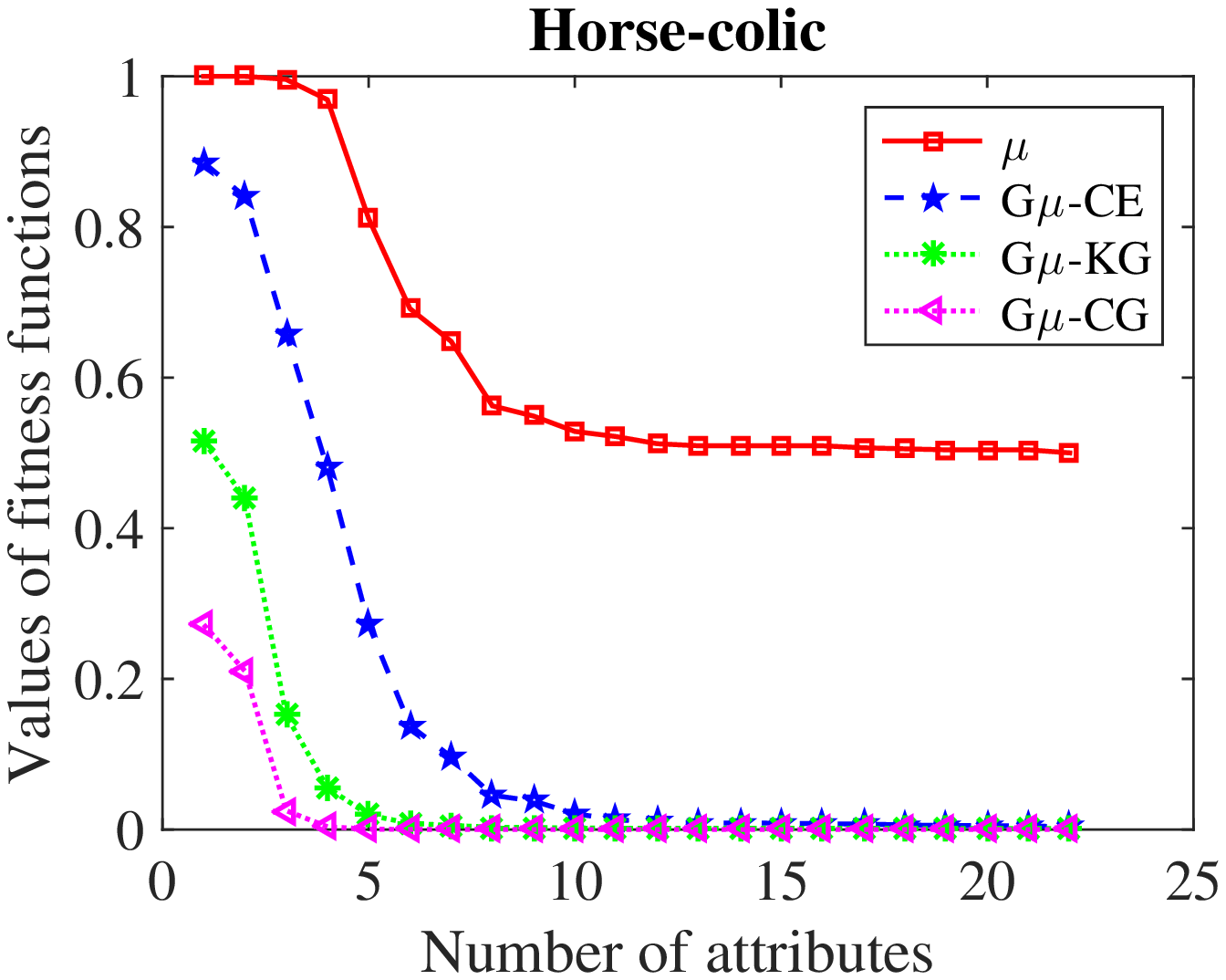}}
    \caption{Monotonicity of fitness functions on data set Horse-colic.}
  \label{fig:Horse-colic}
\end{figure}

\begin{figure}[h]
  \subfigure[$\eta$ and $G\eta$]{
    \label{fig:subfig:a}
    \includegraphics[width=2.5in]{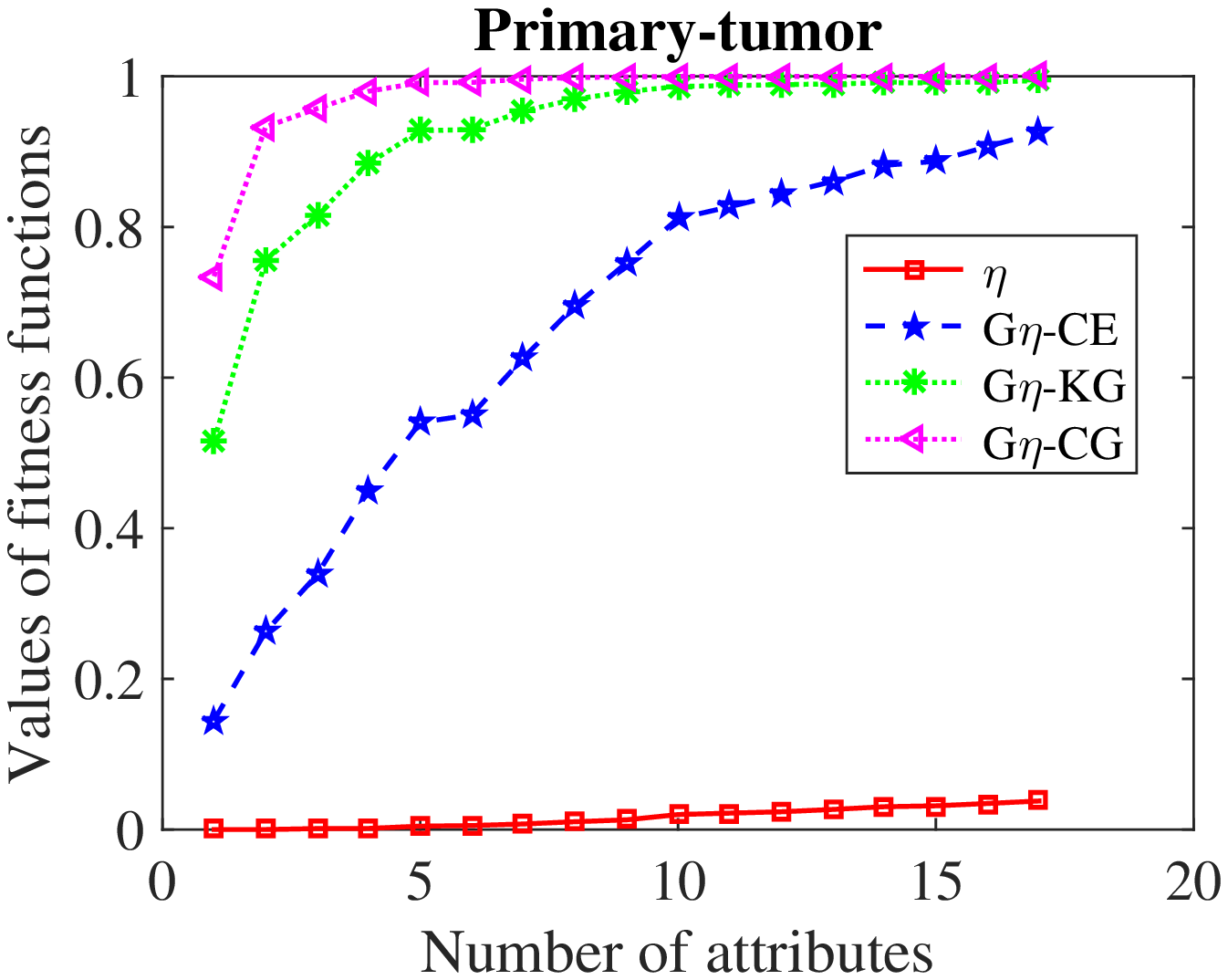}}
  \subfigure[$\mu$ and $G\mu$]{
    \label{fig:subfig:b}
    \includegraphics[width=2.5in]{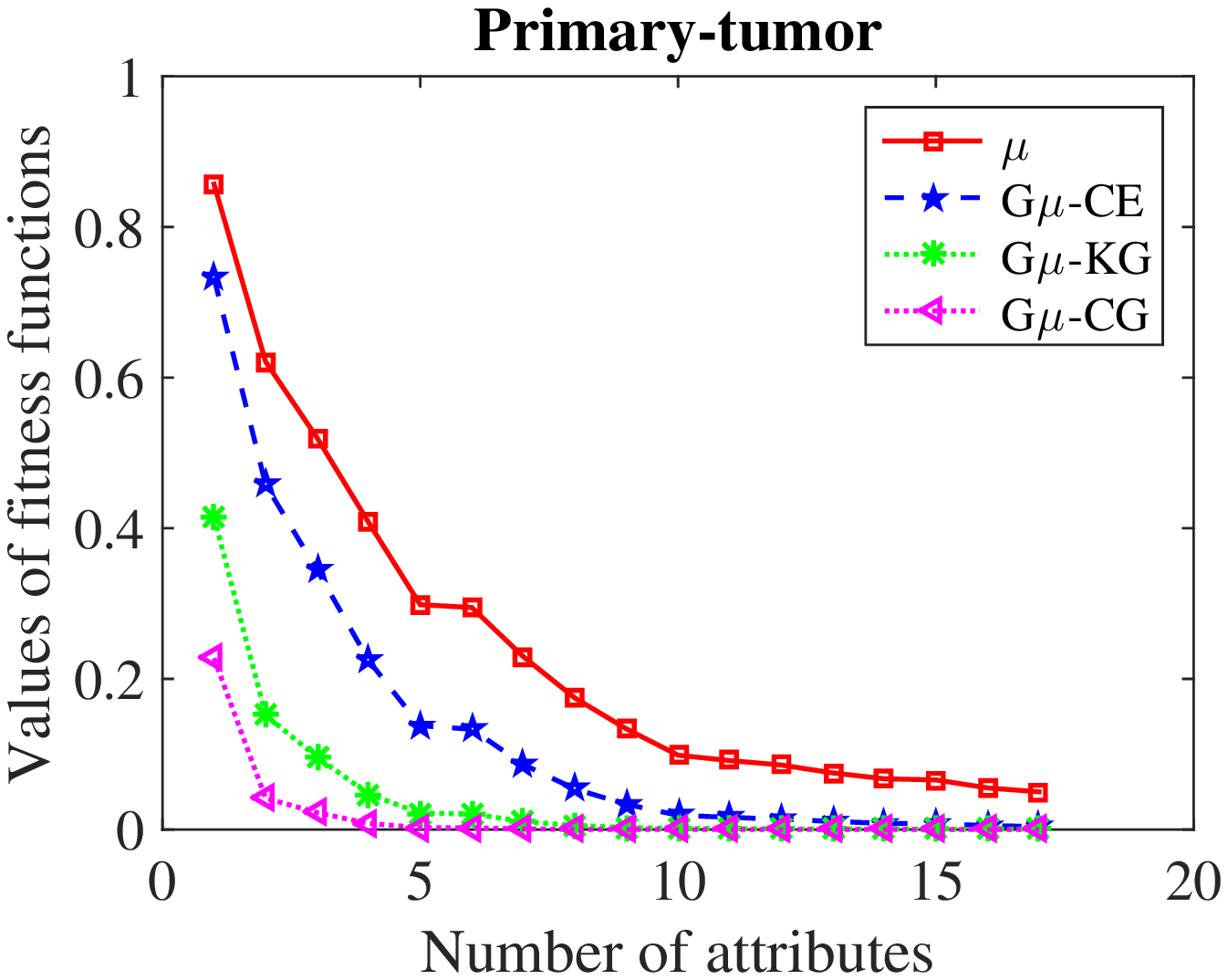}}
    \caption{Monotonicity of fitness functions on data set Primary-tumor.}
  \label{fig:Primary-tumor}
\end{figure}

\begin{figure}[h]
  \subfigure[$\eta$ and $G\eta$]{
    \label{fig:subfig:a}
    \includegraphics[width=2.5in]{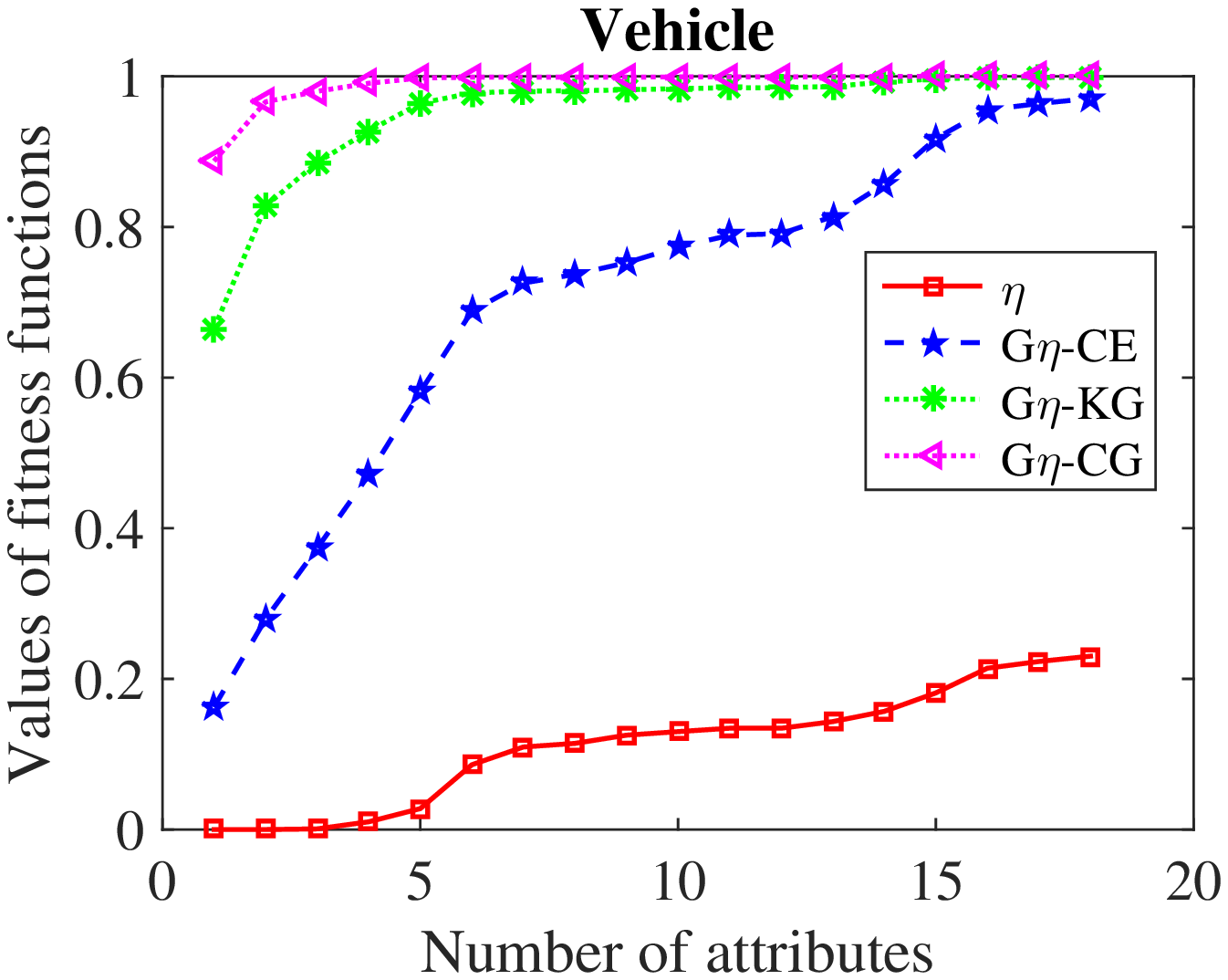}}
  \subfigure[$\mu$ and $G\mu$]{
    \label{fig:subfig:b}
    \includegraphics[width=2.5in]{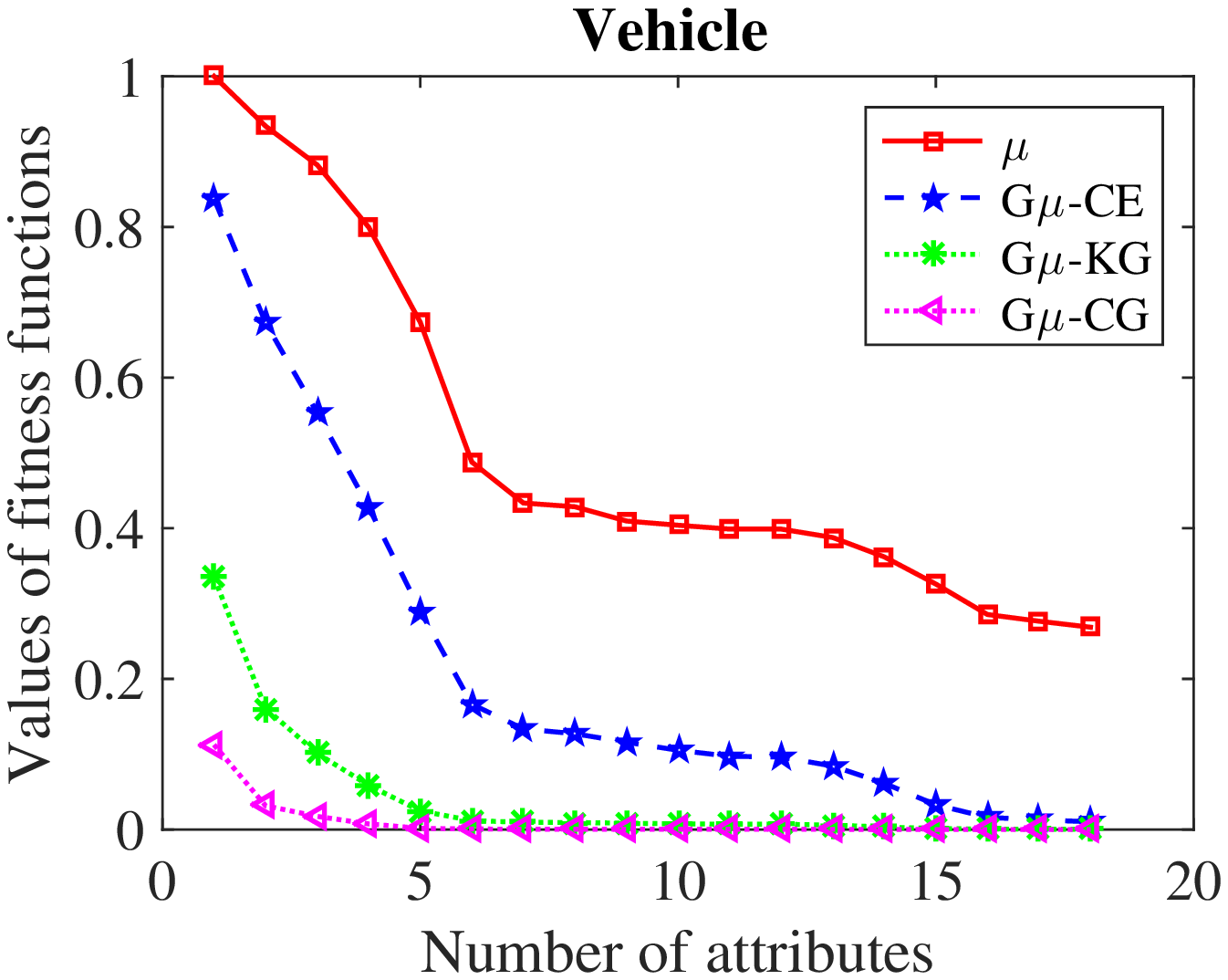}}
    \caption{Monotonicity of fitness functions on data set Vehicle.}
  \label{fig:Vehicle}
\end{figure}

\begin{figure}[h]
  \subfigure[$\eta$ and $G\eta$]{
    \label{fig:subfig:a}
    \includegraphics[width=2.5in]{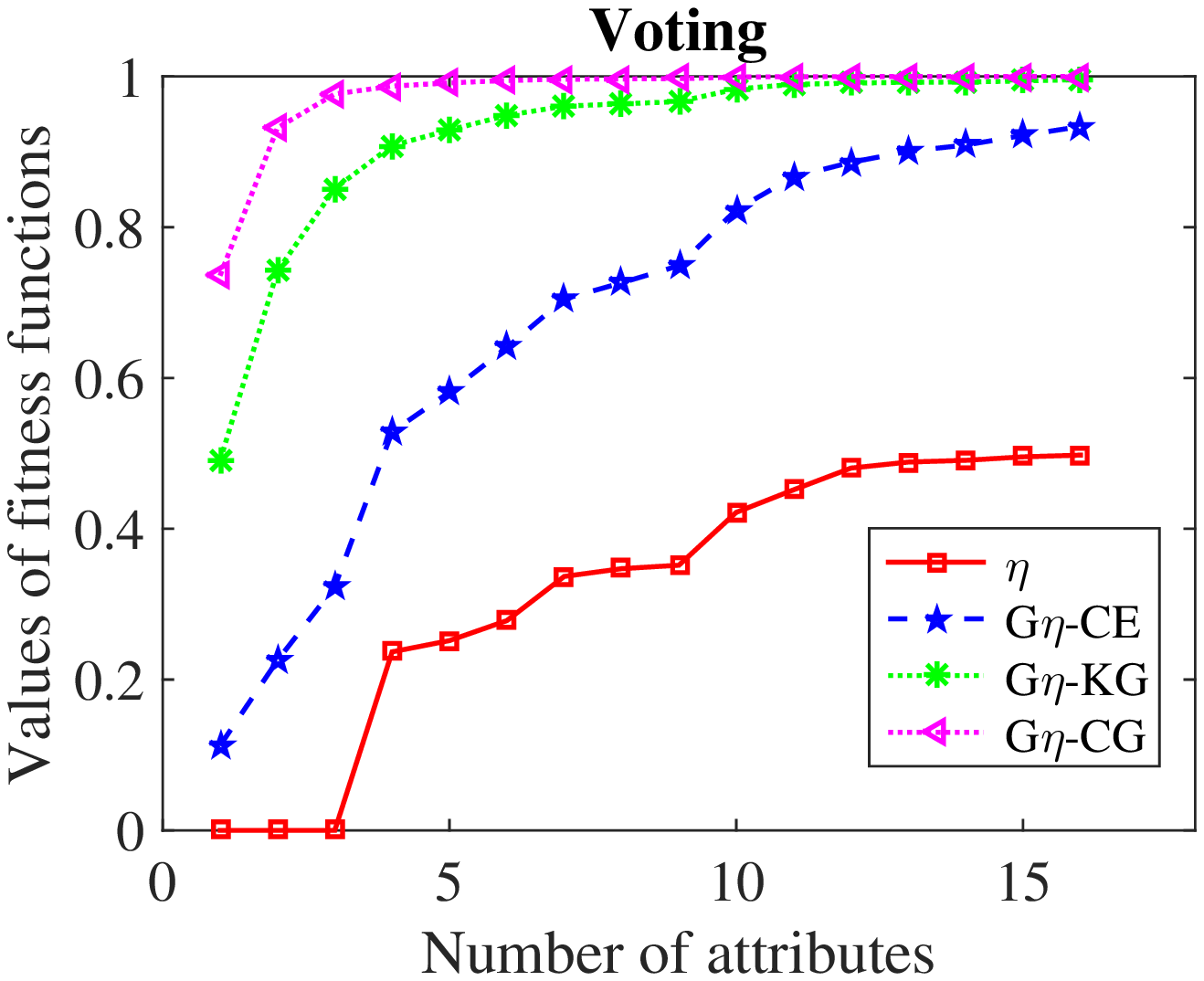}}
  \subfigure[$\mu$ and $G\mu$]{
    \label{fig:subfig:b}
    \includegraphics[width=2.5in]{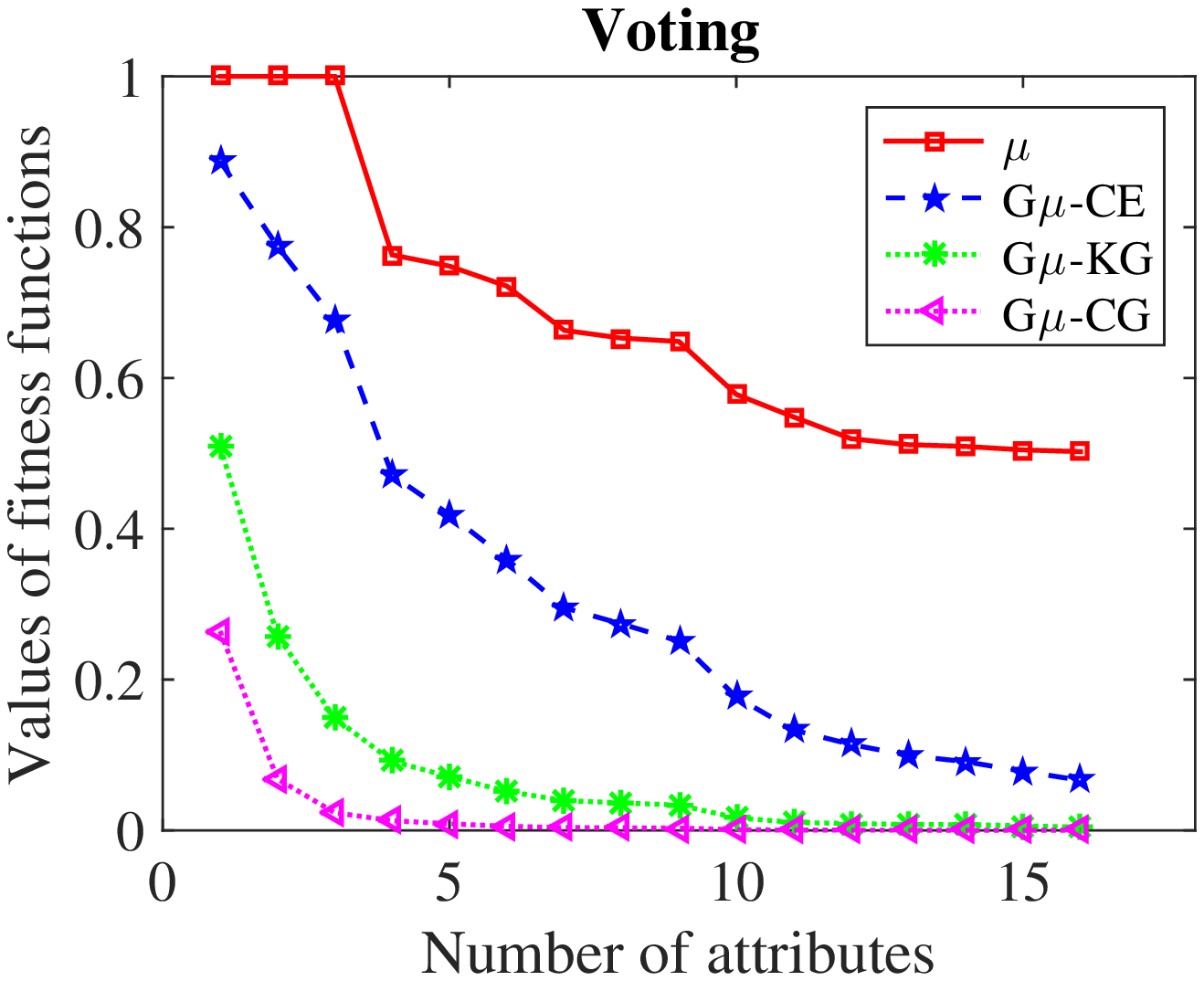}}
    \caption{Monotonicity of fitness functions on data set Voting.}
  \label{fig:Voting}
\end{figure}

\begin{figure}[h]
  \subfigure[$\eta$ and $G\eta$]{
    \label{fig:subfig:a}
    \includegraphics[width=2.5in]{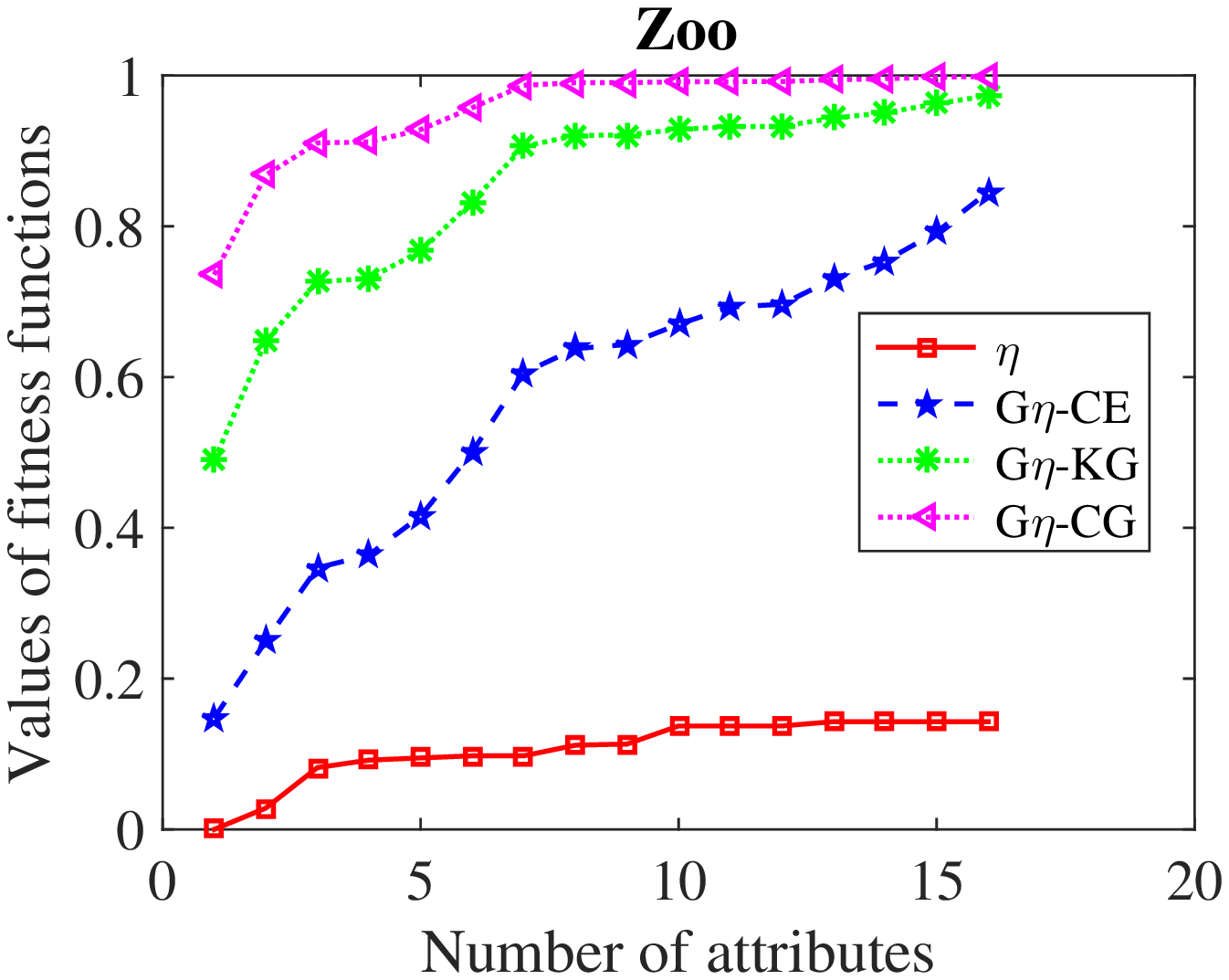}}
  \subfigure[$\mu$ and $G\mu$]{
    \label{fig:subfig:b}
    \includegraphics[width=2.5in]{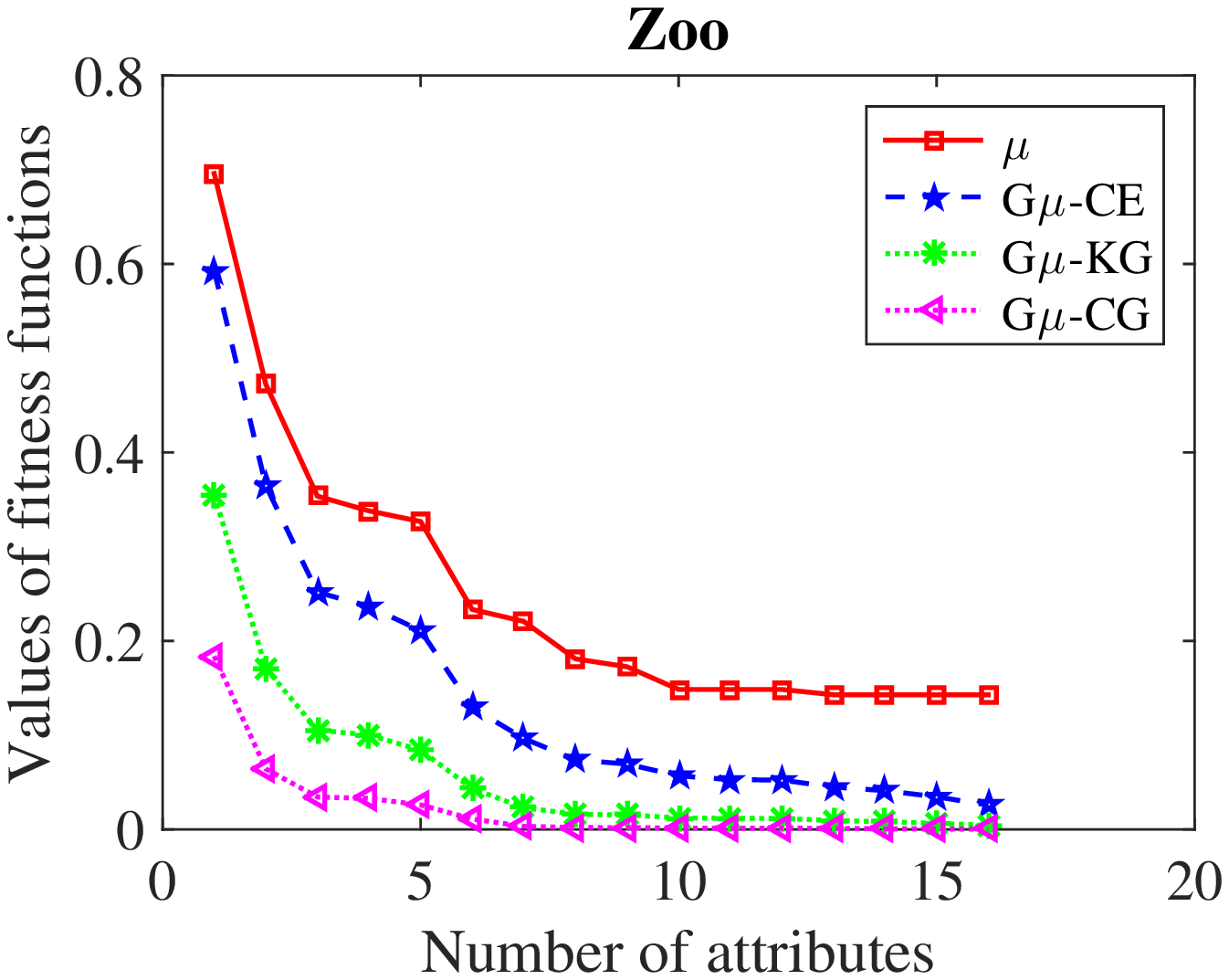}}
    \caption{Monotonicity of fitness functions on data set Zoo.}
  \label{fig:Zoo}
\end{figure}

\begin{figure}[h]
  \subfigure[$\eta$ and $G\eta$]{
    \label{fig:subfig:a}
    \includegraphics[width=2.5in]{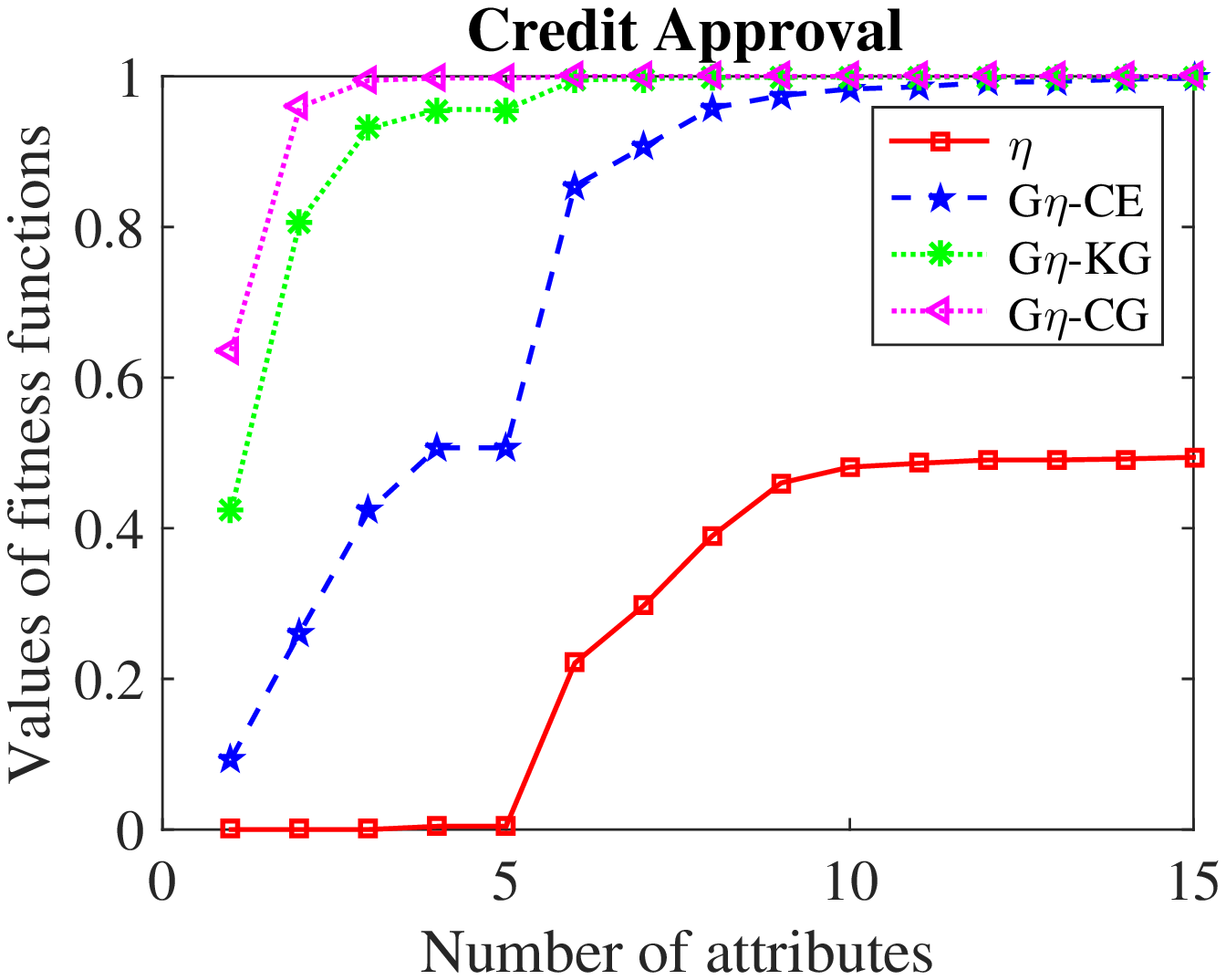}}
  \subfigure[$\mu$ and $G\mu$]{
    \label{fig:subfig:b}
    \includegraphics[width=2.5in]{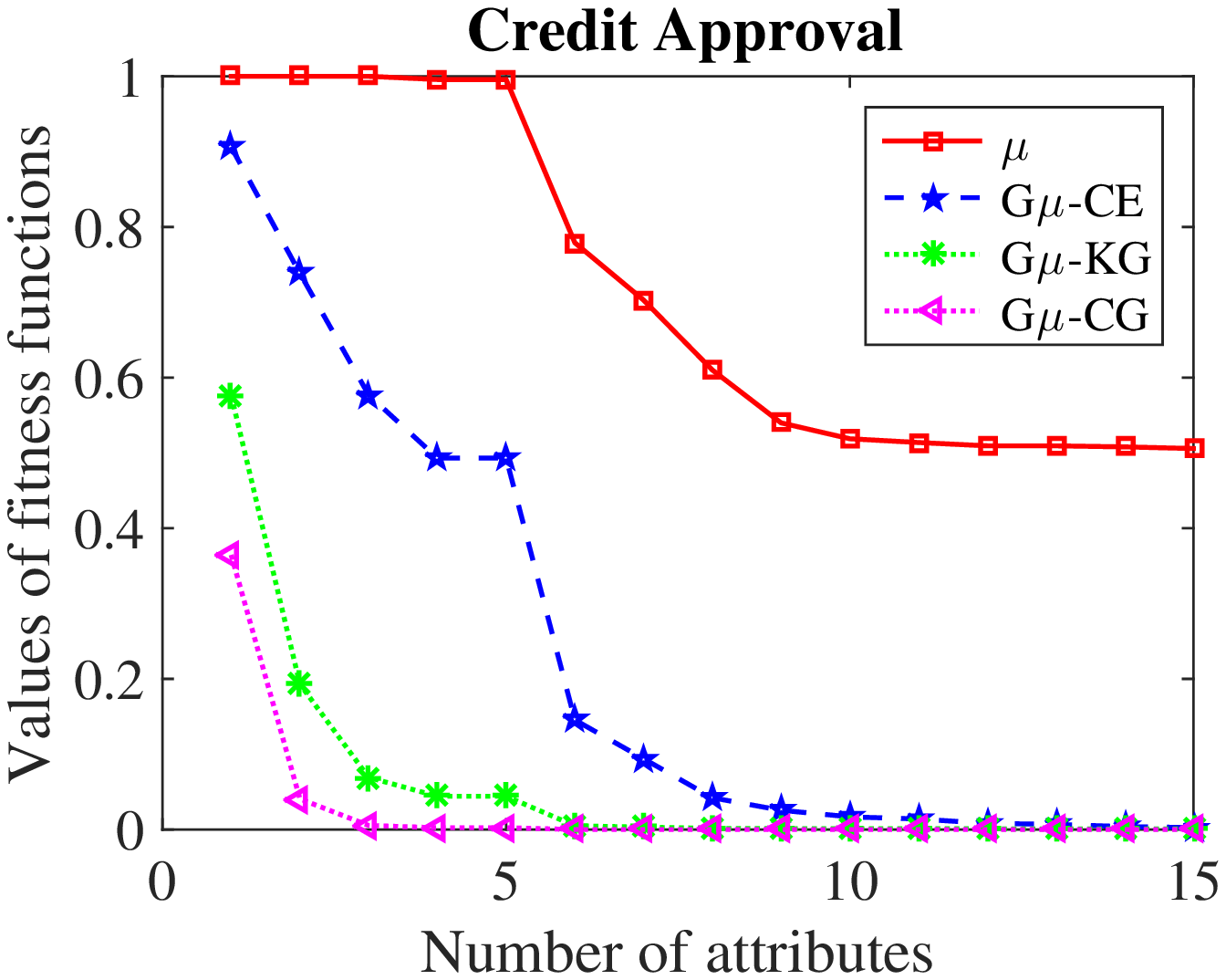}}
    \caption{Monotonicity of fitness functions on data set Credit Approval.}
  \label{fig:Credit Approval}
\end{figure}

\begin{figure}[h]
  \subfigure[$\eta$ and $G\eta$]{
    \label{fig:subfig:a}
    \includegraphics[width=2.5in]{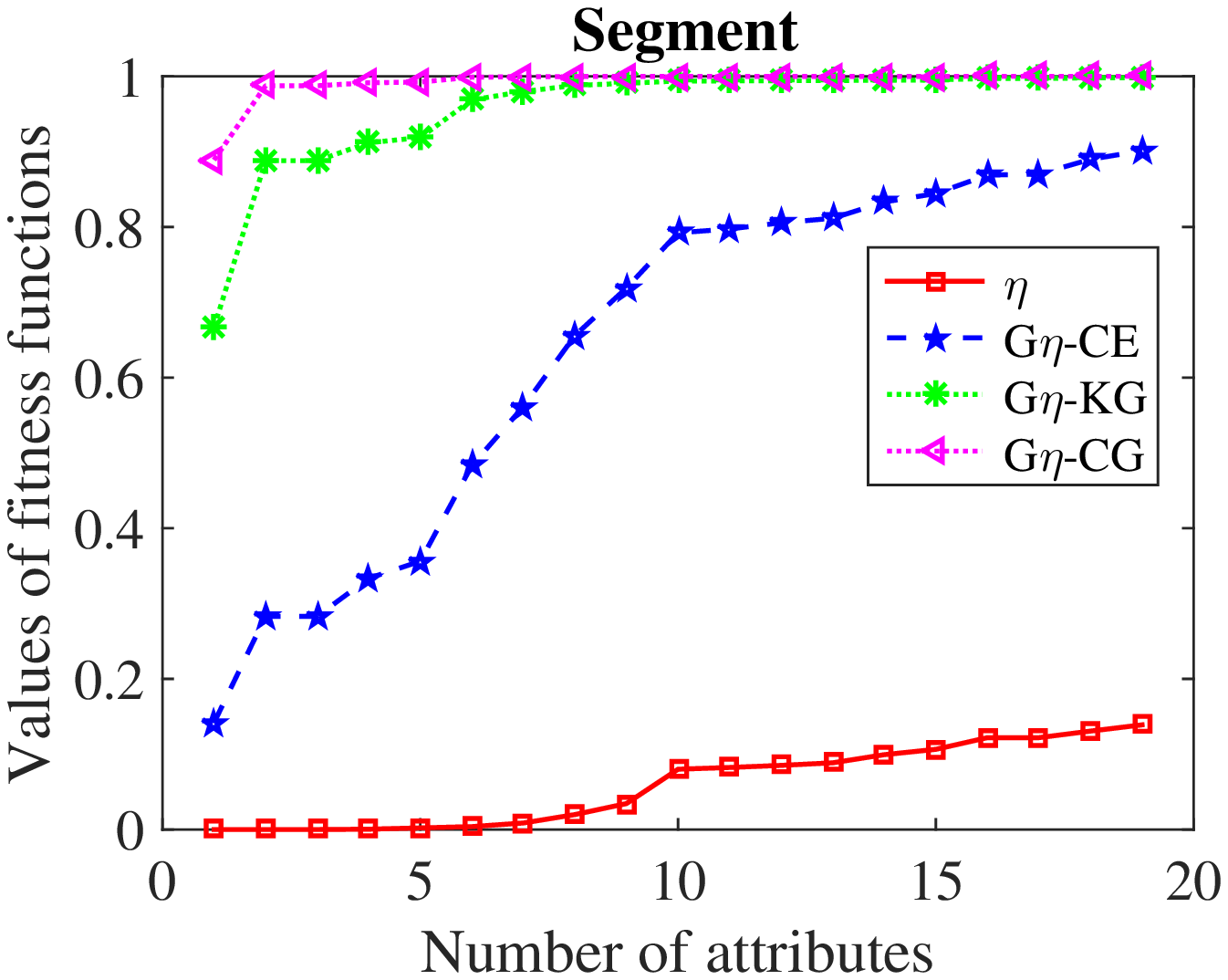}}
  \subfigure[$\mu$ and $G\mu$]{
    \label{fig:subfig:b}
    \includegraphics[width=2.5in]{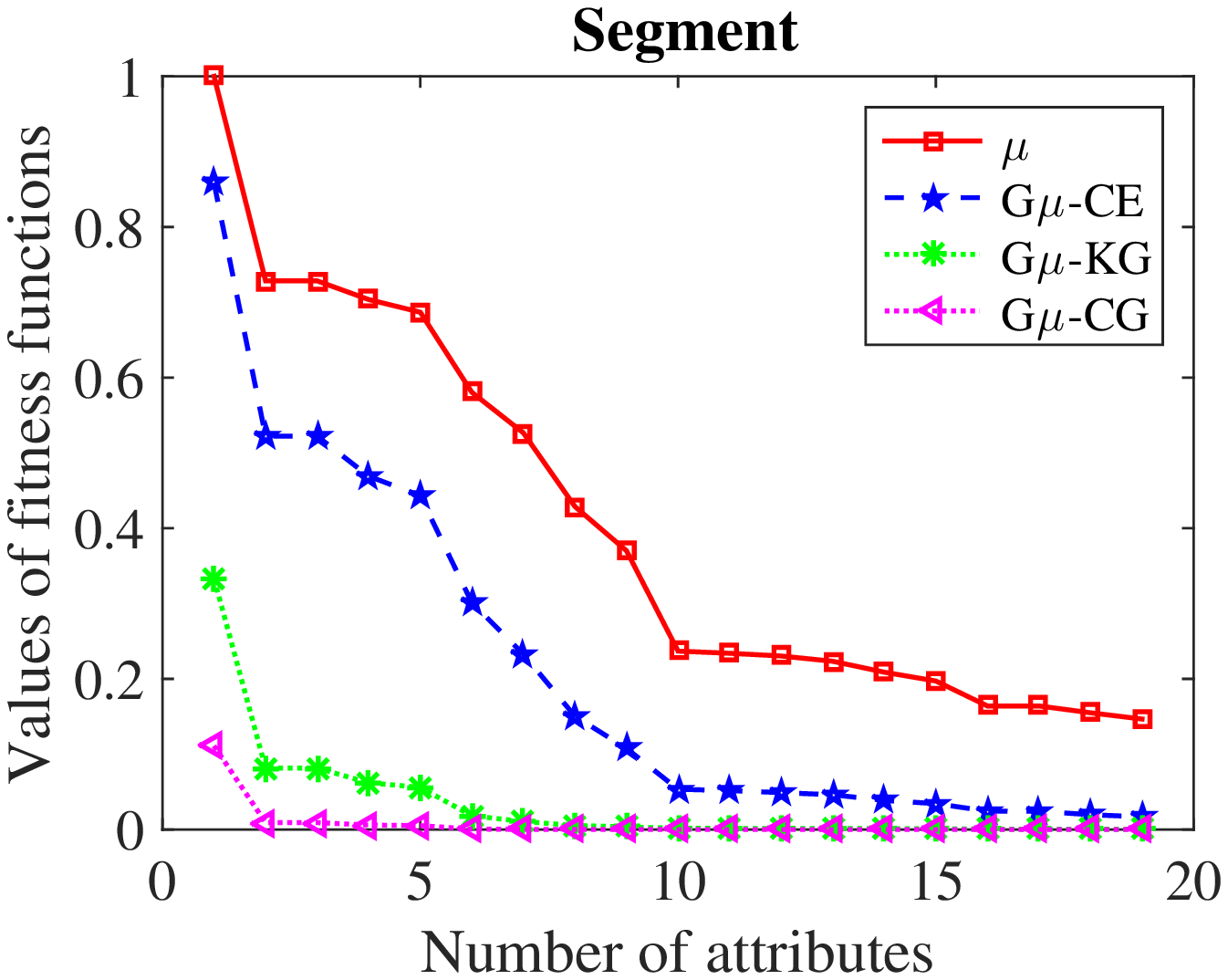}}
    \caption{Monotonicity of fitness functions on data set Segment.}
  \label{fig:Segment}
\end{figure}

\begin{figure}[h]
  \subfigure[$\eta$ and $G\eta$]{
    \label{fig:subfig:a}
    \includegraphics[width=2.5in]{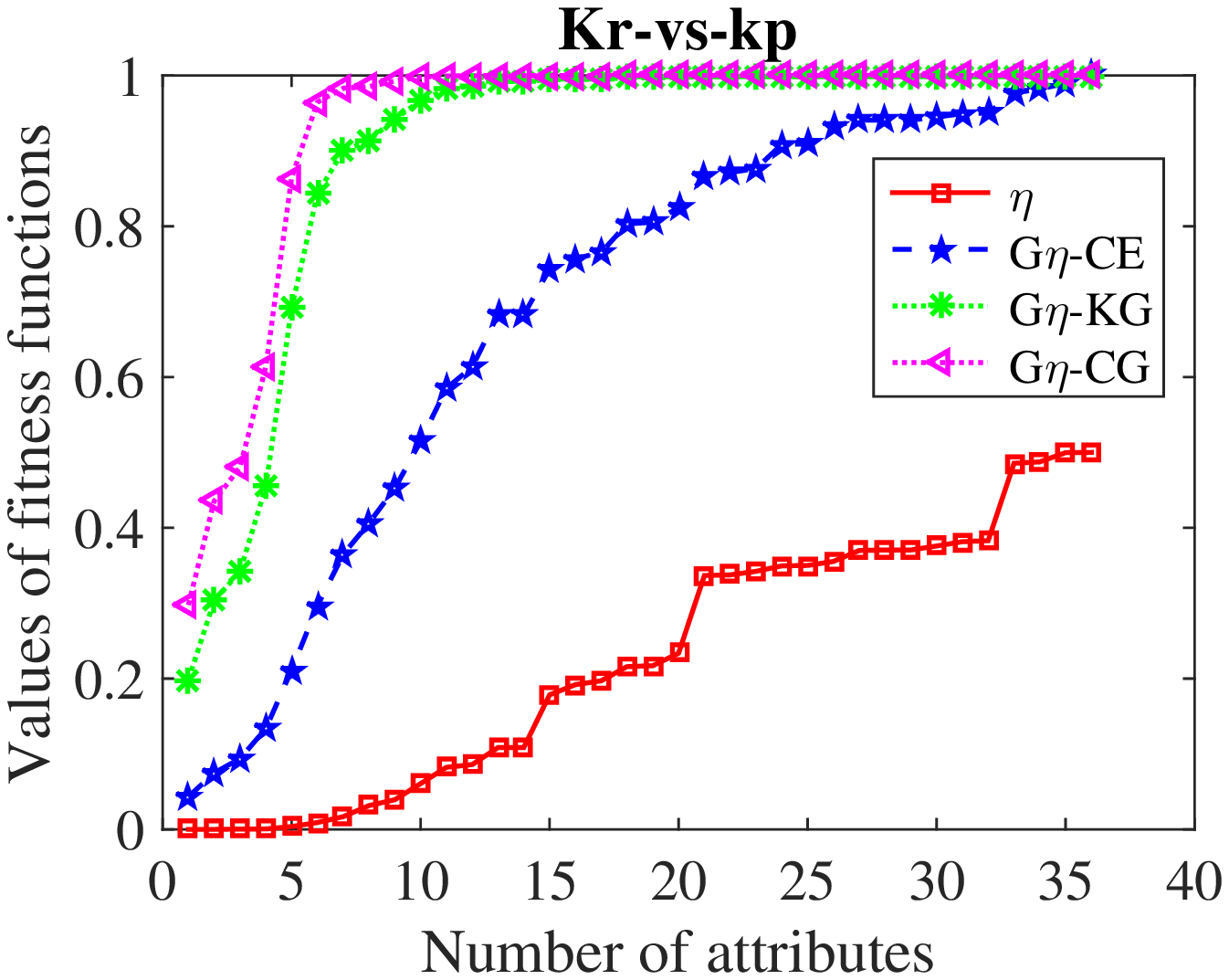}}
  \subfigure[$\mu$ and $G\mu$]{
    \label{fig:subfig:b}
    \includegraphics[width=2.5in]{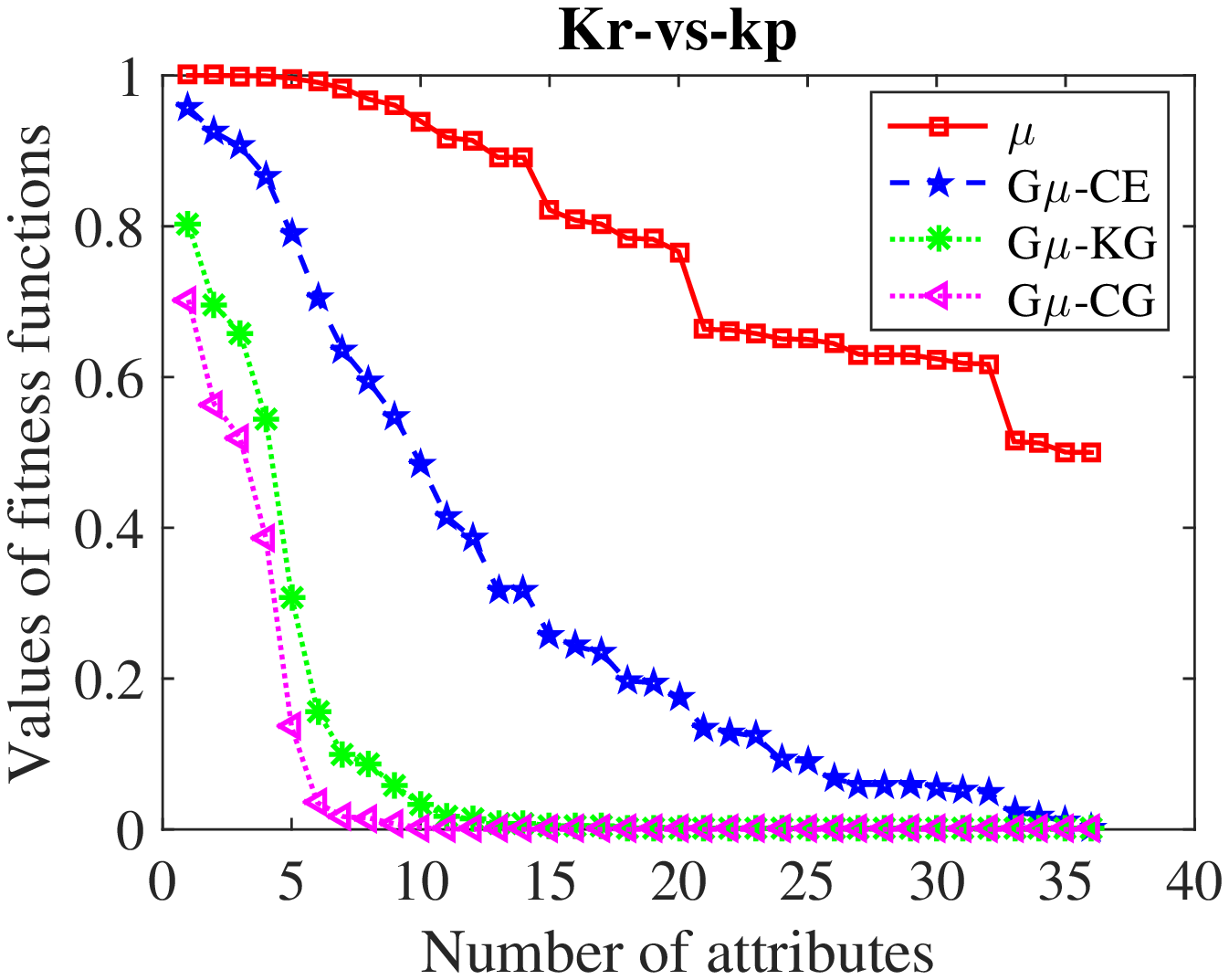}}
    \caption{Monotonicity of fitness functions on data set Kr-vs-kp.}
  \label{fig:Kr-vs-kp}
\end{figure}

\begin{figure}[h]
  \subfigure[$\eta$ and $G\eta$]{
    \label{fig:subfig:a}
    \includegraphics[width=2.5in]{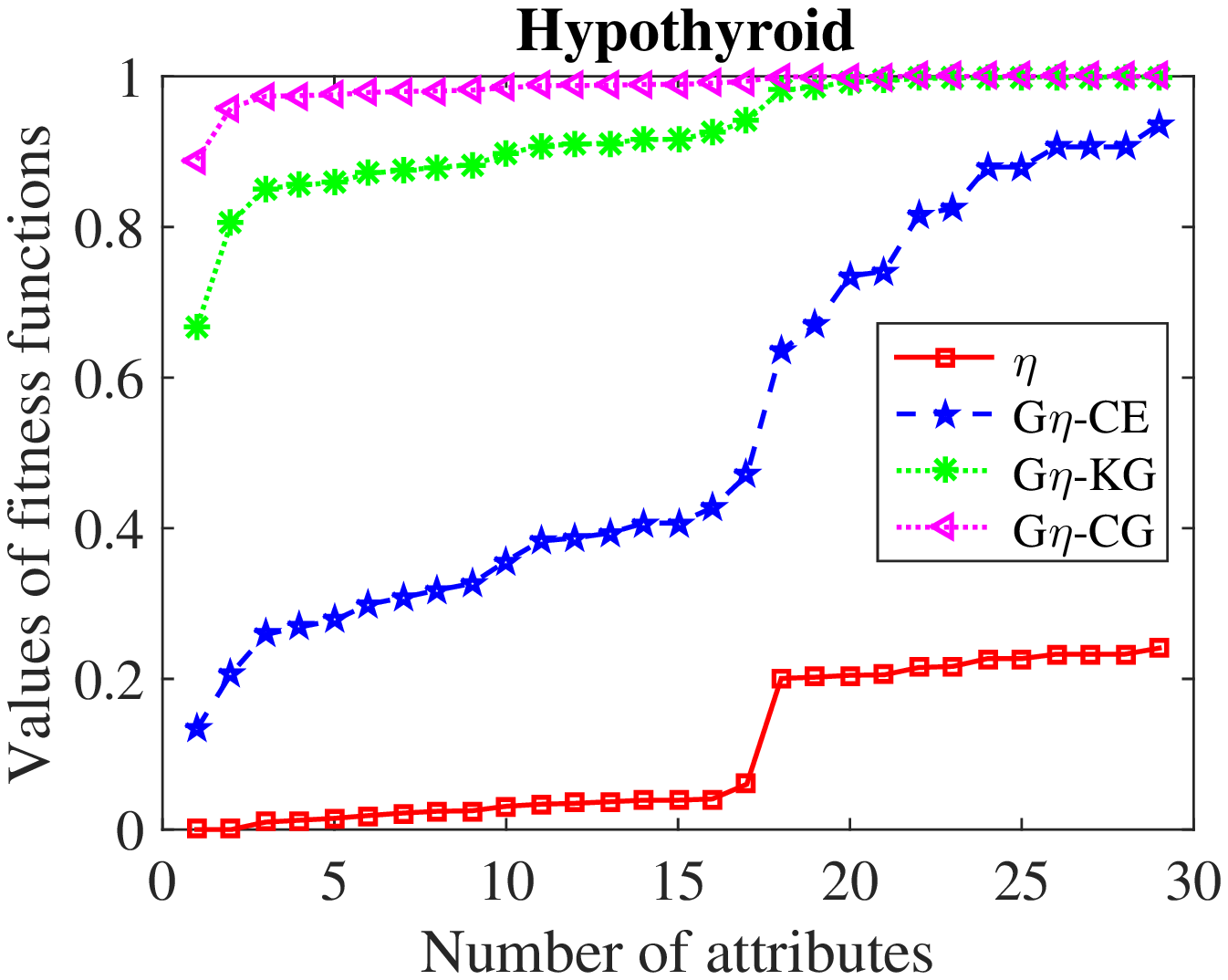}}
  \subfigure[$\mu$ and $G\mu$]{
    \label{fig:subfig:b}
    \includegraphics[width=2.5in]{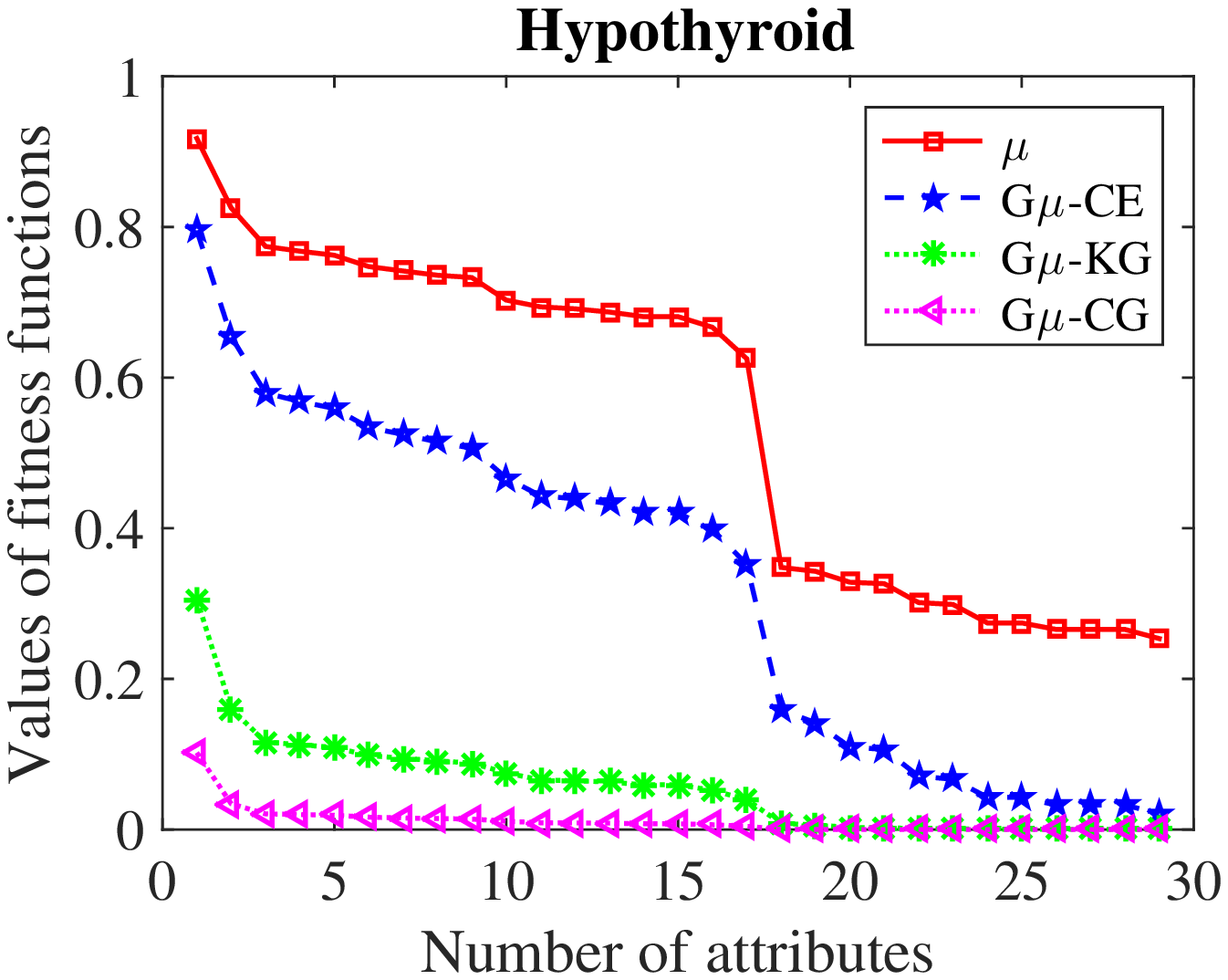}}
    \caption{Monotonicity of fitness functions on data set Hypothyroid.}
  \label{fig:Hypothyroid}
\end{figure}

\begin{figure}[h]
  \subfigure[$\eta$ and $G\eta$]{
    \label{fig:subfig:a}
    \includegraphics[width=2.5in]{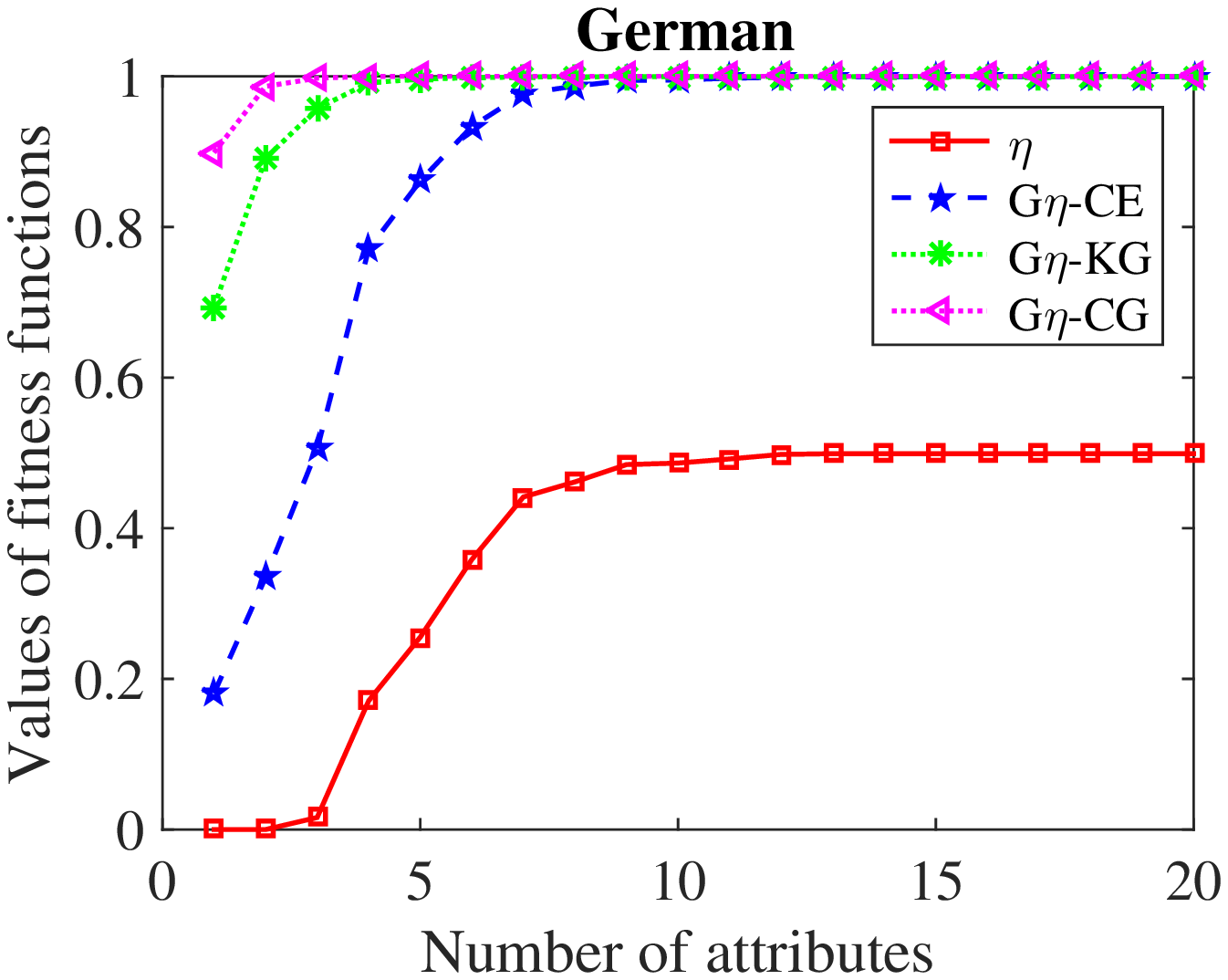}}
  \subfigure[$\mu$ and $G\mu$]{
    \label{fig:subfig:b}
    \includegraphics[width=2.5in]{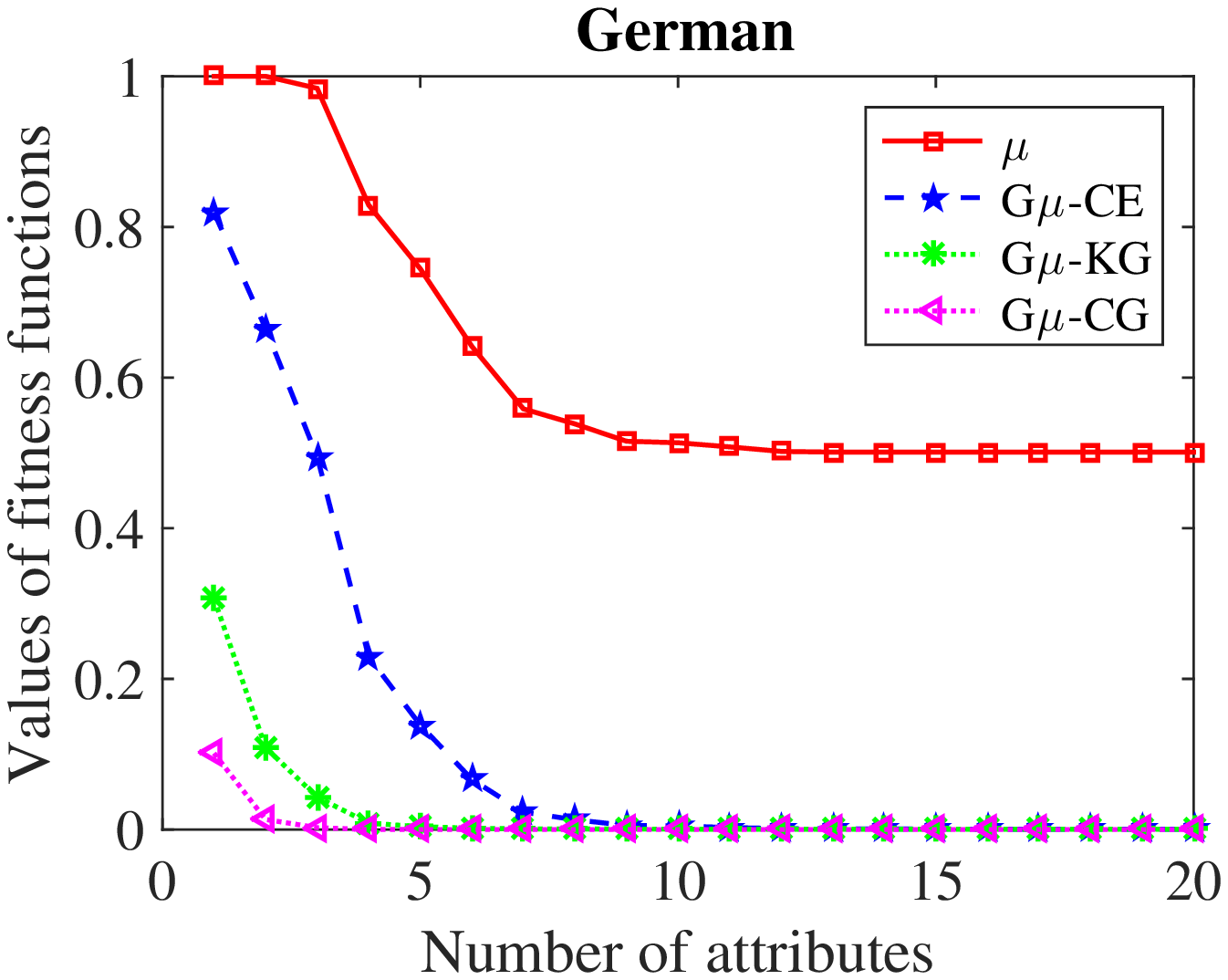}}
    \caption{Monotonicity of fitness functions on data set German.}
  \label{fig:German}
\end{figure}

Figures \ref{fig:Horse-colic} - \ref{fig:German} present the
experimental results of the proposed fitness functions on ten data
sets. In each of figures, the X-axis represents the size of
condition attribute subset. The condition attribute subset is
increased from one attribute to all attributes during the
experiments. The Y-axis pertains to values of fitness functions.
Furthermore, each figure has two subfigures. The subfigure (a) shows
that the experimental results of the fitness functions $\eta$,
$G\eta-CE$, $G\eta-KG$ and $G\eta-CG$, where $G\eta-CE$, $G\eta-KG$
and $G\eta-CG$ represent the CE-based, KG-based and CG-based
implementations of $G\eta$ respectively. The subfigure (b) shows
that the experimental results of the fitness functions $\mu$,
$G\mu-CE$, $G\mu-KG$ and $G\mu-CG$, where $G\mu-CE$, $G\mu-KG$ and
$G\mu-CG$ represent the CE-based, KG-based and CG-based
implementations of $G\mu$ respectively. Moreover, we rescaled the
values of them to the [0, 1] range in order to better visualize the
data because the values of the fitness functions $G\eta-CE$ and
$G\mu-CE$ may be greater than 1.

It can be seen from these figures that the values of the fitness
functions $\eta$, $G\eta-CE$, $G\eta-KG$ and $G\eta-CG$ increase
with the number of selected attributes becoming bigger, and the
values of the fitness functions $\mu$, $G\mu-CE$, $G\mu-KG$ and
$G\mu-CG$ decrease with the number of selected attributes becoming
bigger. It indicates that the proposed fitness functions are
monotonic with respect to the set inclusion of attributes. The
results are consistent with Corollary \ref{MonotonicityCorollary}
and Corollary \ref{ModifiedMonotonicityCorollary}. However, it is
easy to see that the values of the fitness function $\eta$ are the
same when the number of attributes increased from 1 to 2 on all the
data sets except Zoo. Similarly, there is no change in the results
of the fitness function $\mu$ as the number of attributes increases
from 1 to 2 on some data sets, such as Horse-colic, Voting and
Kr-vs-kp. In these cases, the fitness functions $\eta$ and $\mu$ can
not evaluate the significance of attributes effectively. For the
same situation, the values of the fitness functions $G\eta-CE$,
$G\eta-KG$ and $G\eta-CG$ get bigger, while the values of the
fitness functions $G\mu-CE$, $G\mu-KG$ and $G\mu-CG$ get smaller
when the number of attributes increased from 1 to 2. It shows that
the fitness functions $G\eta$ and $G\mu$ can evaluate the
significance of attributes more accurately. The results show that
the fitness functions $G\eta$ and $G\mu$ can provide more
information for evaluating the significance of attributes. In other
words, the fitness functions $G\eta$ and $G\mu$ have a better
discrimination power than the fitness functions $\eta$ and $\mu$.
Hence, we can evaluate the significance of attributes more
effectively by using the fitness functions $G\eta$ and $G\mu$.

\subsection{Significance of single attributes} \label{SingleExperiment}
Evaluating single attributes and ranking them are an important step
in Algorithms \ref{algorithm: addition-deletion} and \ref{algorithm:
deletion}. Hence, in this subsection, we compared the effectiveness
of the proposed fitness functions in evaluating the significance of
single attributes.

\begin{figure}[!ht]\hspace{-2cm}
\includegraphics[width=1.3\textwidth]{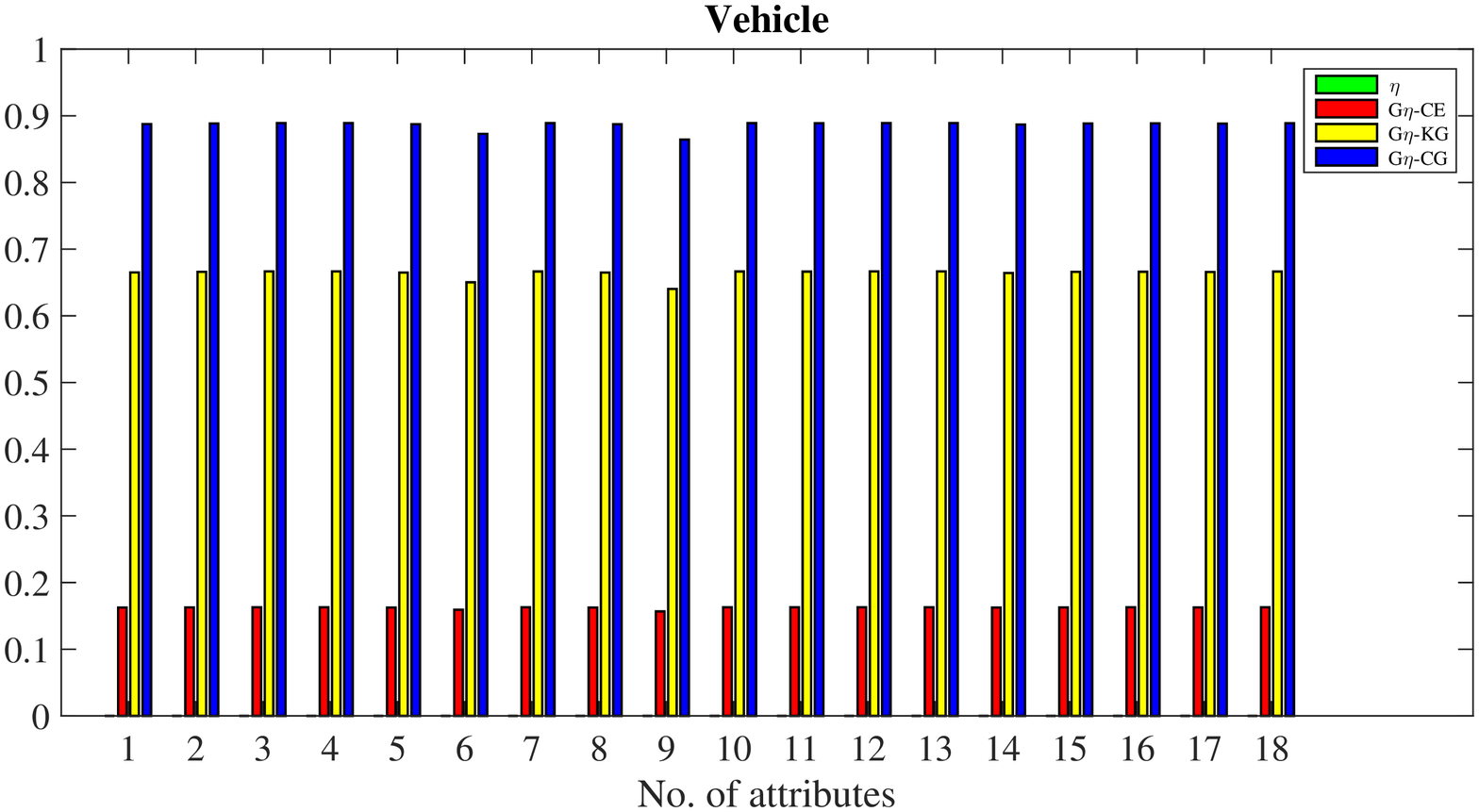}
\vspace{-0.4cm} \caption{Significance of a single attribute computed
with different fitness functions}
\label{figure:CreditApprovalLowerBar}
\end{figure}

\begin{figure}[!ht]\hspace{-2cm}
\includegraphics[width=1.3\textwidth]{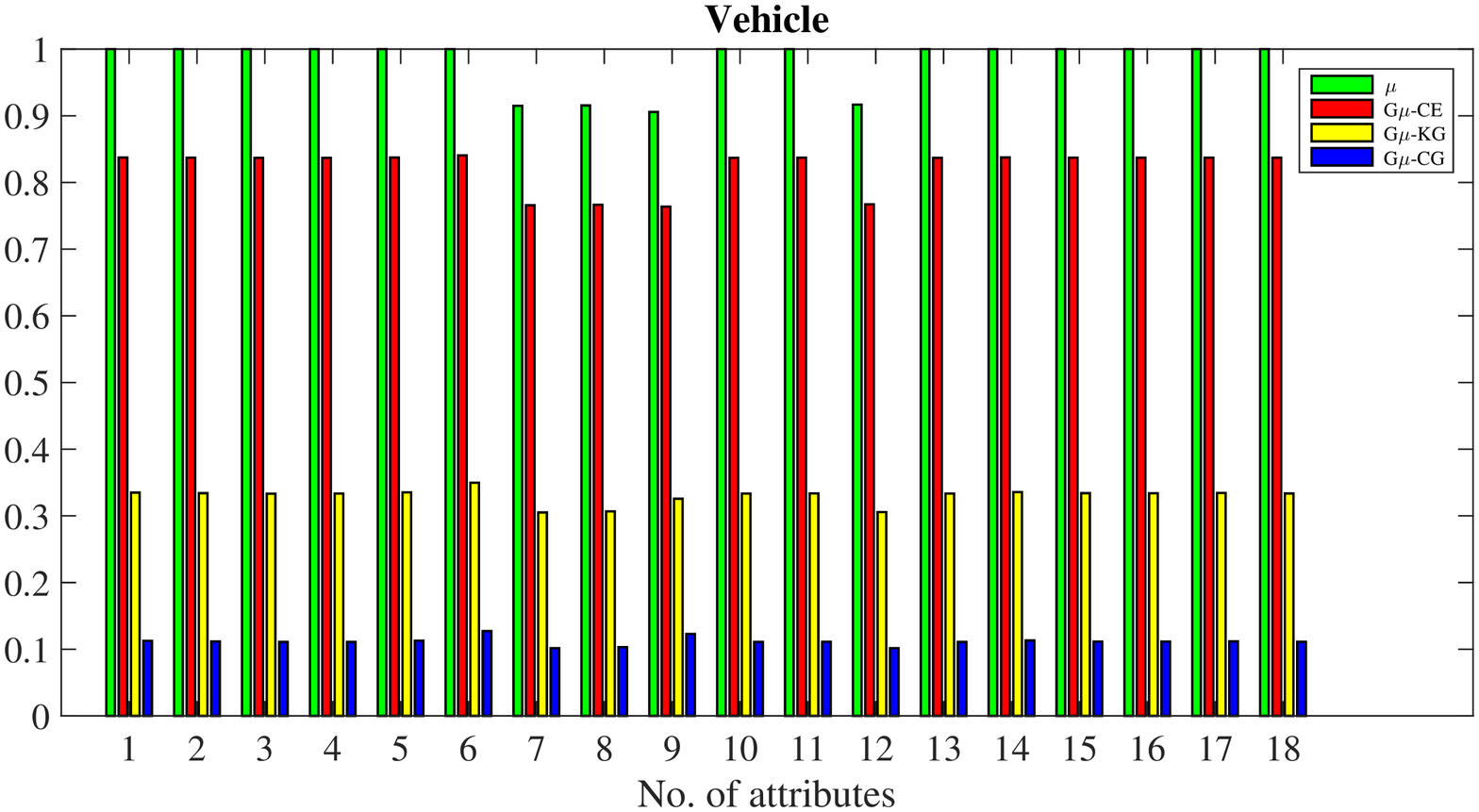}
\vspace{-0.4cm} \caption{Significance of a single attribute computed
with different fitness functions}
\label{figure:CreditApprovalUpperBar}
\end{figure}

\begin{figure}[!ht]\hspace{-2cm}
\includegraphics[width=1.3\textwidth]{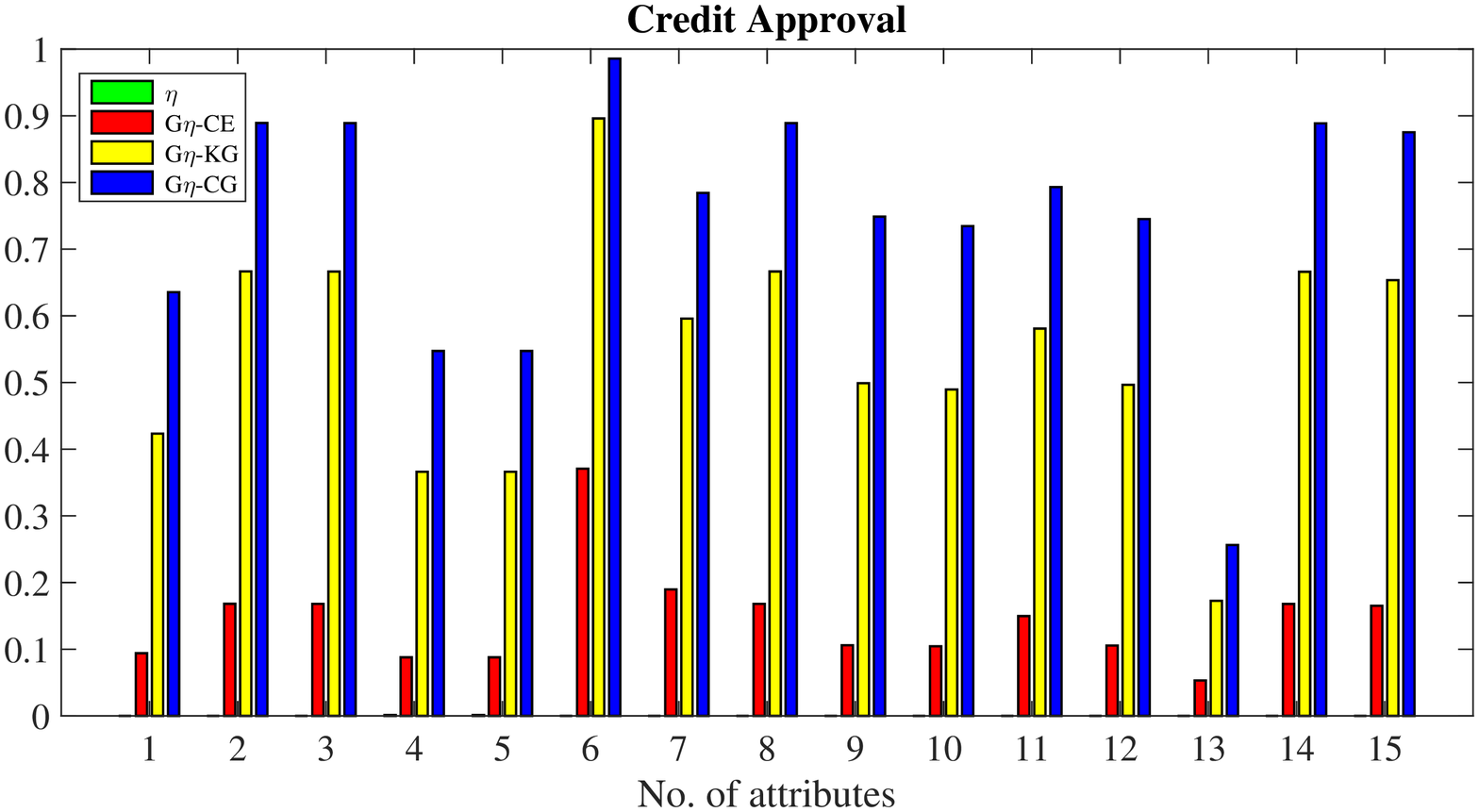}
\vspace{-0.4cm} \caption{Significance of a single attribute computed
with different fitness functions} \label{figure:VehicleLowerBar}
\end{figure}

\begin{figure}[!ht]\hspace{-2cm}
\includegraphics[width=1.3\textwidth]{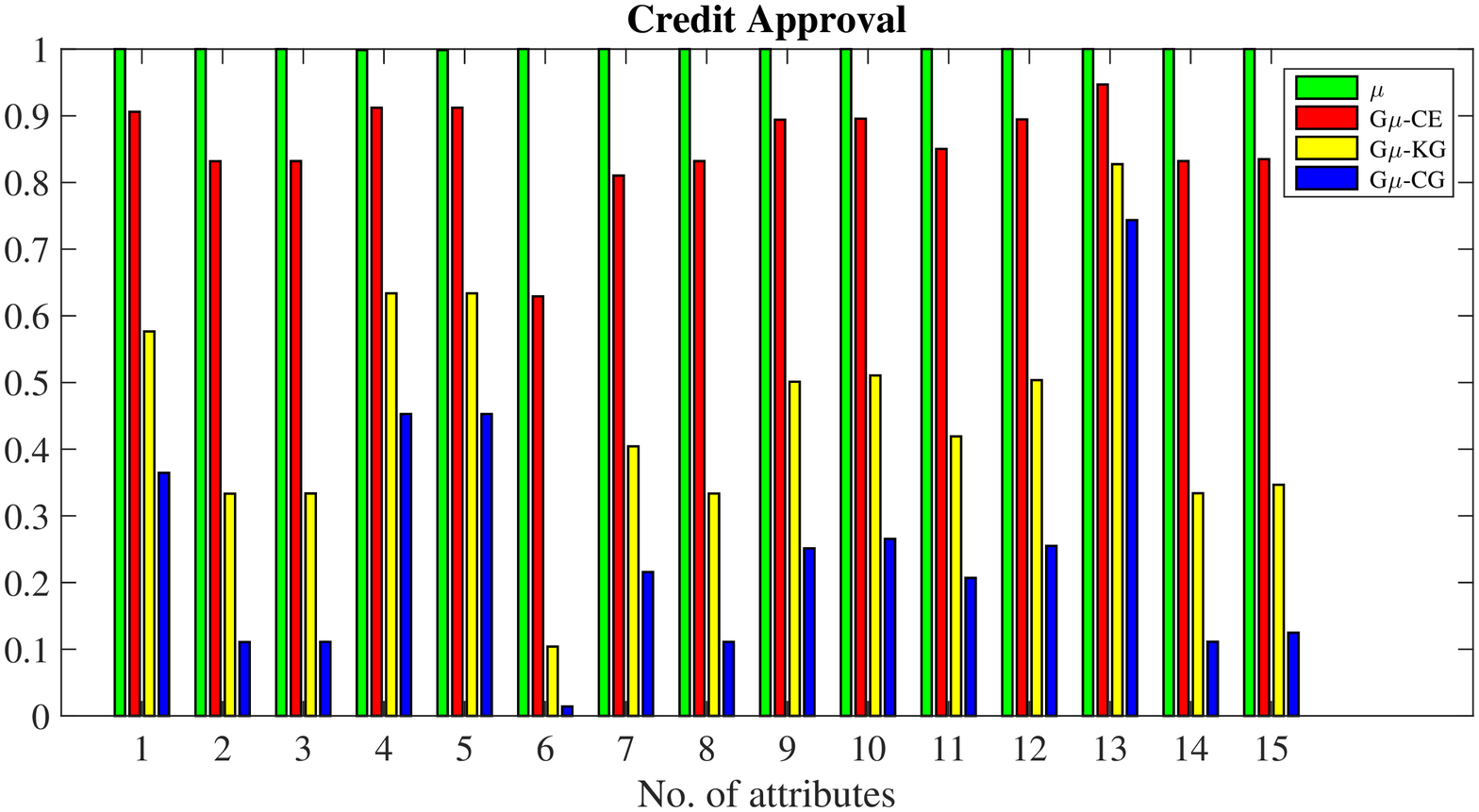}
\vspace{-0.4cm} \caption{Significance of a single attribute computed
with different fitness functions} \label{figure:VehicleUpperBar}
\end{figure}

Two data sets Vehicle and Credit Approval were used in experiments.
There are 18 attributes in Vehicle and 15 attributes in Credit
Approval. The experimental results of the fitness functions $\eta$,
$G\eta-CE$, $G\eta-KG$ and $G\eta-CG$ on two data sets are shown
Figures \ref{figure:CreditApprovalLowerBar} and
\ref{figure:VehicleLowerBar} respectively. The experimental results
of the fitness functions $\mu$, $G\mu-CE$, $G\mu-KG$ and $G\mu-CG$
on two data sets are shown Figures
\ref{figure:CreditApprovalUpperBar} and \ref{figure:VehicleUpperBar}
respectively. Similar to the monotonicity experiments, the values of
the fitness functions $G\eta-CE$ and $G\mu-CE$ were rescaled to the
[0, 1] range.

As to data set Vehicle, for each single attribute, the values of the
fitness function $\eta$ are equal to zero. It indicates that the
fitness function $\eta$ can not evaluate the single attributes
effectively. In this situation, we can not rank attributes by the
fitness function $\eta$. In comparison, the fitness functions
$G\eta-CE$, $G\eta-KG$ and $G\eta-CG$ can differentiate the
different single attributes effectively. In other words, we can rank
attributes effectively by the fitness function $G\eta$. Similarly,
for the most single attributes, the values of the fitness function
$\mu$ are equal to 1. Hence, we can not rank attributes by the
fitness function $\mu$, but we can rank attributes by the fitness
functions $G\mu-CE$, $G\mu-KG$ and $G\mu-CG$. As to data set Credit
Approval, we can obtain a similar result. The above experimental
results show that the fitness functions $\eta$ and $\mu$ are not
appropriate to evaluate the single attributes and rank them
sometimes, and the fitness functions $G\eta$ and $G\mu$ can evaluate
the single attributes and rank them effectively.

\subsection{The classification accuracy experiments} \label{AccuracyExperiment}
In this subsection, we took the $(\alpha, \beta)$ low distribute
reduct as an example of the distribute reducts. Then we compared the
classification accuracy of the $(\alpha, \beta)$ low distribution
reduct with three traditional definitions of attribute reduct,
including the qualitative positive region-preserved reduct (QLPRP)
\cite{Zhao2011NARDTRS}, the quantitative positive region-preserved
reduct (QNPRP) \cite{Zhao2011NARDTRS} and the positive region
extension reduct (EXDPR) \cite{Li2011NMARDTRS}, in probabilistic
rough set model. For each definition, the corresponding reducts are
obtained by the addition-deletion method and the deletion method.

For the $(\alpha, \beta)$ low distribution reduct, in Algorithms
\ref{algorithm: addition-deletion} and \ref{algorithm: deletion}, we
considered three different implementations of $G\eta$ by taking CE,
KG and CG as the examples of the expected granularity. In addition,
for Algorithm \ref{algorithm: addition-deletion} and \ref{algorithm:
deletion}, the $(\alpha, \beta)$ low distribution reducts obtained
by three different implementations are denoted as LDRCE, LDRKG and
LDRCG respectively.

In the experiments, two well-known used classifiers including
BayesNet and linear SVM (SMO) were selected to evaluate the
different definitions of attribute reduct. 10-fold cross-validation
scheme was used to assess the performance of the classifiers. All
approaches were implemented based on the WEKA data mining software
package \cite{WEKA}, where the classifiers were implemented with
default settings.

To make a fair comparison, we took the value of $\alpha$ from 0.1 to
1.0 with step 0.1. For each data set, the corresponding attribute
reducts can be got according to the different $\alpha$. Then we
computed the classification accuracies of all attribute reducts
using BayesNet and SMO based on 10-fold cross-validation. The
average value and standard deviation were recorded as the final
classification accuracies.

Tables \ref{tab:AccuracyByAdditionDeletionBayesNet} -
\ref{tab:AccuracyByDeletionSMO} show the experimental results of
classification accuracies. In each table, the average maximum
classification accuracies are depicted in bold. The classification
accuracies are performed on the raw data sets also.

\tabcolsep 0.022in
\begin{table} [!htb]\fontsize{6.5pt}{\baselineskip}\selectfont
 \caption{Classification accuracies based on BayesNet by
addition-deletion method} \vspace{0.3cm} \hspace{-0.9cm}
\label{tab:AccuracyByAdditionDeletionBayesNet}
\begin{tabular}{ccr@{ $\pm$ }lr@{ $\pm$ }lr@{ $\pm$ }lr@{ $\pm$ }lr@{ $\pm$ }lr@{ $\pm$ }l}
\hline
ID & RawData & \multicolumn{2}{c}{QLPRP} & \multicolumn{2}{c}{QNPRP} & \multicolumn{2}{c}{PRER} & \multicolumn{2}{c}{LDRCE} & \multicolumn{2}{c}{LDRKG} & \multicolumn{2}{c}{LDRCG}\\
\hline
1 & 0.8071 & 0.7174 & 0.0910 & 0.7630 & 0.0653 & 0.8226 & 0.0112 & 0.8234 & 0.0000 & 0.8234 & 0.0000 & \textbf{0.8424} & \textbf{0.0000} \\
2 & 0.4690 & 0.4183 & 0.0841 & 0.4174 & 0.0860 & 0.4212 & 0.0884 & \textbf{0.4720} & \textbf{0.0000} & \textbf{0.4720} & \textbf{0.0000} & \textbf{0.4720} & \textbf{0.0000} \\
3 & 0.5201 & 0.5002 & 0.0793 & 0.4939 & 0.0736 & 0.4852 & 0.0751 & \textbf{0.5402} & \textbf{0.0000} & \textbf{0.5402} & \textbf{0.0000} & \textbf{0.5402} & \textbf{0.0000} \\
4 & 0.9011 & 0.7657 & 0.1520 & 0.7667 & 0.1537 & 0.7605 & 0.1219 & \textbf{0.9195} & \textbf{0.0000} & \textbf{0.9195} & \textbf{0.0000} & \textbf{0.9195} & \textbf{0.0000} \\
5 & 0.9406 & 0.7624 & 0.1343 & 0.7881 & 0.1549 & 0.7733 & 0.1816 & 0.9406 & 0.0000 & \textbf{0.9505} & \textbf{0.0000} & \textbf{0.9505} & \textbf{0.0000} \\
6 & 0.8623 & 0.7116 & 0.0740 & 0.7133 & 0.0757 & 0.7086 & 0.1535 & \textbf{0.7899} & \textbf{0.0000} & \textbf{0.7899} & \textbf{0.0000} & \textbf{0.7899} & \textbf{0.0000} \\
7 & 0.8130 & 0.6715 & 0.2357 & 0.6871 & 0.2078 & 0.6555 & 0.2235 & \textbf{0.8433} & \textbf{0.0000} & 0.8364 & 0.0000 & 0.8355 & 0.0000 \\
8 & 0.8792 & 0.7605 & 0.1045 & 0.7239 & 0.1694 & 0.6856 & 0.1737 & \textbf{0.8820} & \textbf{0.0000} & \textbf{0.8820} & \textbf{0.0000} & \textbf{0.8820} & \textbf{0.0000} \\
9 & 0.9176 & 0.9201 & 0.0020 & 0.9201 & 0.0020 & \textbf{0.9224} & \textbf{0.0014} & 0.9181 & 0.0000 & 0.9181 & 0.0000 & 0.9181 & 0.0000 \\
10 & 0.7450 & 0.7165 & 0.0165 & 0.7275 & 0.0275 & 0.7059 & 0.0331 & 0.7360 & 0.0000 & \textbf{0.7550} & \textbf{0.0000} & 0.7330 & 0.0000 \\
\hline
\end{tabular}
\end{table}

\tabcolsep 0.022in
\begin{table} [!htb]\fontsize{6.5pt}{\baselineskip}\selectfont
 \caption{Classification accuracies based on SMO by
addition-deletion method} \vspace{0.3cm} \hspace{-0.9cm}
\label{tab:AccuracyByAdditionDeletionSMO}
\begin{tabular}{ccr@{ $\pm$ }lr@{ $\pm$ }lr@{ $\pm$ }lr@{ $\pm$ }lr@{ $\pm$ }lr@{ $\pm$ }l}
\hline
ID & RawData & \multicolumn{2}{c}{QLPRP} & \multicolumn{2}{c}{QNPRP} & \multicolumn{2}{c}{PRER} & \multicolumn{2}{c}{LDRCE} & \multicolumn{2}{c}{LDRKG} & \multicolumn{2}{c}{LDRCG}\\
\hline
1 & 0.8071 & 0.7296 & 0.0992 & 0.7603 & 0.0559 & 0.8212 & 0.0102 & 0.8288 & 0.0000 & \textbf{0.8315} & \textbf{0.0000} & 0.8288 & 0.0000 \\
2 & 0.4690 & 0.4121 & 0.0855 & 0.4115 & 0.0866 & 0.4156 & 0.0863 & \textbf{0.4720} & \textbf{0.0000} & \textbf{0.4720} & \textbf{0.0000} & \textbf{0.4720} & \textbf{0.0000} \\
3 & 0.6749 & 0.5913 & 0.1377 & 0.5914 & 0.1377 & 0.5735 & 0.1462 & \textbf{0.6903} & \textbf{0.0000} & \textbf{0.6903} & \textbf{0.0000} & \textbf{0.6903} & \textbf{0.0000} \\
4 & 0.9586 & 0.7851 & 0.1713 & 0.7798 & 0.1664 & 0.7641 & 0.1271 & \textbf{0.9563} & \textbf{0.0000} & \textbf{0.9563} & \textbf{0.0000} & \textbf{0.9563} & \textbf{0.0000} \\
5 & 0.9604 & 0.7713 & 0.1447 & 0.7772 & 0.1442 & 0.7723 & 0.1807 & \textbf{0.9406} & \textbf{0.0000} & 0.9307 & 0.0000 & 0.9307 & 0.0000 \\
6 & 0.8536 & 0.7172 & 0.0803 & 0.7248 & 0.0871 & 0.7033 & 0.1483 & \textbf{0.8116} & \textbf{0.0000} & \textbf{0.8116} & \textbf{0.0000} & \textbf{0.8116} & \textbf{0.0000} \\
7 & 0.9203 & 0.7162 & 0.2690 & 0.7482 & 0.2420 & 0.7106 & 0.2610 & \textbf{0.9242} & \textbf{0.0000} & 0.9216 & 0.0000 & 0.9212 & 0.0000 \\
8 & 0.9543 & 0.7896 & 0.1389 & 0.7566 & 0.2021 & 0.7067 & 0.1991 & \textbf{0.9549} & \textbf{0.0000} & \textbf{0.9549} & \textbf{0.0000} & \textbf{0.9549} & \textbf{0.0000} \\
9 & 0.9234 & 0.9236 & 0.0006 & 0.9236 & 0.0006 & 0.9230 & 0.0004 & \textbf{0.9244} & \textbf{0.0000} & \textbf{0.9244} & \textbf{0.0000} & \textbf{0.9244} & \textbf{0.0000} \\
10 & 0.7440 & 0.7140 & 0.0140 & 0.7255 & 0.0255 & 0.7078 & 0.0348 & 0.7440 & 0.0000 & \textbf{0.7510} & \textbf{0.0000} & 0.7280 & 0.0000 \\
\hline
\end{tabular}
\end{table}

\tabcolsep 0.022in
\begin{table} [!htb]\fontsize{6.5pt}{\baselineskip}\selectfont
 \caption{Classification accuracies based on BayesNet by
deletion method} \vspace{0.3cm} \hspace{-0.9cm}
\label{tab:AccuracyByDeletionBayesNet}
\begin{tabular}{ccr@{ $\pm$ }lr@{ $\pm$ }lr@{ $\pm$ }lr@{ $\pm$ }lr@{ $\pm$ }lr@{ $\pm$ }l}
\hline
ID & RawData & \multicolumn{2}{c}{QLPRP} & \multicolumn{2}{c}{QNPRP} & \multicolumn{2}{c}{PRER} & \multicolumn{2}{c}{LDRCE} & \multicolumn{2}{c}{LDRKG} & \multicolumn{2}{c}{LDRCG}\\
\hline
1 & 0.7853 & 0.7242 & 0.0950 & 0.7522 & 0.0471 & 0.8098 & 0.0111 & \textbf{0.8288} & \textbf{0.0000} & \textbf{0.8288} & \textbf{0.0000} & 0.8098 & 0.0000 \\
2 & 0.4690 & 0.4212 & 0.0812 & 0.4227 & 0.0797 & 0.4283 & 0.0780 & \textbf{0.4720} & \textbf{0.0000} & \textbf{0.4720} & \textbf{0.0000} & \textbf{0.4720} & \textbf{0.0000} \\
3 & 0.5201 & 0.4986 & 0.0790 & 0.4948 & 0.0750 & 0.5030 & 0.0735 & \textbf{0.5402} & \textbf{0.0000} & \textbf{0.5402} & \textbf{0.0000} & \textbf{0.5402} & \textbf{0.0000} \\
4 & 0.9011 & 0.7720 & 0.1582 & 0.7720 & 0.1582 & 0.7963 & 0.1231 & \textbf{0.9310} & \textbf{0.0000} & \textbf{0.9310} & \textbf{0.0000} & \textbf{0.9310} & \textbf{0.0000} \\
5 & 0.9406 & 0.7762 & 0.1531 & 0.7762 & 0.1531 & 0.7941 & 0.1727 & \textbf{0.9505} & \textbf{0.0000} & \textbf{0.9505} & \textbf{0.0000} & \textbf{0.9505} & \textbf{0.0000} \\
6 & 0.8623 & 0.7116 & 0.0740 & 0.7225 & 0.0883 & 0.7123 & 0.1573 & 0.7884 & 0.0000 & \textbf{0.7899} & \textbf{0.0000} & \textbf{0.7899} & \textbf{0.0000} \\
7 & 0.8130 & 0.6885 & 0.2304 & 0.6824 & 0.2303 & 0.6648 & 0.2280 & 0.8355 & 0.0000 & 0.8355 & 0.0000 & \textbf{0.8398} & \textbf{0.0000} \\
8 & 0.8792 & 0.7713 & 0.1108 & 0.7239 & 0.1694 & 0.7065 & 0.1844 & \textbf{0.8820} & \textbf{0.0000} & \textbf{0.8820} & \textbf{0.0000} & \textbf{0.8820} & \textbf{0.0000} \\
9 & 0.9176 & \textbf{0.9201} & \textbf{0.0020} & 0.9196 & 0.0025 & 0.9194 & 0.0039 & 0.9181 & 0.0000 & 0.9181 & 0.0000 & 0.9181 & 0.0000 \\
10 & 0.7450 & 0.7180 & 0.0180 & 0.7275 & 0.0275 & 0.7043 & 0.0337 & 0.7360 & 0.0000 & \textbf{0.7550} & \textbf{0.0000} & \textbf{0.7550} & \textbf{0.0000} \\
\hline
\end{tabular}
\end{table}

\tabcolsep 0.022in
\begin{table} [!htb]\fontsize{6.5pt}{\baselineskip}\selectfont
 \caption{Classification accuracies based on SMO by
deletion method} \label{tab:AccuracyByDeletionSMO} \vspace{0.3cm}
\hspace{-0.9cm}
\begin{tabular}{ccr@{ $\pm$ }lr@{ $\pm$ }lr@{ $\pm$ }lr@{ $\pm$ }lr@{ $\pm$ }lr@{ $\pm$ }l}
\hline
ID & RawData & \multicolumn{2}{c}{QLPRP} & \multicolumn{2}{c}{QNPRP} & \multicolumn{2}{c}{PRER} & \multicolumn{2}{c}{LDRCE} & \multicolumn{2}{c}{LDRKG} & \multicolumn{2}{c}{LDRCG}\\
\hline
1 & 0.8071 & 0.7174 & 0.0910 & 0.7630 & 0.0653 & 0.8226 & 0.0112 & 0.8234 & 0.0000 & 0.8234 & 0.0000 & \textbf{0.8424} & \textbf{0.0000} \\
2 & 0.4690 & 0.4156 & 0.0813 & 0.4174 & 0.0791 & 0.4224 & 0.0770 & \textbf{0.4720} & \textbf{0.0000} & \textbf{0.4720} & \textbf{0.0000} & \textbf{0.4720} & \textbf{0.0000} \\
3 & 0.6749 & 0.5894 & 0.1376 & 0.5882 & 0.1370 & 0.6046 & 0.1270 & \textbf{0.6903} & \textbf{0.0000} & \textbf{0.6903} & \textbf{0.0000} & \textbf{0.6903} & \textbf{0.0000} \\
4 & 0.9586 & 0.7841 & 0.1703 & 0.7841 & 0.1703 & 0.8145 & 0.1409 & \textbf{0.9540} & \textbf{0.0000} & \textbf{0.9540} & \textbf{0.0000} & \textbf{0.9540} & \textbf{0.0000} \\
5 & 0.9604 & 0.7752 & 0.1518 & 0.7752 & 0.1518 & 0.7931 & 0.1719 & \textbf{0.9406} & \textbf{0.0000} & \textbf{0.9406} & \textbf{0.0000} & \textbf{0.9406} & \textbf{0.0000} \\
6 & 0.8536 & 0.7172 & 0.0803 & 0.7288 & 0.0919 & 0.7036 & 0.1486 & 0.8043 & 0.0000 & \textbf{0.8116} & \textbf{0.0000} & \textbf{0.8116} & \textbf{0.0000} \\
7 & 0.9203 & 0.7485 & 0.2548 & 0.7309 & 0.2633 & 0.7246 & 0.2651 & 0.9212 & 0.0000 & 0.9212 & 0.0000 & \textbf{0.9242} & \textbf{0.0000} \\
8 & 0.9543 & 0.8077 & 0.1472 & 0.7566 & 0.2021 & 0.7395 & 0.2173 & \textbf{0.9549} & \textbf{0.0000} & \textbf{0.9549} & \textbf{0.0000} & \textbf{0.9549} & \textbf{0.0000} \\
9 & 0.9234 & 0.9234 & 0.0005 & 0.9233 & 0.0008 & 0.9233 & 0.0005 & \textbf{0.9242} & \textbf{0.0000} & \textbf{0.9242} & \textbf{0.0000} & \textbf{0.9242} & \textbf{0.0000} \\
10 & 0.7440 & 0.7220 & 0.0220 & 0.7255 & 0.0255 & 0.7078 & 0.0348 & 0.7440 & 0.0000 & \textbf{0.7510} & \textbf{0.0000} & \textbf{0.7510} & \textbf{0.0000} \\
\hline
\end{tabular}
\end{table}

It is observed from Tables
\ref{tab:AccuracyByAdditionDeletionBayesNet} -
\ref{tab:AccuracyByDeletionSMO} that, for addition-deletion method,
LDRCE, LDRKG and LDRCG exhibited the best average classification
accuracy based on BayesNet in most cases. In detail, LDRCE and LDRCG
achieved the highest average classification accuracy on six data
sets, and LDRKG achieved the highest average classification accuracy
on seven data sets. It should be emphasized that LDRCE, LDRKG and
LDRCG jointly achieved the highest average classification accuracy
on five data sets. Moveover, EXDPR got the maximum average
classification accuracy on data set Hypothyroid. Compared with the
classification accuracy of raw data, LDRCE, LDRKG and LDRCG obtained
better classification performance on all the data sets except the
data sets Credit Approval and German. In most cases, QLPRPR, QNPRPR
and PRER decrease the classification accuracies of raw data to some
extent. In addition, the classification accuracies of LDRCE, LDRKG
and LDRCG was found to be similar or consistent. As to SMO, LDRCE,
LDRKG and LDRCG have better average classification accuracy than
QLPRP, QNPRP and PRER on all the data sets. Compared with the
classification accuracy of raw data, LDRCE, LDRKG and LDRCG obtained
better classification performance on more than half of data sets.
For the deletion method, the similar results can be obtained by the
classification accuracy based on BayesNet and SMO.

Furthermore, from Tables
\ref{tab:AccuracyByAdditionDeletionBayesNet} -
\ref{tab:AccuracyByDeletionSMO}, we can clearly notice that the
standard deviations of classification accuracies derived from LDRCE,
LDRKG and LDRCG equal to zero on all the data sets. It implies
Algorithms \ref{algorithm: addition-deletion} and \ref{algorithm:
deletion} are insensitive to threshold parameters. This can be
explained because the $(\alpha, \beta)$ low distribution reduct has
a more stringent reduction condition that preserves the $(\alpha,
\beta)$ low approximation of all decision classes. In fact, the
$(\alpha, \beta)$ low distribution reduct is equivalent to reduct
that keeps Pawlak positive region (Definition \ref{PositiveRegion})
unchanged when decision table is consistent ($PO{S_R}(D) = U$). It
means that for the consistent decision talbe, the classification
accuracies derived from LDRCE, LDRKG and LDRCG are the same for all
threshold parameters. What's more, for inconsistent decision table
($PO{S_R}(D) \ne U$), Algorithms \ref{algorithm: addition-deletion}
and \ref{algorithm: deletion} are also insensitive to threshold
parameters when each decision class contains too few inconsistent
objects because the $(\alpha, \beta)$ low distribution reduct must
keep the $(\alpha, \beta)$ low approximation of all decision classes
unchanged.

Table \ref{tab:ReductLengeByDeletion} and
\ref{tab:ReductLengeByAdditionDeletion} outlined the average length
of the derived reduct based on the addition-deletion method and the
deletion method. From Table \ref{tab:ReductLengeByDeletion} and
\ref{tab:ReductLengeByAdditionDeletion}, it can be seen that the
numbers of attributes obtained by QLPRPR, QNPRPR and PRER are less
than those obtained by LDRCE, LDRKG and LDRCG. The reason should be
attributed to the fact that the $(\alpha, \beta)$ low distribution
reduct have more stringent reduction condition than three other
definitions of attribute redcut. For length of the derived reduct,
the standard deviations of LDRCE, LDRKG and LDRCG also equal to zero
on all the data sets. It further indicates Algorithms
\ref{algorithm: addition-deletion} and \ref{algorithm: deletion} are
insensitive to threshold parameters. In addition, for QLPRPR, QNPRPR
and PRER, the relatively high standard deviations of reduct length
also show that QLPRPR, QNPRPR and PRER obtained by the
addition-deletion method and the deletion method are sensitive to
threshold parameters.

\tabcolsep 0.12in
\begin{table}[htbp] \small \fontsize{6.5pt}{\baselineskip}\selectfont
\vspace{-0.3cm} \caption{Average length of a reduct based on the
addition-deletion method} \label{tab:ReductLengeByDeletion}
\begin{center}
\begin{tabular}{cr@{ $\pm$ }lr@{ $\pm$ }lr@{ $\pm$ }lr@{ $\pm$ }lr@{ $\pm$ }lr@{ $\pm$ }l}
\hline
ID & \multicolumn{2}{c}{QLPRP} & \multicolumn{2}{c}{QNPRP} & \multicolumn{2}{c}{PRER} & \multicolumn{2}{c}{LDRCE} & \multicolumn{2}{c}{LDRKG} & \multicolumn{2}{c}{LDRCG}\\
\hline
1 & 5.5 & 4.5 & 4.6 & 4.4 & 3.7 & 4.1 & 10.0 & 0.0 & 10.0 & 0.0 & 10.0 & 0.0 \\
2 & 11.9 & 5.7 & 11.9 & 5.7 & 11.9 & 5.6 & 16.0 & 0.0 & 16.0 & 0.0 & 16.0 & 0.0 \\
3 & 9.5 & 6.1 & 9.4 & 6.1 & 8.6 & 6.5 & 15.0 & 0.0 & 15.0 & 0.0 & 15.0 & 0.0 \\
4 & 6.2 & 5.3 & 6.6 & 5.6 & 2.1 & 3.3 & 12.0 & 0.0 & 12.0 & 0.0 & 12.0 & 0.0 \\
5 & 3.8 & 2.9 & 3.4 & 2.4 & 3.0 & 2.0 & 7.0 & 0.0 & 6.0 & 0.0 & 6.0 & 0.0 \\
6 & 6.0 & 5.0 & 5.9 & 4.9 & 4.0 & 4.6 & 11.0 & 0.0 & 11.0 & 0.0 & 11.0 & 0.0 \\
7 & 8.6 & 6.4 & 9.4 & 5.1 & 7.6 & 5.3 & 15.0 & 0.0 & 14.0 & 0.0 & 14.0 & 0.0 \\
8 & 12.3 & 13.6 & 15.0 & 14.0 & 9.6 & 12.7 & 29.0 & 0.0 & 29.0 & 0.0 & 29.0 & 0.0 \\
9 & 14.5 & 9.5 & 14.2 & 9.3 & 3.2 & 6.6 & 23.0 & 0.0 & 23.0 & 0.0 & 23.0 & 0.0 \\
10 & 5.5 & 4.5 & 5.5 & 4.5 & 4.9 & 4.8 & 10.0 & 0.0 & 10.0 & 0.0 & 10.0 & 0.0 \\
\hline
\end{tabular}
\end{center}
\end{table}

\tabcolsep 0.12in
\begin{table}[htbp] \small \fontsize{6.5pt}{\baselineskip}\selectfont
\vspace{-0.6cm} \caption{Average length of a reduct based on the
deletion method} \label{tab:ReductLengeByAdditionDeletion}
\begin{center}
\begin{tabular}{cr@{ $\pm$ }lr@{ $\pm$ }lr@{ $\pm$ }lr@{ $\pm$ }lr@{ $\pm$ }lr@{ $\pm$ }l}
\hline
ID & \multicolumn{2}{c}{QLPRP} & \multicolumn{2}{c}{QNPRP} & \multicolumn{2}{c}{PRER} & \multicolumn{2}{c}{LDRCE} & \multicolumn{2}{c}{LDRKG} & \multicolumn{2}{c}{LDRCG}\\
\hline
1 &  5.1 & 4.1 & 5.5 & 4.5 & 5.5 & 4.5 & 9.0 & 0.0 & 9.0 & 0.0 & 10.0 & 0.0 \\
2 &  12.0 & 5.6 & 12.1 & 5.5 & 12.5 & 5.2 & 16.0 & 0.0 & 16.0 & 0.0 & 16.0 & 0.0 \\
3 &  9.6 & 6.1 & 9.4 & 6.1 & 9.8 & 5.6 & 15.0 & 0.0 & 15.0 & 0.0 & 15.0 & 0.0 \\
4 &  6.2 & 5.3 & 6.2 & 5.3 & 6.3 & 5.3 & 12.0 & 0.0 & 12.0 & 0.0 & 12.0 & 0.0 \\
5 &  2.9 & 2.1 & 2.9 & 2.1 & 3.2 & 1.9 & 6.0 & 0.0 & 6.0 & 0.0 & 6.0 & 0.0 \\
6 &  6.0 & 5.0 & 6.0 & 5.0 & 5.9 & 4.9 & 11.0 & 0.0 & 11.0 & 0.0 & 11.0 & 0.0 \\
7 &  8.7 & 5.6 & 8.7 & 6.2 & 8.2 & 5.2 & 14.0 & 0.0 & 14.0 & 0.0 & 14.0 & 0.0 \\
8 &  15.0 & 14.0 & 15.0 & 14.0 & 15.0 & 14.0 & 29.0 & 0.0 & 29.0 & 0.0 & 29.0 & 0.0 \\
9 &  14.5 & 9.5 & 14.1 & 9.3 & 12.6 & 9.7 & 23.0 & 0.0 & 23.0 & 0.0 & 23.0 & 0.0 \\
10 & 5.5 & 4.5 & 5.5 & 4.5 & 6.0 & 5.1 & 10.0 & 0.0 & 10.0 & 0.0 & 10.0 & 0.0 \\
\hline
\end{tabular}
\end{center}
\end{table}

From the experimental results, we can draw a conclusion that the
distribution reducts are relatively the better choices in
probabilistic rough set model, as attributes they selected have the
higher classification accuracy.

\subsection{Comparison of distribution reducts with ranking based attribute reduction methods} \label{CDRRBAR}
In this subsection, we compared the classification accuracy of the
distribution reducts with ranking based attribute reduction methods
(RBAR). RBAR is a classic attribute reduction algorithm. In RBAR,
the attributes are globally ranked based on the significance of
single attributes, and the first k best attributes are selected. In
our experiments, the significance of single attributes was evaluated
by using the fitness functions $G\eta-CE$, $G\eta-KG$ and
$G\eta-CG$, and the different values of k were specified based on
the subset size of the corresponding $(\alpha, \beta)$ low
distribution reducts, which are obtained by the size of LDRCE, LDRKG
and LDRCG, to compare the distribution reducts with ranking based
attribute reduction methods. The classification accuracies were
obtained by adopting the same experimental settings and methods that
were used in subsection \ref{AccuracyExperiment}.

Tables
\ref{tab:AccuracyByFilterAdditionDeletionBayesNet}-\ref{tab:AccuracyByFilterDeletionSMO}
present experimental results of RBAR on ten data sets using BayesNet
and SMO, where RBARCE, RBARKG and RBARCG represent the attribute
reducts obtained by using $G\eta-CE$, $G\eta-KG$ and $G\eta-CG$ to
evaluate the significance of single attributes, respectively.
Compare with Tables \ref{tab:AccuracyByAdditionDeletionBayesNet} -
\ref{tab:AccuracyByDeletionSMO}, it is easy to see that, for the
$(\alpha, \beta)$ low distribution reducts obtained by the
addition-deletion method, LDRCE and LDRCG outperform RBARCE and
RBARCG on seven data sets respectively, and LDRKG outperform RBARKG
on eight data sets according to BayesNet. Note that LDRCE, LDRKG and
LDRCG get the same average classification accuracy on data set
Primary-tumor with RBARCE, RBARKG and RBARCG respectively. This is
because the $(\alpha, \beta)$ low distribution reducts obtained by
the addition-deletion method and reducts obtained by ranking method
are the same. As to SMO, the results is similar to that of BayesNet.
For the $(\alpha, \beta)$ low distribution reducts obtained by the
deletion method, we can obtain the similar results also.

Nevertheless, from Table \ref{tab:ReductLengeByDeletion} and
\ref{tab:ReductLengeByAdditionDeletion}, we can easily find that the
standard deviations of classification accuracies derived from
RBARCE, RBARKG and RBARCG equal to zero on all the data sets. This
is because, for given 10 different values of $\alpha$, we always
obtain the same ranking of attributes according to the fitness
function $G\eta$, and the subset size of the $(\alpha, \beta)$ low
distribution reduct obtained by the addition-deletion method or
deletion method is same every time. It further indicates that
Algorithms \ref{algorithm: addition-deletion} and \ref{algorithm:
deletion} are insensitive to threshold parameters.

The above experimental results show that the distribution reducts
are effective.

\tabcolsep 0.022in
\begin{table} [tbp]\fontsize{9pt}{\baselineskip}\selectfont
 \caption{Classification accuracies based on BayesNet by
ranking method (subset size obtained by addition-deletion method)}
\label{tab:AccuracyByFilterAdditionDeletionBayesNet}
\begin{center}
\begin{tabular}{cr@{ $\pm$ }lr@{ $\pm$ }lr@{ $\pm$ }lr@{ $\pm$ }l}
\hline
ID & \multicolumn{2}{c}{RBARCE} & \multicolumn{2}{c}{RBARKG} & \multicolumn{2}{c}{RBARCG}\\
\hline
1 & 0.7745 & 0.0000 & 0.7147 & 0.0000 & 0.7147 & 0.0000 \\
2 & 0.4720 & 0.0000 & 0.4720 & 0.0000 & 0.4720 & 0.0000 \\
3 & 0.4504 & 0.0000 & 0.4504 & 0.0000 & 0.4504 & 0.0000 \\
4 & 0.8943 & 0.0000 & 0.8943 & 0.0000 & 0.8943 & 0.0000 \\
5 & 0.9307 & 0.0000 & 0.8416 & 0.0000 & 0.8416 & 0.0000 \\
6 & 0.8652 & 0.0000 & 0.8652 & 0.0000 & 0.8652 & 0.0000 \\
7 & 0.8260 & 0.0000 & 0.8312 & 0.0000 & 0.8212 & 0.0000 \\
8 & 0.8767 & 0.0000 & 0.8767 & 0.0000 & 0.8767 & 0.0000 \\
9 & 0.9178 & 0.0000 & 0.9178 & 0.0000 & 0.9223 & 0.0000 \\
10 & 0.7430 & 0.0000 & 0.7420 & 0.0000 & 0.7420 & 0.0000 \\
\hline
\end{tabular}
\end{center}
\end{table}

\tabcolsep 0.022in
\begin{table} [h]\fontsize{9pt}{\baselineskip}\selectfont
 \caption{Classification accuracies based on SMO by
ranking method (subset size obtained by addition-deletion method)}
\label{tab:AccuracyByFilterAdditionDeletionSMO}
\begin{center}
\begin{tabular}{cr@{ $\pm$ }lr@{ $\pm$ }lr@{ $\pm$ }lr@{ $\pm$ }l}
\hline
ID & \multicolumn{2}{c}{RBARCE} & \multicolumn{2}{c}{RBARKG} & \multicolumn{2}{c}{RBARCG}\\
\hline
1 & 0.7391 & 0.0000 & 0.7147 & 0.0000 & 0.7147 & 0.0000 \\
2 & 0.4720 & 0.0000 & 0.4720 & 0.0000 & 0.4720 & 0.0000 \\
3 & 0.6040 & 0.0000 & 0.6040 & 0.0000 & 0.6040 & 0.0000 \\
4 & 0.9563 & 0.0000 & 0.9563 & 0.0000 & 0.9563 & 0.0000 \\
5 & 0.9307 & 0.0000 & 0.8416 & 0.0000 & 0.8416 & 0.0000 \\
6 & 0.8551 & 0.0000 & 0.8551 & 0.0000 & 0.8551 & 0.0000 \\
7 & 0.9216 & 0.0000 & 0.9203 & 0.0000 & 0.9048 & 0.0000 \\
8 & 0.9528 & 0.0000 & 0.9528 & 0.0000 & 0.9528 & 0.0000 \\
9 & 0.9236 & 0.0000 & 0.9236 & 0.0000 & 0.9229 & 0.0000 \\
10 & 0.7490 & 0.0000 & 0.7520 & 0.0000 & 0.7520 & 0.0000 \\
\hline
\end{tabular}
\end{center}
\end{table}

\tabcolsep 0.022in
\begin{table} [h]\fontsize{9pt}{\baselineskip}\selectfont
 \caption{Classification accuracies based on BayesNet by
ranking method(subset size obtained by deletion method)}
\label{tab:AccuracyByFilterDeletionBayesNet}
\begin{center}
\begin{tabular}{cr@{ $\pm$ }lr@{ $\pm$ }lr@{ $\pm$ }lr@{ $\pm$ }l}
\hline
ID & \multicolumn{2}{c}{RBARCE} & \multicolumn{2}{c}{RBARKG} & \multicolumn{2}{c}{RBARCG}\\
\hline
1 & 0.7663 & 0.0000 & 0.7092 & 0.0000 & 0.7147 & 0.0000 \\
2 & 0.4720 & 0.0000 & 0.4720 & 0.0000 & 0.4720 & 0.0000 \\
3 & 0.4504 & 0.0000 & 0.4504 & 0.0000 & 0.4504 & 0.0000 \\
4 & 0.8943 & 0.0000 & 0.8943 & 0.0000 & 0.8943 & 0.0000 \\
5 & 0.8416 & 0.0000 & 0.8416 & 0.0000 & 0.8416 & 0.0000 \\
6 & 0.8652 & 0.0000 & 0.8652 & 0.0000 & 0.8652 & 0.0000 \\
7 & 0.8312 & 0.0000 & 0.8312 & 0.0000 & 0.8087 & 0.0000 \\
8 & 0.8767 & 0.0000 & 0.8767 & 0.0000 & 0.8767 & 0.0000 \\
9 & 0.9178 & 0.0000 & 0.9178 & 0.0000 & 0.9223 & 0.0000 \\
10 & 0.7430 & 0.0000 & 0.7420 & 0.0000 & 0.7420 & 0.0000 \\
\hline
\end{tabular}
\end{center}
\end{table}

\tabcolsep 0.022in
\begin{table} [h]\fontsize{9pt}{\baselineskip}\selectfont
 \caption{Classification accuracies based on SMO by
ranking method(subset size obtained by deletion method)}
\label{tab:AccuracyByFilterDeletionSMO}
\begin{center}
\begin{tabular}{cr@{ $\pm$ }lr@{ $\pm$ }lr@{ $\pm$ }lr@{ $\pm$ }l}
\hline
ID & \multicolumn{2}{c}{RBARCE} & \multicolumn{2}{c}{RBARKG} & \multicolumn{2}{c}{RBARCG}\\
\hline
1 & 0.7527 & 0.0000 & 0.6902 & 0.0000 & 0.7147 & 0.0000 \\
2 & 0.4720 & 0.0000 & 0.4720 & 0.0000 & 0.4720 & 0.0000 \\
3 & 0.6040 & 0.0000 & 0.6040 & 0.0000 & 0.6040 & 0.0000 \\
4 & 0.9563 & 0.0000 & 0.9563 & 0.0000 & 0.9563 & 0.0000 \\
5 & 0.8416 & 0.0000 & 0.8416 & 0.0000 & 0.8416 & 0.0000 \\
6 & 0.8551 & 0.0000 & 0.8551 & 0.0000 & 0.8551 & 0.0000 \\
7 & 0.9203 & 0.0000 & 0.9203 & 0.0000 & 0.9078 & 0.0000 \\
8 & 0.9528 & 0.0000 & 0.9528 & 0.0000 & 0.9528 & 0.0000 \\
9 & 0.9236 & 0.0000 & 0.9236 & 0.0000 & 0.9229 & 0.0000 \\
10 & 0.7490 & 0.0000 & 0.7520 & 0.0000 & 0.7520 & 0.0000 \\
\hline
\end{tabular}
\end{center}
\end{table}

\section{Conclusion} \label{conclusion}
The addition-deletion method based and deletion method based
attribute reduction algorithms are two representative heuristic
attribute reduction algorithms in rough set theory. The fitness
functions play a crucial role in designing heuristic attribute
reduction algorithms. The monotonicity of fitness functions is very
important to guarantee the validity of heuristic attribute reduction
algorithms. This paper aims at developing heuristic attribute
reduction algorithms for finding distribution reducts in
probabilistic rough set model. To begin with, we proposed two
monotonic fitness functions $\eta _R^{(\alpha ,\beta )}$ and  $\mu
_R^{(\alpha ,\beta )}$. The equivalence definitions of the $(\alpha
,\beta )$ low and upper distribution reducts are given respectively
based on $\eta _R^{(\alpha ,\beta )}$ and $\mu _R^{(\alpha ,\beta
)}$. The equivalence definitions of two distribution reducts can be
use to design the stopping criteria of heuristic attribute reduction
algorithms. Additionally, we presented two modified fitness
functions $G\eta _R^{(\alpha ,\beta )}$ and $G\mu _R^{(\alpha ,\beta
)}$ by multiplying measures of granularity of partitions by $\eta
_R^{(\alpha ,\beta )}$ and $\mu _R^{(\alpha ,\beta )}$ to evaluate
the significance of attributes more effectively. On this basis, we
developed two heuristic attribute reduction algorithms to find
distribution reducts based on the addition-deletion method and
deletion method. The monotonicity of fitness functions proposed
guarantees that the reduction result is right as well. Finally,
experiments on several real-life data sets are conducted to test the
effectiveness of the proposed fitness functions and distribution
reducts. The experimental results show that $G\eta _R^{(\alpha
,\beta )}$ and $G\mu _R^{(\alpha ,\beta )}$ are a more effective
alternative than $\eta _R^{(\alpha ,\beta )}$ and $\mu _R^{(\alpha
,\beta )}$ for evaluating the significance of attributes. The
experimental results also indicate that distribution reducts can
achieve better performance in probabilistic rough set model and
outperforms ranking based attribute reduction methods.

This investigation provides a new insight into the distribution
reducts in probabilistic rough set model. In the future, the study
work will be focused on designing heuristic attribute reduction
algorithms to find other definitions of attribute reduct in
probabilistic rough set model by studying monotonic fitness
functions.

\section*{Acknowledgements}
This work is supported by the National Natural Science Foundation of
China (Grant Nos. 61502419, 61272060, 61379114), and the Key Natural
Science Foundation of Chongqing (No. CSTC2013jjB40003).


\bibliographystyle{plain}
\bibliographystyle{elsarticle-num}
\bibliography{maxiao}

\end{document}